\documentclass{article}
\usepackage[a4paper,margin=0.75in,includehead,includefoot]{geometry}

\usepackage[numbers, compress]{natbib}

\makeatletter
\newif\if@restonecol
\makeatother

\usepackage[ruled,vlined]{algorithm2e}

\usepackage[utf8]{inputenc} 
\usepackage[T1]{fontenc}    
\usepackage[hyphens]{url}
\usepackage{hyperref}
\usepackage[hyphenbreaks]{breakurl}

\usepackage{booktabs}       
\usepackage{amsfonts}       
\usepackage{nicefrac}       
\usepackage{microtype}      
\usepackage{xcolor}         

\usepackage{listings}

\definecolor{codegreen}{rgb}{0,0.6,0}
\definecolor{codegray}{rgb}{0.5,0.5,0.5}
\definecolor{codepurple}{rgb}{0.58,0,0.82}
\definecolor{backcolour}{rgb}{0.95,0.95,0.92}

\lstdefinestyle{mystyle}{
    backgroundcolor=\color{backcolour},   
    commentstyle=\color{codegreen},
    keywordstyle=\color{magenta},
    numberstyle=\tiny\color{codegray},
    stringstyle=\color{codepurple},
    basicstyle=\ttfamily\footnotesize,
    breakatwhitespace=false,         
    breaklines=true,                 
    captionpos=b,                    
    keepspaces=true,                 
    showspaces=false,                
    showstringspaces=false,
    showtabs=false,                  
    tabsize=2
}

\usepackage{pgfplots}
\DeclareUnicodeCharacter{2212}{−}
\usepgfplotslibrary{groupplots,dateplot}
\usetikzlibrary{patterns,shapes.arrows}
\pgfplotsset{compat=newest}

\usepackage{longtable}[=v4.13]

\usepackage{multirow}
\usepackage{caption}
\usepackage{subcaption}
\usepackage{rotating}

\usepackage{amsmath}
\usepackage{amssymb}
\usepackage{mathtools}
\usepackage{amsthm}

\theoremstyle{plain}
\newtheorem{theorem}{Theorem}[section]

\newtheorem{lemma}[theorem]{Lemma}

\newtheorem{remark}[theorem]{Remark}
\theoremstyle{definition}
\newtheorem{definition}[theorem]{Definition}

\usepackage{physics}

\newcommand{\projunit}{\ensuremath\Pi_{[0, 1]}}

\title{Rethinking Log Odds: Linear Probability Modelling and Expert Advice in Interpretable Machine Learning} 

\author{%
  Danial Dervovic\\
  JP Morgan AI Research\\
  Edinburgh, UK \\
  \texttt{danial.dervovic@jpmchase.com} \\
  \and
  Nicol\'as Marchesotti\\
  JP Morgan AI Research\\
  London, UK \\
  \texttt{nicolas.p.marchesotti@jpmorgan.com} \\
  \and
  Freddy L\'ecu\'e\\
  JP Morgan AI Research\\
  New York City, NY, USA \\
  \texttt{freddy.lecue@jpmchase.com} \\
  \and
  Daniele Magazzeni\\
  JP Morgan AI Research\\
  London, UK \\
  \texttt{daniele.magazzeni@jpmorgan.com} \\
}

\begin{document}

\maketitle

\begin{abstract}
We introduce a family of interpretable machine learning models, with two broad additions: \emph{Linearised Additive Models (LAMs)} which replace the ubiquitous logistic link function in General Additive Models (GAMs); and \textsc{SubscaleHedge}, an expert advice algorithm for combining base models trained on subsets of features called \emph{subscales}. LAMs can augment any additive binary classification model equipped with a sigmoid link function. Moreover, they afford direct global and local attributions of additive components to the model output in probability space. We argue that LAMs and \textsc{SubscaleHedge} improve the interpretability of their base algorithms. Using rigorous null-hypothesis significance testing on a broad suite of financial modelling data, we show that our algorithms do not suffer from large performance penalties in terms of ROC-AUC and calibration.
\end{abstract}

\section{Introduction}
\label{sec:intro}

In sensitive domains such as finance and healthcare, there is renewed interest in \emph{inherently interpretable} models~\cite{Lipton2018,molnar2022,DARPA}, where the model form is such that it admits useful explanations of its output without any post-processing.
Within finance there is already increased regulatory scrutiny~\cite{OCC2021,EU2021,RFI2021} being introduced with regards to the usage of AI, incorporating demands for algorithmic decisions to be accompanied by rationales.
Furthermore, for AI models to be satisfactory for all stakeholders in this domain, there are also additional specific desiderata beyond interpretability, a subset of which we shall focus on in this paper.

Within consumer credit lending, monotonicity and other constraints are crucial for both interpretability and legal requirements in modelling~\cite{SR11-7}, with mathematical guarantees that as risk factors increase, a customer's computed risk should also increase.
Another important consideration within credit modelling is being able to attribute the decision of a model to one or more specific risk factors of a customer, also known as \emph{reason codes}~\cite{reason-codes}.
In the US this is enshrined in law~\cite{ecoa} whereby any negative credit decision requires up to four reason codes being specified.
In the EU the GDPR directive~\cite{regulation2016general} grants any person subject to an automated decision a `right to explanation'. 
From the perspective of a machine learning model, a risk factor is effectively a specific grouping of features and their values for a particular customer.
Such groupings are known as \emph{subscales}.
Classes of model reasoning explicitly about subscales lend themselves well to reason code-based explanations and allow for ``multi-level'' explanations, namely subscale-level explanations augmented with individual feature-level explanations within each subscale. 
Explanations of risk modelling decisions with respect to broad factors can also be less overwhelming for stakeholders~\cite{MILLER20191,fico-score}.

In this work we focus on improving two fundamental building blocks of interpretable modelling for binary classification: \emph{i.} modelling in log-odds space; and \emph{ii.} modelling subscales separately then combining in a hierarchical fashion.
In the first case, we will provide a detailed example demonstrating that even in the simplest modelling setting using log-odds, namely logistic regression, the model coefficients do not admit easy interpretation in the model output space (probabilities), contrary to received wisdom.
We then propose a scheme that aims to alleviate this lack of interpretability within a wide class of additive models without a major sacrifice in performance.
Subsequently, we present a global subscale model based on a classical mixture-of-experts framework.
Experiments on models augmented with these developments are conducted on credit modelling datasets and compared with the state of the art in interpretable modelling, incorporating both monotonicity and subscale modelling.

\textbf{Contributions.}
This work makes a number of technical contributions.
\emph{Linearised Additive Models (LAM).} We define a new class of models, that act as a `thin wrapper' around generalised additive models (GAMs), based on a piecewise linear approximation to the sigmoid function.
We demonstrate this approximation is the universally optimal one of this form in terms of squared error.
\emph{Subscale Probabilistic Mixture Models (SSPM).} We define a family of hierarchical models based on a linear combination of subscale models and provide an efficient training scheme for them which we call \textsc{SubscaleHedge}.
\emph{Experimental Comparison.} We conduct a statistically rigorous comparison on a collection of public credit modelling datasets between LAM-based algorithms, subscale mixture models and baselines from the interpretability literature, namely Additive Risk Models~\cite{CHEN2022113647} and XGBoost~\cite{xgboost}. 
We compare classification performance alongside calibration and find that LAMs and subscale mixture models have little negative impact on performance while providing interpretability and transparency benefits.

\textbf{Related Work.}
The closest work is the paper by~\cite{CHEN2022113647}, in which they define Additive Risk Models (ARMs).
These models are transparent and flexible, incorporating monotone constraints.
Moreover, they come in one- and two-layer varieties, with the latter being comprised of subscale models.
The inclusion of monotone constraints along with subscale modelling is unique in the literature as far as we are aware.
ARMs are described in greater detail in Section~\ref{sec:lin_prob_modelling}.
Indeed, the present work is inspired in part by this family of models, aiming to improve their transparency and interpretablility yet further while retaining the same classification performance.
We observe that comparing a subscale mixture of one-layer ARMs against the two-layer ARM, \emph{there is no statistically significant drop in classifier performance}, whereas the subscale mixture model affords better interpretability.
We systematically benchmark all relevant combinations of models against ARMs in Section~\ref{sec:experiments}.
Moreover, we are interested in evaluation of the ARM models' performance on a larger corpus of datasets (two datasets are considered in~\cite{CHEN2022113647}).

There is a vast literature on inherently interpretable models in general, a non-exhaustive group of which is~\cite{Sanjeeb2018, Caruana13,ExNN,vaughan2018explainable,kraus2019forecasting,de2021spline,pmlr-v119-vidal20a,NEURIPS2019_567b8f5f}.
None of these works incorporate both monotone constraints and explicit subscale modelling.

Our LAM construction is based on a piecewise-linear approximation to the sigmoid of specific form.
In the field of private machine learning, a polynomial approximation is often employed in encrypted logistic regression, since contemporary Fully Homomorphic Encryption schemes only allow for a restricted set of computations including polynomials.
For instance, in the work~\cite{Kim2019} they find the optimal least-squares Taylor polynomials of degrees $\{3, 7, 8\}$ approximating $\sigma(x)$ over the interval $[-8, 8]$.
Works such as~\cite{Chen2018,Kim2018b} use similar polynomial approximations. 

Probabilistic combinations of classifiers are a ubiquitous technique in machine learning, with various methods for obtaining the mixture coefficients.
For example, in stacking~\cite{WOLPERT1992241,Breiman1996} cross-validation and retraining are used to obtain weights for individual model.
A similar technique based on bootstrapping as opposed to cross-validation is presented in~\cite{LeBlanc1996}.
Expert Advice algorithms~\cite{Kivinen1994,ExpertAdvice1997,FREUND1997119} centre around iteratively updating weights with each incoming datum.
Random Forests~\cite{Breiman2001} work via a mixture of trees trained on different feature subsets, although these subsets are random as opposed to being in fixed groupings as in subscale modeling. 
An alternative approach to operating on groups of features independently is Hierarchical Bayesian Modelling~\cite{GyftodimosFlach2004}.
In the wider literature, there are many techniques for combining probability assignments or forecasts, dependent on the goals of the decision-maker, often taking into account the reliability of different experts~\cite{GenestZidek1986,doi:10.1287/deca.2014.0293,SATOPAA2014344}.
Our \textsc{SubscaleHedge} approach for combining subscale models in is most closely aligned to~\cite{FREUND1997119}, constituting a simple instantiation of the Hedge algorithm with the experts comprising individual models separately pre-trained on each subscale.
In the Explainability and Fairness literature, the work~\cite{FEAMOE} uses a mixture of experts approach to partition the feature space into regions acted on by different linear models, whereas the approach in the present work is to allow underlying models to be arbitrary, with each expert independently acting on a different feature group.

\textbf{Structure.} In the remainder of this section we fix notation and some useful definitions.
In Section~\ref{sec:lin_prob_modelling} we discuss additive modelling and formally define LAMs, showing optimality of approximation with respect to GAMs.
Section~\ref{sec:mixture_models} explores subscale modelling, introducing the \textsc{SubscaleHedge} training algorithm for linear opinion pools.
Experiments benchmarking the various introduced algorithms are in Section~\ref{sec:experiments}, followed by concluding statements in Section~\ref{sec:conclusion}.

\textbf{Notation.} We use $\{(\vb*{x}^{(j)}, y^{(j)})\}_{j=1}^M$ to denote the data, where $\vb*{x} \in \mathbb{R}^d$ is a vector of numerical features in a given dataset.
The labels are indicators of either defaulting on a loan, or bankruptcy: $y_i \in \{0 , 1\}$.
We assume data points are drawn i.i.d. 
For a point $\vb*{x} \in \mathbb{R}^d$, we denote the $i$\textsuperscript{th} element of $\vb*{x}$ by $x_i$.
The set $[d] := \{1, \ldots, d\}$ for $d \in \mathbb{N}$.
Individual data points are indexed in the superscript, so that datapoint $j$ is given by $\vb*{x}^{(j)}$ with $i$\textsuperscript{th} element denoted by $x_i^{(j)}$.
The $i$\textsuperscript{th} Euclidean basis vector is denoted by $\vb{e}_i$.
The sigmoid function is defined as $\sigma(x) := (1 + e^{-x})^{-1}$ for $x \in \mathbb{R}$.
We follow credit modelling terminology, where \emph{risk} $\hat{y}(\vb*{x}^{(j)})$ is a model's subjective probability in $[0, 1]$ for a data point $\vb*{x}^{(j)}$ to be of positive class, that is $y^{(j)} = 1$.

\textbf{Monotone Constraints.} The set of monotone increasing variables is denoted as $\mathcal{I} \subseteq [d]$, monotone decreasing variables $\mathcal{D} \subseteq [d]$ and the remaining variables $\mathcal{U} = [d] \setminus (\mathcal{I} \cup \mathcal{D})$.
Note that a variable can either be monotone increasing, decreasing, or neither, that is, $\mathcal{I}$, $\mathcal{D}$ and $\mathcal{U}$ form a partition of $[d]$.
Monotone constraints for a particular dataset are specified by the modeller using domain knowledge.
Categorical features are one-hot encoded and belong to $\mathcal{U}$.

\textbf{Subscales.} As discussed, it can be helpful for explanations~\cite{fico-score} to partition features into \emph{subscales}, namely semantically meaningful groups of features.
Formally, the subscales partition the feature set: a dataset with $d$ input features has every subscale $S_i$ satisfying $S_i \subseteq [d]$, $S_i \cap S_j = \emptyset$ for all $i \neq j$ and $\bigcup_i S_i = [d]$.
The set of all subscales we denote by $\mathcal{S}$ and $\mathcal{S}$ is decided by the modeller a priori.

\section{Logistic and Linear Probability Modelling}
\label{sec:lin_prob_modelling}

GAMs~\cite{hastie2009elements,molnar2022} are a widely-known and long standing class of models considered to be inherently interpretable due to their composite nature.
In this work we restrict attention to GAMs without feature interactions.
We formalise the notion of additive models as understood in this paper in Definition~\ref{def:logistic_additive_model}.

\begin{definition}[Logistic Additive Model]
\label{def:logistic_additive_model}
Let $\vb*{x} \in \mathbb{R}^d$. We call $\hat{y} : \mathbb{R}^d \to [0, 1]$ a \emph{logistic additive model} if takes the form
$\hat{y}(\vb*{x}) = \sigma(f(\vb*{x})) := \sigma ( \beta_0 + \sum_{i=1}^{d} \beta_i f_i(x_i) )$,
where the \emph{bias} $\beta_0 \in \mathbb{R}$ and for all $i \in [d]$, $\beta_i, x_i \in \mathbb{R}$ and the $f_i : \mathbb{R} \to \mathbb{R}$ are univariate functions.
We may also refer to $f(\vb*{x})$ as a logistic additive model with $\sigma \circ f$ implicit when the context is clear.
\end{definition}

We refer to logistic additive models as defined in Definition~\ref{def:logistic_additive_model} simply as additive models.
The simplest and most common additive model in wide usage is logistic regression, where $f_i(x_i) = x_i$ for all $i \in [d]$~\cite{hastie2009elements}.
We briefly describe a number of additive models from the literature that will be referred to throughout the text.

\textbf{Nonnegative Logistic Regression.}
We define \emph{Nonnegative Logistic Regression models (NNLR)} as those solving the following optimisation problem~\cite{CHEN2022113647}. 
\begin{equation}
\label{eq:nnlr_opt}
    \min_{\beta_i} \frac{1}{M}\sum_{j=1}^M L(\hat{y}(\vb*{x}^{(j)}), y^{(j)})) + C \sum_{i=1}^{d} \beta_i^2 \qq{subject to} \beta_i \geq 0 \ \forall  i \in \mathcal{I};\ \beta_i \leq 0 \ \forall i \in \mathcal{D}.  
\end{equation}
where the logistic loss $L(\hat{y}, y) = -y \log \hat{y} - (1 - y ) \log(1 - \hat{y})$ and 
$\hat{y}(\vb*{x}) = \sigma( \beta_0 + \sum_{i=1}^d \beta_i x_i )$, with $C > 0$ being a regularisation hyperparameter.
We solve the optimisation problem~\eqref{eq:nnlr_opt} via Sequential Least Squares Programming (SLSQP)~\cite{Kraft1994,kraft1988software} as implemented in \texttt{scipy}~\cite{2020SciPy-NMeth}. 
NNLR models follow the familiar $\ell_2$-regularised (or `ridge') logistic regression~\cite{hastie2009elements}, with the added constraint on the model coefficients enforcing the monotone directions on the variables in $\mathcal{I}$ and $\mathcal{D}$.
The ``nonnegative'' label arises from the constraints all being positive in the Additive Risk Models of~\cite{CHEN2022113647}.

\textbf{Additive Risk Models (ARMs).} ARMs are additive models introduced in~\cite{CHEN2022113647} and come in two flavours, \emph{One-Layer ARMs (ARM1)} and \emph{Two-Layer ARMs (ARM2)}.
We begin with a brief description of ARM1, and refer the reader to~\cite{CHEN2022113647} for a more detailed description.

The ARM1 model comprises a NNLR model with input features transformed like so: categorical features and special values are one-hot encoded.
For a continuous feature $u \in \mathcal{D}$, a series of $L_u \in \mathbb{N}$ indicator variables are created in its stead, signifying membership of half intervals $(-\infty, \theta_j]$ with right-side boundaries $\theta_1 < \theta_2 < \cdots < \theta_{L_u} = +\infty$.
The corresponding NNLR model coefficients $\beta_{u; \theta_j}$ for $j \in [L_u]$ are all constrained to be nonnegative, i.e. $x_{u; \theta_j} \in \mathcal{I}$.
This feature processing scheme enforces that the ARM1 model output is monotone decreasing in $x_u$ --  in the notation of Definition~\ref{def:logistic_additive_model}, $f_u(x_u)$ is piecewise constant and monotone decreasing.
Monotone increasing features receive a similar treatment, with the indicator variables corresponding to the half-intervals $[\theta_j, +\infty)$ with \emph{left-side} boundaries $\theta_1 > \theta_2 > \cdots > \theta_{L_u} = \phi_u$, where $\phi_u$ lower bounds $x_u$.
Again, $\beta_{u; \theta_j} \in \mathcal{I}$, ensuring that $f_u(x_u)$ is piecewise constant and monotone \emph{increasing}.
For unconstrained variables $u \in \mathcal{U}$, the intervals $(\theta_j, \theta_{j+1}]$ are \emph{two-sided}, with bin edges $\phi_u = \theta_1 < \theta_2 < \cdots < \theta_{L_u} = + \infty$.
The bin edges $\theta_j$ for all continuous variables are decided using an entropy-based scheme~\cite{CHEN2022113647} that is essentially equivalent to training a decision tree separately on $x_u$ with fixed number of leaves $L_u$ pre-determined as a hyperparameter.

The ARM2 model is built out of ARM1 models trained on solely the variables for a particular subscale, with the output of the subscale $S$ model designated by $r^{[S]}(\vb*{x})$.
The output for each subscale model $r^{[S]}$ lies within $[0, 1]$ and is interpreted as the risk arising solely from $S$.
The second layer of ARM2, $r(\vb*{x}) = \sigma(\beta_0 + \sum_{S \in \mathcal{S}} \beta_{S} r^{[S]}(\vb*{x}))$, is an additional NNLR model with the subscale risk scores as input variables.
The input variables are fixed to be monotone increasing, that is, $\beta_S \geq 0$.
Note that the individual subscale ARM1 models $r^{[S]}$ are trained first and their outputs treated as fixed for the training of the global model $r(\vb*{x})$.

\textbf{Logistic Modelling and Interpretability.} Historically, linear probability modelling, that is linear regression on dichotomous variables, was used prior to the advent of efficient methods for fitting logistic regression models~\cite{Aldrich1984,hastie2009elements}.
It is generally accepted that this application of linear regression to binary classification problems is unwise due to the propensity of the model returning probabilities outside the $[0, 1]$ interval and sensitivity to outliers~\cite{NgLogReg,hastie2009elements,molnar2022}.
In certain circumstances, these issues are not observed, with linear regression obtaining similar classification performance to logistic regression~\cite{Hellevik2009-cw,vonHippel}.
Nonetheless, the usage of logistic regression on classification problems is widespread.

To a modelling expert, the logistic regression model coefficients are interpreted as follows~\cite{molnar2022,hastie2009elements}: a unit change in variable $x_i$ leads to an increase in odds for the positive class of $\exp(\beta_i)$.
However, as observed by~\cite{harris,vonHippel} this interpretation can be unwieldy for experts and nigh-on impossible for non-experts to reason with when we are concerned with \emph{probabilities}, which is how the model outputs are typically presented.

\paragraph{Illustrative Example.} Suppose user A has a predicted risk of $\hat{y}(\vb*{x}^{(A)}) = 0.1$ and user B has a predicted risk of $\hat{y}(\vb*{x}^{(B)}) = 0.25$.
Users with risk $\geq 0.5$ are considered ``high-risk'' and ``low-risk'' otherwise.
Both users are interested in what happens to their risk if they increase the value of feature $x_i$ by one unit.
They are told the model coefficient $\beta_i = 1.61 \approx \ln 5$.
First, both users need to exponentiate $\beta_i$ which gives 5. 
The users now know that increasing feature $x_i$ by increases their odds by a factor of five.
Since they have their probability, they now have to compute what increasing their odds corresponds to in terms of probabilities.
In this instance for user A, $\operatorname{odds}(A) = \hat{y}(\vb*{x}^{(A)}) / (1 - \hat{y}(\vb*{x}^{(A)})) = 0.1 / (1 - 0.1) \approx 0.111$.
They then multiply their odds by 5 to $0.556$.
Converting back to probabilities we have $\hat{y}(\vb*{x}^{(A)} + \vb{e}_i) \approx 0.556 / (1 + 0.556) \approx 0.357$.
Subtracting their original risk score, we have that increasing $x_i$ by one unit increases the risk by a probability of $0.257$ and they remain low-risk.
A similarly laborious computation gives  an increase in risk for user B of approximately $0.625$. 
User B would be high-risk under a unit increase in $x_i$! This was not obvious before carrying out the computation explicitly. 
The nonlinearity of odds as a function of probabilities (and vice versa) means that  \emph{users with different risk scores cannot attribute logistic regression model outputs to the model coefficients in the same way.}
Moreover, the necessary computations are such that one cannot easily reason about the model's input-output relationship, even if the user is a seasoned practitioner in the use of logistic regression models.
On the contrary, the coefficients $\beta_i$ of a linear probability model admit the more direct interpretation of the increase in output model probability arising from a unit increase in $x_i$, regardless of the risk value of the user in question.
We now outline an approximation that provides the legibility of linear probability modelling with the precision of logistic modelling. 



\begin{figure}
    \centering
    \begin{subfigure}[b]{0.435\textwidth}
        \centering
        \resizebox{\columnwidth}{!}{%
\begin{tikzpicture}

\begin{axis}[
clip=false,
height=8cm,
legend cell align={left},
legend style={
  fill opacity=0.8,
  draw opacity=1,
  text opacity=1,
  at={(0.97,0.03)},
  anchor=south east,
  draw=white!80!black
},
tick align=outside,
tick pos=left,
width=12cm,
x grid style={white!69.0196078431373!black},
xlabel={\(\displaystyle \alpha\)},
xmin=-0.49895, xmax=10.49995,
xtick style={color=black},
y grid style={white!69.0196078431373!black},
ylabel={Squared Error},
ymin=-0.0311610017733549, ymax=0.872213469432495,
ytick style={color=black}
]
\addplot [semithick, black, forget plot]
table {%
0.001 0.385961111120247
0.0512462311557789 0.369431117718733
0.101492462311558 0.35332171273151
0.151738693467337 0.337632412652916
0.201984924623116 0.322362417802712
0.252231155778894 0.307510615191842
0.302477386934673 0.293075582492518
0.352723618090452 0.279055593079707
0.402969849246231 0.265448622102982
0.45321608040201 0.252252353539933
0.503462311557789 0.23946418817513
0.553708542713568 0.227081252442394
0.603954773869347 0.215100408062531
0.654201005025126 0.203518262404262
0.704447236180905 0.192331179492294
0.754693467336684 0.181535291583966
0.804939698492462 0.171126511234094
0.855185929648241 0.161100543766891
0.90543216080402 0.151452900074015
0.955678391959799 0.142178909658688
1.00592462311558 0.133273733847655
1.05617085427136 0.124732379095265
1.10641708542714 0.116549710306954
1.15666331658291 0.108720464113248
1.20690954773869 0.101239262029429
1.25715577889447 0.0941006234405752
1.30740201005025 0.0872989783565181
1.35764824120603 0.0808286798862533
1.40789447236181 0.0746840163865079
1.45814070351759 0.0688592232443939
1.50838693467337 0.0633484942592246
1.55863316582915 0.0581459925937176
1.60887939698492 0.0532458612697599
1.6591256281407 0.0486422331887337
1.70937185929648 0.0443292406609463
1.75961809045226 0.0403010244330563
1.80986432160804 0.0365517422064123
1.86011055276382 0.0330755766429967
1.9103567839196 0.0298667428590804
1.96060301507538 0.0269194954098286
2.01084924623116 0.0242281347709406
2.06109547738693 0.0217870133258369
2.11134170854271 0.0195905408691736
2.16158793969849 0.0176331896393115
2.21183417085427 0.0159094988939936
2.26208040201005 0.0144140790448647
2.31232663316583 0.0131416153674647
2.36257286432161 0.0120868713043507
2.41281909547739 0.0112446913794512
2.46306532663317 0.0106100037423955
2.51331155778894 0.0101778223616607
2.56355778894472 0.0099432488856377
2.6138040201005 0.00990147419054735
2.66405025125628 0.0100477796341037
2.71429648241206 0.0103775380334353
2.76454271356784 0.0108862143855143
2.81478894472362 0.0115693663478071
2.8650351758794 0.0124226444964397
2.91528140703518 0.0134417923785184
2.96552763819095 0.0146226463747096
3.01577386934673 0.0159611353874945
3.06602010050251 0.0174532803698515
3.11626633165829 0.0190951937084941
3.16651256281407 0.0208830784749807
3.21675879396985 0.022813227557485
3.26700502512563 0.0248820226851223
3.31725125628141 0.0270859333562607
3.36749748743719 0.0294215156813865
3.41774371859296 0.0318854111505416
3.46798994974874 0.0344743453346866
3.51823618090452 0.0371851265297126
3.5684824120603 0.0400146443512288
3.61872864321608 0.0429598682876753
3.66897487437186 0.0460178462187793
3.71922110552764 0.0491857029058264
3.76946733668342 0.0524606384596942
3.8197135678392 0.0558399267921794
3.86995979899498 0.0593209140556436
3.92020603015075 0.0629010170755928
3.97045226130653 0.0665777217804074
4.02069849246231 0.0703485816320635
4.07094472361809 0.0742112160613012
4.12119095477387 0.078163308910421
4.17143718592965 0.0822026068864899
4.22168341708543 0.0863269180275521
4.27192964824121 0.0905341101840744
4.32217587939699 0.0948221095176622
4.37242211055276 0.0991888990187999
4.42266834170854 0.103632517045242
4.47291457286432 0.108151055882353
4.5231608040201 0.112742660326602
4.57340703517588 0.11740552629324
4.62365326633166 0.122137899448977
4.67389949748744 0.126938073870403
4.72414572864322 0.131804390728659
4.774391959799 0.13673523700093
4.82463819095477 0.141729044208969
4.87488442211055 0.146784287185015
4.92513065326633 0.151899482865154
4.97537688442211 0.157073189110283
5.02562311557789 0.162304003554615
5.07586934673367 0.167590562481687
5.12611557788945 0.172931539727683
5.17636180904523 0.178325645611985
5.22660804020101 0.183771625894629
5.27685427135678 0.189268260760423
5.32710050251256 0.194814363829414
5.37734673366834 0.200408781193379
5.42759296482412 0.206050390477906
5.4778391959799 0.211738099929729
5.52808542713568 0.217470847528862
5.57833165829146 0.223247600125096
5.62857788944724 0.229067352598409
5.67882412060302 0.234929127042801
5.72907035175879 0.240831971973151
5.77931658291457 0.246774961554482
5.82956281407035 0.252757194853294
5.87980904522613 0.258777795110334
5.93005527638191 0.264835909034405
5.98030150753769 0.270930706116675
6.03054773869347 0.277061377964991
6.08079396984925 0.283227137657715
6.13104020100503 0.289427219116579
6.1812864321608 0.295660876498075
6.23153266331658 0.301927383602936
6.28177889447236 0.308226033303131
6.33202512562814 0.314556136986025
6.38227135678392 0.320917024015164
6.4325175879397 0.327308041207232
6.48276381909548 0.333728552324774
6.53301005025126 0.340177937584203
6.58325628140704 0.346655593178652
6.63350251256282 0.353160930815287
6.68374874371859 0.359693377266662
6.73399497487437 0.366252373935625
6.78424120603015 0.372837376433532
6.83448743718593 0.379447854171204
6.88473366834171 0.386083289962393
6.93497989949749 0.392743179639313
6.98522613065327 0.399427031679864
7.03547236180905 0.406134366846258
7.08571859296482 0.412864717834648
7.1359648241206 0.419617628935436
7.18621105527638 0.426392655703959
7.23645728643216 0.433189364641221
7.28670351758794 0.440007332884337
7.33694974874372 0.44684614790645
7.3871959798995 0.453705407225755
7.43744221105528 0.460584718123415
7.48768844221106 0.467483697370053
7.53793467336684 0.474401970960554
7.58818090452261 0.481339173856952
7.63842713567839 0.488294949739092
7.68867336683417 0.495268950762901
7.73891959798995 0.50226083732593
7.78916582914573 0.509270277840061
7.83941206030151 0.516296948511
7.88965829145729 0.523340533124548
7.93990452261307 0.530400722839193
7.99015075376885 0.537477215985039
8.04039698492462 0.544569717868755
8.0906432160804 0.551677940584365
8.14088944723618 0.558801602829747
8.19113567839196 0.565940429728569
8.24138190954774 0.573094152657641
8.29162814070352 0.580262509079308
8.3418743718593 0.587445242378907
8.39212060301508 0.594642101707063
8.44236683417085 0.601852841826589
8.49261306532663 0.609077222964019
8.54285929648241 0.616315010665449
8.59310552763819 0.623565975656665
8.64335175879397 0.63082989370742
8.69359798994975 0.638106545499603
8.74384422110553 0.645395716499385
8.79409045226131 0.652697196833043
8.84433668341708 0.660010781166385
8.89458291457287 0.667336268587768
8.94482914572864 0.674673462494414
8.99507537688442 0.682022170482114
9.0453216080402 0.689382204238046
9.09556783919598 0.696753379436749
9.14581407035176 0.704135515639025
9.19606030150754 0.711528436193806
9.24630653266332 0.718931968142774
9.2965527638191 0.726345942127767
9.34679899497488 0.733770192300752
9.39704522613065 0.741204556236409
9.44729145728643 0.748648874847188
9.49753768844221 0.756102992300733
9.54778391959799 0.763566755939664
9.59803015075377 0.771040016203622
9.64827638190955 0.778522626553433
9.69852261306533 0.786014443397507
9.74876884422111 0.793515326020191
9.79901507537688 0.801025136512151
9.84926130653266 0.80854373970268
9.89950753768844 0.816071003093885
9.94975376884422 0.823606796796696
10 0.831150993468593
};
\addplot [semithick, black, dashed]
table {%
2.59964289804225 -0.0311610017733549
2.59964289804225 0.872213469432495
};
\addlegendentry{$\alpha^\star = 2.5996...$}
\end{axis}

\end{tikzpicture}
        }
     \end{subfigure} 
     \begin{subfigure}[b]{0.555\textwidth}
        \centering
        \resizebox{\columnwidth}{!}{%
            \input{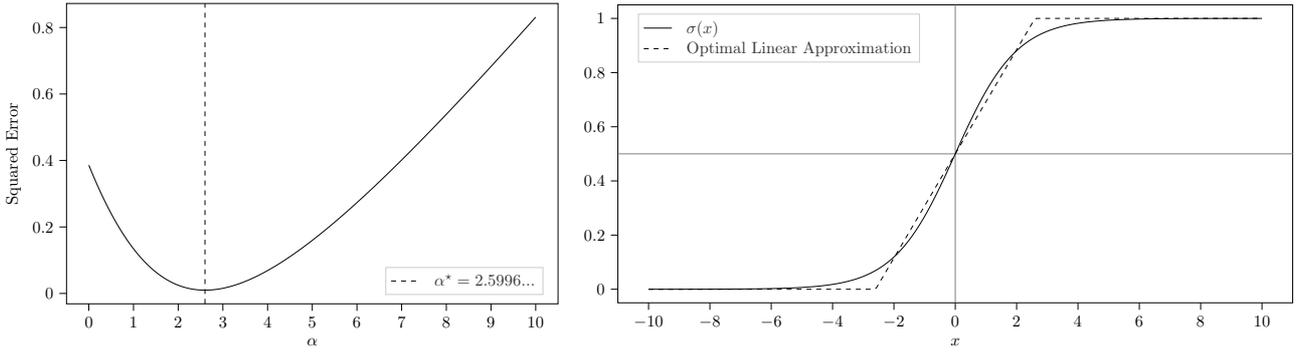}
        }
     \end{subfigure} 
    \caption{Squared error for linear approximation to sigmoid (left panel) and optimal (in square error) clipped linear approximation $\widetilde{\sigma}(x; \alpha \approx 2.5996)$ to sigmoid function $\sigma(x)$ (right panel).}
    \label{fig:opt_sig_approx}
\end{figure}

\paragraph{Approximating the Sigmoid.}
Consider an approximation $\widetilde{\sigma}(x; \alpha)$ to the sigmoid function $\sigma(x)$ parameterised by $\alpha > 0$ of the form
\begin{equation}
\label{eq:clip_lin_approx}
    \widetilde{\sigma}(x, \alpha) = 
    \begin{cases}
        0, & x \in (-\infty, -\alpha) ; \\
        \frac{1}{2} + \frac{x}{2\alpha}, & x \in [-\alpha, +\alpha] ; \\
        1, &  x \in (\alpha, +\infty) .
    \end{cases}
\end{equation}
In Fig.~\ref{fig:opt_sig_approx} (right) we see an example of $\widetilde{\sigma}(x, \alpha^\star)$ with the optimally chosen value for $\alpha^\star \approx 2.5996 \approx \frac{80000}{30773}$, which is optimal in terms of squared error.
More precisely, for a function $f(x)$ that is to be approximated by $\widetilde{f}(x)$, the \emph{squared error} is defined as
$\operatorname{SE} = \int_{-\infty}^{+\infty} (\widetilde{f}(x) - f(x))^2 \dd x$.
The squared error is finite when $\widetilde{f}(x) - f(x)$ is square integrable.
In the case of the approximation to $\sigma(x)$ we have that
\begin{align*}
    \operatorname{SE}(\alpha) &= \int_{-\infty}^{-\alpha} \qty(\sigma(x))^2 \, \dd x + \int_{-\alpha}^{+\alpha} \qty(\frac{1}{2} + \frac{x}{2\alpha} - \sigma(x))^2 \, \dd x + \int_{+\alpha}^{+\infty} \qty(1 - \sigma(x))^2 \, \dd x \\
    &= -\frac{7 \, {\alpha}^{2} + 6 \, {\alpha} \log {\left(1 + e^{{-\alpha}}\right)} - 6 \, {\alpha} \log\left(e^{{\alpha}} + 1\right) + 3 \, {\alpha} - 3 \, {\rm Li}_2\left(-e^{-{\alpha}}\right) + 3 \, {\rm Li}_2\left(-e^{{\alpha}}\right)}{3 \, {\alpha}},
\end{align*}
where ${\rm Li}_2$ is Spence's dilogarithm function, defined as ${\rm Li}_2(z) = -\int^z_0 \frac{\ln(1-u)}{u} \dd u$ for $z \in \mathbb{C}$.
The closed form of $\operatorname{SE}(\alpha)$ was obtained using the SageMath system~\cite{sagemath}. 
The function $\operatorname{SE}(\alpha) $ is plotted in Fig.~\ref{fig:opt_sig_approx} (left) for $\alpha > 0$, is convex and takes its minimum at $\alpha^\star \approx 2.5996 \approx \frac{80000}{30773}$, with this minimum obtained via Newton’s method.
We remark that the approximation $\widetilde{\sigma}(x, \alpha^\star)$ is \emph{universal}, in that for all models such that $\hat{y} = \sigma(f(\vb*{x}))$, we can simply make the substitution $\widetilde{\sigma}(f(\vb*{x}), \alpha^\star)$ with the same value for $\alpha^\star$.
From here on, we write $\widetilde{\sigma}(x) \equiv \widetilde{\sigma}(x, \alpha^\star)$ for brevity.

\paragraph{Linearised Additive Models (LAMs)}
Recall that an \emph{additive} model takes the form $\hat{y}(\vb*{x}) = \sigma(f(\vb*{x})) = \sigma ( \sum_{i=0}^{d} \beta_i f_i(x_i) )$,
where $\beta_i, x_i \in \mathbb{R}$ and $f_i : \mathbb{R} \to \mathbb{R}$ for all $i \in [d]$, $f_0: x \mapsto 1$.
Observe in the case of an additive model we have that when $f(\vb*{x}) \in [-\alpha, \alpha]$, we have
$\widetilde{\sigma}(f(\vb*{x})) = \widetilde{f}(\vb*{x}) := \frac{1}{2} + \sum_{i=0}^{d} \frac{\beta_i}{2 \alpha^\star} f_i (x_i)$, which follows immediately from linearity.
We shall call $\widetilde{f}$ the \emph{$\alpha^\star$-linearised model} relative to $f$ and shorten to \emph{linearised} model for brevity.
Let $\projunit$ be the projector from $\mathbb{R}$ onto the unit interval.
We then have the following result with proof in the appendix.
\begin{lemma}
\label{lem:add_model_equiv}
    Suppose $f$ is an additive model as defined in Definition~\ref{def:logistic_additive_model} with $\widetilde{f}$  its $\alpha^\star$-linearised version and $\widetilde{\sigma}$ as in~\eqref{eq:clip_lin_approx}.
    Then,
    $\widetilde{\sigma}(f(\vb*{x})) = \projunit(\widetilde{f}(\vb*{x}))$,
    that is, for an additive model $f$ the optimal clipped linear approximation to the sigmoid evaluated on the output of $f$ is formally equivalent to the output of the linearised model relative to $f$, projected to the unit interval.
\end{lemma}

Lemma~\ref{lem:add_model_equiv} inspires a family of models that may be derived from any additive model. 

\begin{definition}[Linearised Additive Models (LAM)]
\label{def:linearised_additive_models}
Let $\vb*{x} \in \mathbb{R}^{d}$ and let $f(\vb*{x})$ be an additive model per Definition~\ref{def:logistic_additive_model}.
Moreover, set $\alpha^\star := \frac{80000}{30773} \approx 2.5996$ as a universal constant.
Then, the \emph{linearised additive model}, $\hat{y}_{\text{lin}}$, relative to $f$ is given by
$$\hat{y}_{\text{lin}}(\vb*{x}) = \projunit\qty( \widetilde{f}(\vb*{x})) = \projunit\qty(\frac{1}{2} + \frac{\beta_0}{2\alpha^\star} + \sum_{i=1}^{d} \frac{\beta_i}{2 \alpha^\star} f_i(x_i)).$$
\end{definition}
\textbf{Example: Linearised Logistic Regression.}
For linear regression, we have $\hat{y}(\vb*{x}) = \sigma ( \sum_{i=0}^{d} \beta_i x_i)$, where we fix $x_0 = 1$ according to convention.
Then, the linearised version will be $\hat{y}_{\text{lin}}(\vb*{x}) = \projunit(\frac{1}{2} + \sum_{i=0}^{d} \frac{\beta_i}{2 \alpha^\star} x_i)$ for fixed $x_0 = 1$.
Inspection of linearised logistic regression and LAMs in general leads to the following considerations.
\emph{Computational Effort.} The linearised version of an additive model $f$ can leverage the existing well-optimised routines for training $f$. 
The linearisation component constitutes a thin wrapper around the existing model.
\emph{Interpretability.} The model coefficients are now directly interpretable in the same units of the model output, namely probability. Concretely, the coefficients $(\beta_i / 2\alpha^\star) \times f_i(x_i)$ are maximally faithful `true-to-the-model'~\cite{true_to_model_true_to_data} local feature attributions for the model $\hat{y}_{\text{lin}}$ evaluated on $\vb*{x}$, when the output of the original logistic model, $f(\vb*{x})$, falls in the interval $[-\alpha^\star, \alpha^\star]$, corresponding to probability interval $\approx[0.069, 0.931]$.

\begin{remark}
To train a LAM, all that is required is to train the underlying logistic additive model in the usual way, then apply Definition~\ref{def:linearised_additive_models} using the trained model coefficients $\{\beta_i\}_{i=0}^{d}$.
\end{remark}

In the case of linearised logistic regression we can interpret the coefficients $\beta_i / 2\alpha^\star$ as the contribution to model output in probability space for a unit increase in $x_i$.
As illustrated in the example earlier in the text, this direct interpretation incurs smaller cognitive overhead in answering certain questions.
Specifically, in the example users A and B had to interpret a unit increase in feature $x_i$ as having different and unintuitive effects on their risk score.
Using the LAM, a unit increase in $x_i$ gives rise to \emph{the same change in model output, regardless of the input}, i.e. $\frac{\beta_i}{2\alpha^\star} = \frac{1.61}{2 \times 2.5996}\approx 0.310$.

One must be cognisant of a number of caveats:
\emph{Model Certainty.} Output risk scores of $\leq 7\%$ and $\geq 93\%$ correspond to confident predictions.
The LAM is effectively rounding these risk scores to certainty with respect to the logistic additive model.
Whether this makes sense is down to the application.
For example, if a difference between 97\% and 99.99\% is important or between 5\% and 0.05\%, then clipping to $[0, 1]$ may not be appropriate. 
If one is uncomfortable with outputs in $\{0, 1\}$, one can either: \emph{i.} clip to the interval $[\epsilon, 1 - \epsilon]$ for some small $\epsilon > 0$; or \emph{ii.} increase $\alpha$ in the clipped linear approximation (thereby growing the set of model inputs with output in $(0, 1)$) incurring a penalty in approximation.
We delegate further investigation of the consequences of these interventions to future work.
\emph{Interpretation of model coefficients under model certainty.} Supposing the model output lies in $\{0, 1\}$, our stated interpretation of $\beta_i / 2\alpha^\star$ requires care, since there will be many $\Delta \vb*{x}$ such that the model output does not change at all.
Given that LAMs modify the probability assignments given by the underlying additive models, we are particularly interested in the effect this may have on classifier calibration, which we investigate in Section~\ref{sec:experiments}. 

\section{Subscale Probabilistic Mixture Models.}

\label{sec:mixture_models}

We turn now to subscale modelling, as exemplified in the two-layer ARM defined in~\cite{CHEN2022113647}. 
We present a novel class of subscale models, \emph{Subscale Probabilistic Mixture Models (SSPM)}, that we argue is more interpretable than two-layer ARMs, owing to comprising a linear combination as opposed to a logistic combination (see the arguments in Section~\ref{sec:lin_prob_modelling}). Moreover, we show in Section~\ref{sec:experiments} that there is no statistically significant difference in classification performance.

First, we outline the need for subscale modelling when interpretability is required.
Suppose we have a model that takes as input 250 features split into 10 subscales. Perhaps one subscale contains 24 features corresponding to $\left\{\text{Count of events of type X in past $n$ hours} \mid n \in \{1, \ldots, 24\}  \right\}$ grouped together in the subscale EventsTypeX. Using the subscale attribution EventsTypeX tells us broadly how important these events are to the model on a given input with respect to, say EventsTypeY. 
This may not be clear from the individual feature contributions -- even when underlying subscale models are themselves interpretable -- for example when many of the individual feature attributions take similar values. 
Importantly, we will see that the linear opinion pool structure of SSPMs ensures that the subscale attributions are completely faithful to the action of the model, as may not be the case by just adding up individual feature contributions in a non-hierarchical model.

As an alternative to combining subscale models via NNLR as in ARM2 models~\cite{CHEN2022113647}, we propose to combine the subscale risk scores using probabilistic weighting.
Namely, for each subscale $S\in\mathcal{S}$ we have a weight $w_S$ and combine the risk factors in a global model $r(\vb*{x})$ linearly, that is $r(\vb*{x}) = \sum_{S \in \mathcal{S}} w_S r^{[S]}(\vb*{x})$, where $w_S \in [0, 1]$ for all $S \in \mathcal{S}$ and $\sum_{S \in \mathcal{S}} w_S = 1$.
We will refer to such models that are trained using \textsc{SubscaleHedge} (Algorithm~\ref{alg:subscale_hedge}) as SSPMs, or \emph{mixture models} for short. 
The probabilistic sum is also known as a \emph{linear opinion pool}~\cite{GenestZidek1986}. 
If one wants to provide an local subscale attribution of the risk score $r$ for a mixture model evaluated on datapoint $\vb*{x}$ , one simply returns $w_S r^{[S]}(\vb*{x})$, obviating the need for a conversion between log-odds and probabilities.

As discussed in Section~\ref{sec:intro}, there are many known ways to combine model outputs.
For example, the technique of \emph{stacking}~\cite{WOLPERT1992241,Breiman1996} uses cross-validation data and least squares under non-negativity constraints to to determine the coefficients of a linear combination of different models.
Preliminary experiments we conducted determined this method was both computationally costly and would often lead to many of the $w_S$ being zero, resulting in poor generalisation performance.
Instead we adapt a simple classical family of algorithms from the online learning community, \emph{Hedge}~\cite{FREUND1997119} or \emph{Multiplicative Weights} to find the weights.
The simplicity of this algorithm lends additional transparency for model consumers.

\begin{algorithm}
\DontPrintSemicolon
\caption{\textsc{SubscaleHedge}}
\label{alg:subscale_hedge}
\KwData{Training data $\mathcal{D} = {(\vb*{x}^{(j)}, y^{(j)})}^M_{j=1}$, Subscale set $\mathcal{S}$, Trained models for each subscale $\{r^{[S]}(\vb*{x})\}_{S \in \mathcal{S}}$, Loss function $L$ (logistic loss)}
\SetKwProg{subscalehedge}{SubscaleHedge}{}{end}
\subscalehedge{$(\mathcal{D}, \mathcal{S}, \{r^{[S]}(\vb*{x})\}_{S \in \mathcal{S}}, M, L)$}{
    Set $\eta \leftarrow 8\log(\abs{S} / M)$ \tcp*{learning rate for mixture}
    Set $w_S^{(0)} \leftarrow 1 / \abs{\mathcal{S}}$ for all $S \in \mathcal{S}$\\
    Randomly shuffle $\mathcal{D}$\\
    \For{$t = 0$ \KwTo $M - 1$}
        {
        $w_S^{(t+1)} \leftarrow \exp(-\eta L\qty(r^{[S]}(\vb*{x}^{(t)}, y^{(t)})))w_S^{(t)}$ for all $S \in \mathcal{S}$ \tcp*{update expert weights}
        $w_S^{(t+1)} \leftarrow \frac{w_S^{(t+1)}}{\sum_{S' \in \mathcal{S}} w_{S'}^{(t+1)}}$ for all $S \in \mathcal{S}$  \tcp*{renormalise expert weights}
        }
        $w_S \leftarrow w_S^{(M)}$ for all $S \in \mathcal{S}$ \tcp*{finalise weights for each subscale}
    \Return $\{w_S\}_{S \in \mathcal{S}}$ 
}
\end{algorithm}

We describe \textsc{SubscaleHedge} in Algorithm~\ref{alg:subscale_hedge}.
Each subscale model $r^{[S]}$ is assumed to be pretrained on the $M$ training points.
Every subscale model is now considered as one of $\abs{\mathcal{S}}$ experts, with each `opinion' for a given training point $\vb*{x}^{(j)}$ given as $r^{[S]}(\vb*{x}^{(j)}) \in [0, 1]$.
The starting weight for all subscales is $w_S = 1/\abs{\mathcal{S}}$ for all $S \in \mathcal{S}$.
We then randomly shuffle the $M$ training points and each training point is fed to the $r^{[S]}$ in sequential fashion without replacement until the training set is exhausted.
The weights are updated after each data point according to the Hedge algorithm so as to ``reward'' more accurate experts and ``punish'' the less accurate experts, multiplying each $w_S$ by $\exp(-\eta L(r^{[S]}(\vb*{x}^{(j)}), y^{(j)}))$, where the learning rate $\eta$ takes the recommended value $\eta = 8 \log(\abs{\mathcal{S}}) / M$ and $L$ is the logistic loss.
Before the next data point the weights are re-normalised such that $\sum_{S \in \mathcal{S}} w_S = 1$.
Note that the time complexity of \textsc{SubscaleHedge} scales as $O(M \abs{\mathcal{S}} \tau)$, where $\tau$ is the subscale model evaluation time, owing to the evaluation of each of the $\abs{\mathcal{S}}$ subscale models on each of the $M$ training points.
%
%
Observe that we can define an SSPM with an arbitrary model architecture for each subscale, as long as $r^{[S]}(\vb*{x}) \in [0, 1]$ for every subscale $S \in \mathcal{S}$.
In the subsequent text, for an SSPM where the subscale models $r^{[S]}(\vb*{x})$ are of type X, we denote the overall mixture model $\sum_{S \in \mathcal{S}} w_S r^{[S]}(\vb*{x})$ as MixX.

\section{Experimental Comparison}

\label{sec:experiments}

We are chiefly interested to what extent linearising relative to logistic models, and/or partitioning features into subscales and using a probabilistic mixture introduce degredation (if any) of both classification performance and calibration performance.
For each metric the 10-fold stratified cross-validation score is computed for every (classifier, dataset) combination. 

\textbf{Models.}
We compare all models to XGBoost~\cite{CHEN2022113647}, with shorthand XGB, and XGB with monotone constraints imposed (MonoXGB).
XGB classification performance serves as an effective upper-bound on the other models.
Indeed, there has been much discussion in recent years about a tradeoff between model accuracy and interpretability~\cite{Rudin19,Dziugaite2020} and XGB is included here to illustrate this tradeoff.
Further baselines include NNLR with raw feature inputs as an effective lower bound on model performance, alongside ARM1 and ARM2, as the ARM algorithms represent the state-of-the-art in interpretable modelling in terms of respecting monotonicity and explicit subscale modelling (ARM2).
For an additive model with shorthand X, we denote its linearised version (in the sense of Definition~\ref{def:linearised_additive_models}) by LinX.
We include the linearised models LinNNLR, LinARM1 and LinARM2 in our experiments.
LinARM2 has both the individual subscale models and global NNLR model linearised.
In terms of the mixture models as defined Section~\ref{sec:mixture_models}, we test four such models: MixARM1 and MixLinARM1, corresponding to a mixture of ARM1 subscale models and linearised ARM1 subscale models respectively; alongside MixMonoXGB and MixXGB, corresponding to a mixture of XGBoost~\cite{xgboost} subscale models with and without monotone constraints imposed.
Table~\ref{tab:classifier_desc} provides a quick lookup table of the $k=12$ models.

\begin{table}[h!]
\centering
        \caption{Shorthand and descriptions of $k=12$ classifiers under comparison. 
        Models in the second section have two layers, with first layer operating on subscales and the second layer combining the individual subscale scores.
        }
        \label{tab:classifier_desc}
        {\small
        \begin{tabular}{lc}
\toprule
\textbf{Classifier} & \textbf{Description} \\
\midrule
NNLR & Non-negative logistic regression on unprocessed features. \\ 
LinNNLR & Linearised model relative to NNLR. \\ 
ARM1 & One-Layer Additive Risk Model from~\cite{CHEN2022113647}. \\ 
LinARM1 & Linearised model relative to ARM1. \\
XGB & XGBoost~\cite{xgboost} model. \\
MonoXGB & XGBoost~\cite{xgboost} model with monotone constraints imposed. \\
\midrule
MixARM1 & Probabilistic mixture of ARM1 models fitted to each subscale. \\
MixLinARM1 & Probabilistic mixture of LinARM1 models fitted to each subscale. \\
ARM2 & Two-Layer Additive Risk Model from~\cite{CHEN2022113647}. \\
LinARM2 & Linearised model relative to ARM2. Both logistic layers are linearised. \\
MixXGB & Probabilistic mixture of XGB models fitted to each subscale. \\
MixMonoXGB & Probabilistic mixture of MonoXGB models fitted to each subscale.  \\
\bottomrule
\end{tabular}
        }
\end{table}
Model hyperparameters are provided in the Appendix and no attempts were made to tune them due to computational constraints.

\textbf{Datasets.}
In this work we are principally interested in the consumer credit domain, as typically there are monotone constraints and natural feature groupings that one can use as subscales.
Bankruptcy prediction datasets are also included due to their similar problem structure and origin. 
We consider publically available datasets from the UCI repository~\cite{UCI}, namely, the German Credit dataset, Australia credit approvals~\cite{QUINLAN1987221}, Taiwanese bankruptcy~\cite{LIANG2016561} prediction, Japanese credit screening and the Polish companies bankruptcy~\cite{zikeba2016ensemble} dataset.
Moreover, we consider the FICO Home Equity Line of Credit dataset~(HELOC)~\cite{heloc}, Give Me Some Credit~(GMSC) and Lending Club~(LC)~\cite{lending-club} datasets.
If we wish to distinguish metrics of interest of the $k=12$ different algorithms under consideration, then it is important for the number of statistically independent datasets, $N$, to be as large as possible so that the power of any statistical tests is maximal.
The work~\cite{Campelo2020} provides an algorithm for determining the appropriate $N$ for a desired statistical power.
We restrict our experiments to publically available datasets that are not synthetically generated, so we cannot keep generating datasets until the desired $N$ is reached.
Nonetheless, we are able to use $N=24$ independent datasets.
For the Poland dataset, there are 5 separate datasets provided, corresponding to consecutive years of data.
We consider these datasets separately, and denote them by $\text{Poland}\_n$ for $n\in\{ 0, \ldots, 4\}$.
The LC dataset is an order of magnitude larger than the others, thus we split the data into temporally contiguous (and disjoint) regions comprising 100000 datapoints each (apart from the last region).
This procedure leaves us with 13 datasets, which we denote by $\text{LC}\_n$ for $n\in\{ 0, \ldots, 12\}$.
The subscales $\mathcal{S}$ and monotone constraints $(\mathcal{I}, \mathcal{D}, \mathcal{U})$ for the remaining datasets were decided based on domain knowledge and are included in full in the appendix.

\textbf{Classification Performance.}
To measure of classification performance, we use the area under the curve of the receiver operating characteristic~\cite{BRADLEY19971145,Hanley1983}, denoted as AUC.

\textbf{Calibration.}
Generally, a set of predictions of a binary outcome is well calibrated if the outcomes predicted to occur with probability $p$ occur about $p$ fraction of the time, for any probability $p \in [0, 1]$.
A common method for assessing the calibration of a binary classifier is the \emph{reliability diagram}~\cite{DeGroot1983,Niculescu_Mizil2005}, wherein the empirical frequency of the positive class is plotted against the predicted positive class probability by the model.
In this work, we will consider two widely-used numerical summary statistics for the reliability diagram, \emph{Expected Calibration Error (ECE)} and \emph{Maximum Calibration Error, (MCE)}.
To compute these metrics, the test set predictions are sorted and partitioned into $K$ equally spaced bins over $[0, 1]$ ($K = 15$ in our experiments). 
We then have
$\operatorname{ECE} = \sum^K_{i=1} P(i) \cdot \abs{o_i - e_i}$ and $\operatorname{MCE} =\max_{i\in\{1, \ldots, K\}} \abs{o_i - e_i}$,
where $o_i$ is the true fraction of positive instances in bin $i$, $e_i$ is the arithmetic mean of the model outputs for the instances in bin $i$, and $P(i)$ is the empirical probability (fraction) of all instances that fall into bin $i$.
Lower values of ECE and MCE correspond to better calibration of a particular model, with the idealised model having a value of zero for both.

\begin{figure}[!hbtp]
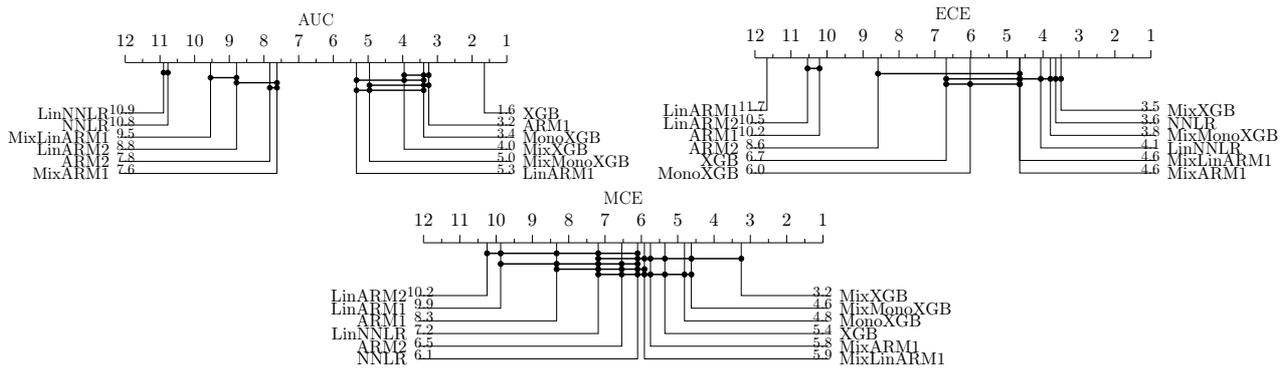

     \centering
     \begin{subfigure}[b]{0.49\textwidth}
        \centering
        \resizebox{\columnwidth}{!}{%
            \input{auc}
        }
     \end{subfigure} 
     \begin{subfigure}[b]{0.49\textwidth}
        \centering
        \resizebox{\columnwidth}{!}{%
\begin{tikzpicture}

\begin{axis}[
clip=false,
height=6cm,
hide x axis,
hide y axis,
tick align=outside,
tick pos=left,
title style={yshift=-0.2cm,font=\Large},
title={ECE},
width=16cm,
x grid style={white!69.0196078431373!black},
xmin=0, xmax=1,
xtick style={color=black},
y dir=reverse,
y grid style={white!69.0196078431373!black},
ymin=0, ymax=1,
ytick style={color=black}
]
\addplot [draw=black, fill=black, mark=*, only marks]
table{%
x  y
0.213409090909091 0.288135593220339
0.237045454545455 0.288135593220339
};
\addplot [draw=black, fill=black, mark=*, only marks]
table{%
x  y
0.631477272727273 0.322033898305085
0.352272727272727 0.322033898305085
};
\addplot [draw=black, fill=black, mark=*, only marks]
table{%
x  y
0.631477272727273 0.355932203389831
0.486704545454546 0.355932203389831
0.631477272727273 0.355932203389831
0.672840909090909 0.355932203389831
0.692045454545455 0.355932203389831
0.702386363636364 0.355932203389831
0.712727272727273 0.355932203389831
};
\addplot [draw=black, fill=black, mark=*, only marks]
table{%
x  y
0.631477272727273 0.389830508474576
0.486704545454546 0.389830508474576
0.631477272727273 0.389830508474576
0.533977272727273 0.389830508474576
};
\addplot [semithick, white]
table {%
0 0
1 1
};
\addplot [thick, black]
table {%
0.11 0.220338983050847
0.89 0.220338983050847
};
\addplot [thick, black]
table {%
0.89 0.169491525423729
0.89 0.220338983050847
};
\addplot [thick, black]
table {%
0.854545454545454 0.194915254237288
0.854545454545454 0.220338983050847
};
\addplot [thick, black]
table {%
0.819090909090909 0.169491525423729
0.819090909090909 0.220338983050847
};
\addplot [thick, black]
table {%
0.783636363636364 0.194915254237288
0.783636363636364 0.220338983050847
};
\addplot [thick, black]
table {%
0.748181818181818 0.169491525423729
0.748181818181818 0.220338983050847
};
\addplot [thick, black]
table {%
0.712727272727273 0.194915254237288
0.712727272727273 0.220338983050847
};
\addplot [thick, black]
table {%
0.677272727272727 0.169491525423729
0.677272727272727 0.220338983050847
};
\addplot [thick, black]
table {%
0.641818181818182 0.194915254237288
0.641818181818182 0.220338983050847
};
\addplot [thick, black]
table {%
0.606363636363636 0.169491525423729
0.606363636363636 0.220338983050847
};
\addplot [thick, black]
table {%
0.570909090909091 0.194915254237288
0.570909090909091 0.220338983050847
};
\addplot [thick, black]
table {%
0.535454545454546 0.169491525423729
0.535454545454546 0.220338983050847
};
\addplot [thick, black]
table {%
0.5 0.194915254237288
0.5 0.220338983050847
};
\addplot [thick, black]
table {%
0.464545454545455 0.169491525423729
0.464545454545455 0.220338983050847
};
\addplot [thick, black]
table {%
0.429090909090909 0.194915254237288
0.429090909090909 0.220338983050847
};
\addplot [thick, black]
table {%
0.393636363636364 0.169491525423729
0.393636363636364 0.220338983050847
};
\addplot [thick, black]
table {%
0.358181818181818 0.194915254237288
0.358181818181818 0.220338983050847
};
\addplot [thick, black]
table {%
0.322727272727273 0.169491525423729
0.322727272727273 0.220338983050847
};
\addplot [thick, black]
table {%
0.287272727272727 0.194915254237288
0.287272727272727 0.220338983050847
};
\addplot [thick, black]
table {%
0.251818181818182 0.169491525423729
0.251818181818182 0.220338983050847
};
\addplot [thick, black]
table {%
0.216363636363636 0.194915254237288
0.216363636363636 0.220338983050847
};
\addplot [thick, black]
table {%
0.180909090909091 0.169491525423729
0.180909090909091 0.220338983050847
};
\addplot [thick, black]
table {%
0.145454545454545 0.194915254237288
0.145454545454545 0.220338983050847
};
\addplot [thick, black]
table {%
0.11 0.169491525423729
0.11 0.220338983050847
};
\addplot [thick, black]
table {%
0.133636363636364 0.220338983050847
0.133636363636364 0.559322033898305
0.1 0.559322033898305
};
\addplot [thick, black]
table {%
0.213409090909091 0.220338983050847
0.213409090909091 0.640677966101695
0.1 0.640677966101695
};
\addplot [thick, black]
table {%
0.237045454545455 0.220338983050847
0.237045454545455 0.722033898305085
0.1 0.722033898305085
};
\addplot [thick, black]
table {%
0.352272727272727 0.220338983050847
0.352272727272727 0.803389830508475
0.1 0.803389830508475
};
\addplot [thick, black]
table {%
0.486704545454546 0.220338983050847
0.486704545454546 0.884745762711864
0.1 0.884745762711864
};
\addplot [thick, black]
table {%
0.533977272727273 0.220338983050847
0.533977272727273 0.966101694915254
0.1 0.966101694915254
};
\addplot [thick, black]
table {%
0.631477272727273 0.220338983050847
0.631477272727273 0.966101694915254
0.9 0.966101694915254
};
\addplot [thick, black]
table {%
0.631477272727273 0.220338983050847
0.631477272727273 0.884745762711864
0.9 0.884745762711864
};
\addplot [thick, black]
table {%
0.672840909090909 0.220338983050847
0.672840909090909 0.803389830508475
0.9 0.803389830508475
};
\addplot [thick, black]
table {%
0.692045454545455 0.220338983050847
0.692045454545455 0.722033898305085
0.9 0.722033898305085
};
\addplot [thick, black]
table {%
0.702386363636364 0.220338983050847
0.702386363636364 0.640677966101695
0.9 0.640677966101695
};
\addplot [thick, black]
table {%
0.712727272727273 0.220338983050847
0.712727272727273 0.559322033898305
0.9 0.559322033898305
};
\addplot [very thick, black]
table {%
0.215409090909091 0.288135593220339
0.235045454545455 0.288135593220339
};
\addplot [very thick, black]
table {%
0.354272727272727 0.322033898305085
0.629477272727273 0.322033898305085
};
\addplot [very thick, black]
table {%
0.488704545454545 0.355932203389831
0.710727272727273 0.355932203389831
};
\addplot [very thick, black]
table {%
0.488704545454545 0.389830508474576
0.629477272727273 0.389830508474576
};
\draw (axis cs:0.89,0.152542372881356) node[
  scale=1.5,
  anchor=south,
  text=black,
  rotate=0.0
]{1};
\draw (axis cs:0.819090909090909,0.152542372881356) node[
  scale=1.5,
  anchor=south,
  text=black,
  rotate=0.0
]{2};
\draw (axis cs:0.748181818181818,0.152542372881356) node[
  scale=1.5,
  anchor=south,
  text=black,
  rotate=0.0
]{3};
\draw (axis cs:0.677272727272727,0.152542372881356) node[
  scale=1.5,
  anchor=south,
  text=black,
  rotate=0.0
]{4};
\draw (axis cs:0.606363636363636,0.152542372881356) node[
  scale=1.5,
  anchor=south,
  text=black,
  rotate=0.0
]{5};
\draw (axis cs:0.535454545454546,0.152542372881356) node[
  scale=1.5,
  anchor=south,
  text=black,
  rotate=0.0
]{6};
\draw (axis cs:0.464545454545455,0.152542372881356) node[
  scale=1.5,
  anchor=south,
  text=black,
  rotate=0.0
]{7};
\draw (axis cs:0.393636363636364,0.152542372881356) node[
  scale=1.5,
  anchor=south,
  text=black,
  rotate=0.0
]{8};
\draw (axis cs:0.322727272727273,0.152542372881356) node[
  scale=1.5,
  anchor=south,
  text=black,
  rotate=0.0
]{9};
\draw (axis cs:0.251818181818182,0.152542372881356) node[
  scale=1.5,
  anchor=south,
  text=black,
  rotate=0.0
]{10};
\draw (axis cs:0.180909090909091,0.152542372881356) node[
  scale=1.5,
  anchor=south,
  text=black,
  rotate=0.0
]{11};
\draw (axis cs:0.11,0.152542372881356) node[
  scale=1.5,
  anchor=south,
  text=black,
  rotate=0.0
]{12};
\draw (axis cs:0.14,0.533898305084746) node[
  scale=1.2,
  anchor=east,
  text=black,
  rotate=0.0
]{11.7};
\draw (axis cs:0.09,0.559322033898305) node[
  scale=1.4,
  anchor=east,
  text=black,
  rotate=0.0
]{LinARM1};
\draw (axis cs:0.14,0.615254237288136) node[
  scale=1.2,
  anchor=east,
  text=black,
  rotate=0.0
]{10.5};
\draw (axis cs:0.09,0.640677966101695) node[
  scale=1.4,
  anchor=east,
  text=black,
  rotate=0.0
]{LinARM2};
\draw (axis cs:0.14,0.696610169491525) node[
  scale=1.2,
  anchor=east,
  text=black,
  rotate=0.0
]{10.2};
\draw (axis cs:0.09,0.722033898305085) node[
  scale=1.4,
  anchor=east,
  text=black,
  rotate=0.0
]{ARM1};
\draw (axis cs:0.14,0.777966101694915) node[
  scale=1.2,
  anchor=east,
  text=black,
  rotate=0.0
]{8.6};
\draw (axis cs:0.09,0.803389830508475) node[
  scale=1.4,
  anchor=east,
  text=black,
  rotate=0.0
]{ARM2};
\draw (axis cs:0.14,0.859322033898305) node[
  scale=1.2,
  anchor=east,
  text=black,
  rotate=0.0
]{6.7};
\draw (axis cs:0.09,0.884745762711864) node[
  scale=1.4,
  anchor=east,
  text=black,
  rotate=0.0
]{XGB};
\draw (axis cs:0.14,0.940677966101695) node[
  scale=1.2,
  anchor=east,
  text=black,
  rotate=0.0
]{6.0};
\draw (axis cs:0.09,0.966101694915254) node[
  scale=1.4,
  anchor=east,
  text=black,
  rotate=0.0
]{MonoXGB};
\draw (axis cs:0.86,0.940677966101695) node[
  scale=1.2,
  anchor=west,
  text=black,
  rotate=0.0
]{4.6};
\draw (axis cs:0.91,0.966101694915254) node[
  scale=1.4,
  anchor=west,
  text=black,
  rotate=0.0
]{MixARM1};
\draw (axis cs:0.86,0.859322033898305) node[
  scale=1.2,
  anchor=west,
  text=black,
  rotate=0.0
]{4.6};
\draw (axis cs:0.91,0.884745762711864) node[
  scale=1.4,
  anchor=west,
  text=black,
  rotate=0.0
]{MixLinARM1};
\draw (axis cs:0.86,0.777966101694915) node[
  scale=1.2,
  anchor=west,
  text=black,
  rotate=0.0
]{4.1};
\draw (axis cs:0.91,0.803389830508475) node[
  scale=1.4,
  anchor=west,
  text=black,
  rotate=0.0
]{LinNNLR};
\draw (axis cs:0.86,0.696610169491525) node[
  scale=1.2,
  anchor=west,
  text=black,
  rotate=0.0
]{3.8};
\draw (axis cs:0.91,0.722033898305085) node[
  scale=1.4,
  anchor=west,
  text=black,
  rotate=0.0
]{MixMonoXGB};
\draw (axis cs:0.86,0.615254237288136) node[
  scale=1.2,
  anchor=west,
  text=black,
  rotate=0.0
]{3.6};
\draw (axis cs:0.91,0.640677966101695) node[
  scale=1.4,
  anchor=west,
  text=black,
  rotate=0.0
]{NNLR};
\draw (axis cs:0.86,0.533898305084746) node[
  scale=1.2,
  anchor=west,
  text=black,
  rotate=0.0
]{3.5};
\draw (axis cs:0.91,0.559322033898305) node[
  scale=1.4,
  anchor=west,
  text=black,
  rotate=0.0
]{MixXGB};
\end{axis}

\end{tikzpicture}
        }
     \end{subfigure} \\
     \begin{subfigure}[b]{0.49\textwidth}
        \centering
        \hspace{-0.02\columnwidth}
        \resizebox{\columnwidth}{!}{%
            \input{mce}
        }
     \end{subfigure}
        \caption{Critical Difference diagrams for AUC, ECE and MCE. The $x$-axis represents the mean rank averaged over the $N=24$ datasets, with each classifier's mean rank reported adjacent to its name. Classifiers connected by an edge \emph{cannot} be distinguished with significance $\alpha=0.05$.}
        \label{fig:three graphs}
\end{figure}

\textbf{Statistical Methodology.}
We follow the advice of~\citet{Demsar2006} and first conduct the Friedman omnibus test~\cite{Friedman1940} with Iman-Davenport correction~\cite{Iman1980}.
Having rejected the null hypothesis that the ranks of each of the algorithms are identical, we conduct the pairwise post-hoc analysis recommended by~\cite{JMLR:v17:benavoli16a}, whereby a Wilcoxon signed-rank test~\cite{Wilcoxon1945} is conducted with Holm’s alpha correction~\cite{Holm1979,JMLR:v9:garcia08a} to control the family-wise error rate.
We provide details of these computations in Section~\ref{sec:stats} in the Appendix.
Consider a graph where for each algorithm $i$ we draw a node. 
We draw an edge between any nodes corresponding to algorithm pairs $(i, i')$ such $i$ cannot be distinguished from $i'$ with significance $\alpha=0.05$.
We display this graph alongside the average ranks $R_i$ using the Critical Difference (CD) diagrams of~\cite{Demsar2006} for AUC, ECE and MCE in Figure~\ref{fig:three graphs}.
Not only are we interested in whether an observed difference in cross validated score between two algorithms is statistically significant, but also the size of this difference.
The Wilcoxon signed-rank test has an associated \emph{Hodges-Lehmann estimator}~\cite{WILCOX202245}, namely, a point estimate of this observed difference across the $k$ datasets.
This estimator is the \emph{psuedomedian},  $\hat{\theta}_{\text{HL}}$, and is defined as
$\hat{\theta}_{\text{HL}} = \operatorname{median}_{1 \leq i \leq j \leq N} \{ \frac{d_i + d_j}{2} \}$,
namely the median of the pairwise \emph{Walsh averages} $\frac{d_i + d_j}{2}$.
The quantities $d_i$ and $d_j$ correspond to the difference between a fixed pair of algorithms in performance on the $i$\textsuperscript{th} and $j$\textsuperscript{th} of $N$ datasets respectively.
The pseudomedian constitutes a robust estimate of the difference in score between a pair of algorithms.
Care must be taken~\cite{Demsar2006} since we are implicitly assuming that the pairwise score differences between two algorithms are commensurable across datasets.
Nonetheless we believe these point estimates are useful to report for the reader to have an idea of the scales involved.
In Tables~\ref{tab:comparison_auc},~\ref{tab:comparison_ece} (Appendix)~and~\ref{tab:comparison_mce}~(Appendix) we tabulate the differences $\hat{\theta}_{\text{HL}}$ between all pairings of the $k=12$ algorithms for the AUC, ECE and MCE metrics respectively.

\textbf{Results for Classification Performance.}
Raw AUC scores for each classifier, dataset combination are reported in Table~\ref{tab:auc_raw} in the Appendix.
From the CD diagram for AUC (Figure~\ref{fig:three graphs}) we see four distinct groups in increasing order of mean rank over the $N$ datasets:  \emph{A.} \{XGB\};  \emph{B.} \{LinARM1, MixMonoXGB, MixXGB, MonoXGB, ARM1\}; \emph{C.} \{MixLinARM1, LinARM2, ARM2, MixARM1\}; and  \emph{D.} \{NNLR, LinNNLR\}. 
These groups are defined by dint of their being approximately mutually indistinguishable, since they are the connected components of the CD graph.
As one might expect, XGBoost with no constraints on its functional form is the highest performing algorithm.
The worst performing group, D, having both its constituent classifiers based on Logistic Regression with no feature preprocessing, performs as expected.
The middle two groups, B and C, are more interesting.
We can draw the following conclusions from Group B: \emph{i.} on average if one is happy with the performance of MonoXGB, one is likely to be happy with the classification performance of MixMonoXGB, thus gaining useful hierarchical structure to the model with little performance degradation; and \emph{ii.} if one is satisfied with the performance of ARM1, one is likely to be satisfied with the linearised version of this model, gaining the interpretability advantages of directly working in probability space with negligible decrease in performance.
Group C is the next best performing group.
Similarly to the conclusions we draw from Group B, if one is content with the performance of ARM2, then without significant average decrease in performance one can use the linearised counterpart, gaining transparency from working directly in probability space.
By the same token, if one is considering using a mixture of ARM1 models, then without major degradation in performance one can linearise the subscale models.
Note that these groupings are at odds with the results reported in~\cite{CHEN2022113647}, namely that ARM2 displays similar classification performance to both XGB and ARM1, whereas here we have that AUC(XGB) $>$ AUC(ARM1) $>$ AUC(ARM2).
We attribute this discrepancy to the fact that the number of datasets used in this work is much greater than that in~\cite{CHEN2022113647} that only reports AUC results on the HELOC and German Credit datasets.
In terms of the absolute AUC score, Table~\ref{tab:comparison_auc} reports the average difference $\hat{\theta}_{\text{HL}}$ between each pair of classifiers over the $N$ datasets.
Indeed, apart from the Group D classifiers that have an average AUC score 0.341 below that of Group A (XGB), the magnitude of the difference in performance with respect to Group A is small, namely a median difference of 0.007 for classifiers in Group B and median difference of 0.0185 for classifiers in Group C.  

It is notable that there is no statistically significant difference in AUC between ARM2 vs MixARM1 and the difference between ARM2 and MixLinARM1 is 0.021.
This shows that the proposed ARM2 model from~\cite{CHEN2022113647} can in general have the outer layer replaced with SSPM with no penalty in AUC, and further replacing the inner layer with LAM only degrades performance slightly.

\textbf{Results for Calibration.}
Raw ECE and MCE scores for each (classifier, dataset) combination are reported in Tables~\ref{tab:ece_raw}~and~\ref{tab:mce_raw} respectively in the Appendix.
The critical difference diagrams for ECE and MCE (Figure~\ref{fig:three graphs}) do not have such distinct groups as with the AUC CD diagram -- the CD graphs have three and one connected component respectively.
The mean ranks for the algorithms are less widely distributed than for AUC.
The main emergent themes are: \emph{i.} The linearised algorithms are generally less well-calibrated as compared with their non-linearised counterparts; and \emph{ii.} The four mixture models are all in the top half performing algorithms for both ECE and MCE, with MixXGB coming out on top for both metrics.
Comparing the size of the difference in calibration (Tables~\ref{tab:comparison_ece}~and~\ref{tab:comparison_mce} in the Appendix), we see that under linearisation, ECE increases by 0.003 and MCE increases by 0.009 (taking the median over the three (X, LinX) pairs).
Mixture models trained with \textsc{SubscaleHedge} improve ECE by 0.015 and improve MCE by 0.034 (taking the median over the four (X, MixX) pairs).

\setlength{\tabcolsep}{2pt}

\begin{table}
\centering
        \caption{Point estimate for difference in AUC scores between classifiers. Negative values mean column model is better than row model. Bold values indicate statistical significance.}
        \label{tab:comparison_auc}
        {\tiny
        \begin{tabular}{lrrrrrrrrrrrr}
\toprule
{} & \rotatebox{0}{NNLR} & \rotatebox{0}{LinNNLR} & \rotatebox{0}{ARM1} & \rotatebox{0}{ARM2} & \rotatebox{0}{LinARM1} & \rotatebox{0}{LinARM2} & \rotatebox{0}{MixARM1} & \rotatebox{0}{MixLinARM1} & \rotatebox{0}{MixMonoXGB} & \rotatebox{0}{MixXGB} & \rotatebox{0}{MonoXGB} & \rotatebox{0}{XGB} \\
\midrule
NNLR       &                  --- &                   0.000 &      \textbf{-0.327} &      \textbf{-0.314} &         \textbf{-0.319} &         \textbf{-0.311} &         \textbf{-0.312} &            \textbf{-0.294} &            \textbf{-0.322} &        \textbf{-0.324} &         \textbf{-0.327} &     \textbf{-0.341} \\
LinNNLR    &                  --- &                     --- &      \textbf{-0.327} &      \textbf{-0.314} &         \textbf{-0.319} &         \textbf{-0.311} &         \textbf{-0.312} &            \textbf{-0.294} &            \textbf{-0.322} &        \textbf{-0.324} &         \textbf{-0.327} &     \textbf{-0.341} \\
ARM1       &                  --- &                     --- &                  --- &       \textbf{0.008} &          \textbf{0.003} &          \textbf{0.013} &          \textbf{0.009} &             \textbf{0.034} &                      0.002 &                  0.002 &                  -0.000 &     \textbf{-0.011} \\
ARM2       &                  --- &                     --- &                  --- &                  --- &         \textbf{-0.004} &          \textbf{0.003} &                  -0.000 &             \textbf{0.021} &            \textbf{-0.009} &        \textbf{-0.010} &         \textbf{-0.008} &     \textbf{-0.017} \\
LinARM1    &                  --- &                     --- &                  --- &                  --- &                     --- &          \textbf{0.009} &          \textbf{0.006} &             \textbf{0.027} &                     -0.002 &                 -0.002 &                  -0.003 &     \textbf{-0.013} \\
LinARM2    &                  --- &                     --- &                  --- &                  --- &                     --- &                     --- &                  -0.004 &                      0.018 &            \textbf{-0.012} &        \textbf{-0.013} &         \textbf{-0.014} &     \textbf{-0.020} \\
MixARM1    &                  --- &                     --- &                  --- &                  --- &                     --- &                     --- &                     --- &             \textbf{0.018} &            \textbf{-0.008} &        \textbf{-0.008} &         \textbf{-0.007} &     \textbf{-0.012} \\
MixLinARM1 &                  --- &                     --- &                  --- &                  --- &                     --- &                     --- &                     --- &                        --- &            \textbf{-0.028} &        \textbf{-0.031} &         \textbf{-0.026} &     \textbf{-0.047} \\
MixMonoXGB &                  --- &                     --- &                  --- &                  --- &                     --- &                     --- &                     --- &                        --- &                        --- &        \textbf{-0.001} &                  -0.002 &     \textbf{-0.007} \\
MixXGB     &                  --- &                     --- &                  --- &                  --- &                     --- &                     --- &                     --- &                        --- &                        --- &                    --- &                  -0.002 &     \textbf{-0.005} \\
MonoXGB    &                  --- &                     --- &                  --- &                  --- &                     --- &                     --- &                     --- &                        --- &                        --- &                    --- &                     --- &     \textbf{-0.002} \\
XGB        &                  --- &                     --- &                  --- &                  --- &                     --- &                     --- &                     --- &                        --- &                        --- &                    --- &                     --- &                 --- \\
\bottomrule
\end{tabular}

        }
\end{table}

\section{Conclusion}
\label{sec:conclusion}

This work introduces techniques for improving the interpretability characteristics of existing models while incurring only very small penalties in classification performance, as evaluated on a large suite of real credit modelling datasets in a statistically rigorous manner. 
For an additive logistic model such as ARM, one can use our linearisation scheme (LAM) with the model and incur only a very small reduction in AUC.
Moreover, when meaningful feature groups called subscales are modelled separately and combined with our linear opinion pooling scheme (\textsc{SubscaleHedge}), there is again only a minor decrease in classification performance, while gaining the interpretability insights afforded by hierarchical modelling.
Linear opinion pooling with weights trained using \textsc{SubscaleHedge} improves model calibration relative to the underlying subscale model trained on all features.
There is evidence provided in the experiments supporting the existence of an accuracy-interpretability tradeoff~\cite{Rudin19,Dziugaite2020,DARPA}, at least when the models under consideration are provided common resources such as data pre-processing and minimal hyperparameter tuning.

\textbf{Future Work.}
This work leads to a number of open questions.
With respect to linearisation, can we extend LAMs to multivariate shape functions, such as those including linear interaction terms and tree ensembles, while maintaining performance and interpretablity?
In the present work we only consider binary classification, will extending LAMs to the multiclass setting reduce performance?
We have only considered logistic link functions, can we define LAM for e.g. probit regression~\cite{hastie2009elements}?
In terms of very small or very large probabilites output by a logistic model, can we construct a more faithful approximation than LAMs as defined in Definition~\ref{def:linearised_additive_models} while retaining interpretability?
In terms of combining subscale models, we have obtained good results with a linear opinion pool with a Hedge-like training algorithm.
Can we gain anything in terms of performance or interpretability from considering other combination methods for probabilities~\cite{GenestZidek1986}, such as logarithmic opinion pools~\cite{NIPS1997_59f51fd6}?
While we have rigorously benchmarked classification performance and calibration, the arguments supporting increased interpretability of LAMs and SSPMs are looser.
A user study would help to remedy this.
We keenly anticipate future exploration of these research questions.

\paragraph{Disclaimer.}
This paper was prepared for informational purposes by
the Artificial Intelligence Research group of JPMorgan Chase \& Co. and its affiliates (``JP Morgan''),
and is not a product of the Research Department of JP Morgan.
JP Morgan makes no representation and warranty whatsoever and disclaims all liability,
for the completeness, accuracy or reliability of the information contained herein.
This document is not intended as investment research or investment advice, or a recommendation,
offer or solicitation for the purchase or sale of any security, financial instrument, financial product or service,
or to be used in any way for evaluating the merits of participating in any transaction,
and shall not constitute a solicitation under any jurisdiction or to any person,
if such solicitation under such jurisdiction or to such person would be unlawful.

\bibliography{references}

\appendix

\section{Proof of Lemma~\ref{lem:add_model_equiv}}
\textbf{Lemma~\ref{lem:add_model_equiv}.}
    Let $f$ be an additive model as defined in Definition~\ref{def:logistic_additive_model} with $\widetilde{f}$  its $\alpha^\star$-linearised version and $\widetilde{\sigma}$ as in~\eqref{eq:clip_lin_approx}.
    Then,
    $\widetilde{\sigma}(f(\vb*{x})) = \projunit(\widetilde{f}(\vb*{x}))$,
    that is, for an additive model $f$ the optimal clipped linear approximation to the sigmoid evaluated on the output of $f$ is formally equivalent to the output of the linearised model relative to $f$, projected to the unit interval.
\begin{proof}
We proceed via case analysis. 
\emph{Case i.} When $f(\vb*{x}) \in [-\alpha^\star, \alpha^\star]$ the required identity follows immediately from the definitions of $\widetilde{\sigma}$, $\widetilde{f}$ and $\projunit$.
Now suppose that $f(\vb*{x}) = \sum_{i=1}^{d} \beta_i f_i(x_i) = \alpha$.
It follows from substitution into the definition of $\widetilde{f}$ that 
\begin{equation}
\label{eq:approx_lemma_case_ii_iii}
\widetilde{f}(\vb*{x}) = \frac{1}{2}\qty(1 + \frac{\alpha}{\alpha^\star})
\end{equation}
\emph{Case ii.} When $\alpha > \alpha^\star$ we have that $\widetilde{f}(\vb*{x}) > 1$ by substitution into~\eqref{eq:approx_lemma_case_ii_iii}, in which case $\projunit\qty(\widetilde{f}(\vb*{x})) = 1 = \widetilde{\sigma}(f(\alpha)) = \widetilde{\sigma}(f(\vb*{x}))$, yielding the required identity.
\emph{Case iii.} When $\alpha < \alpha^\star$ we have that $\widetilde{f}(\vb*{x}) < 0$ by substitution into~\eqref{eq:approx_lemma_case_ii_iii}, in which case $\projunit\qty(\widetilde{f}(\vb*{x})) = 0 = \widetilde{\sigma}(f(\alpha)) = \widetilde{\sigma}(f(\vb*{x}))$, yielding the required identity.
We have shown that the identity holds for all values of $f(\vb*{x})$ and so the result is proved.
\end{proof}




\section{Raw Experimental Metrics}
\label{sec:raw_metrics}
See Tables~\ref{tab:auc_raw},~\ref{tab:ece_raw}~and~\ref{tab:mce_raw} for AUC, ECE and MCE respectively.

\newgeometry{top=0.1cm,bottom=0.1cm}
\setlength{\tabcolsep}{10pt}
\begin{sidewaystable}
        \captionof{table}{Raw AUC scores (mean over 10 CV folds) along with their ranks. Scores in bold are best for a particular dataset.}
        \label{tab:auc_raw}
        {\tiny
        \begin{longtable}{lrrrrrrrrrrrr}
\toprule
\textbf{Classifier} &       \rotatebox{60}{NNLR} &    \rotatebox{60}{LinNNLR} &             \rotatebox{60}{ARM1} &       \rotatebox{60}{ARM2} &          \rotatebox{60}{LinARM1} &    \rotatebox{60}{LinARM2} &    \rotatebox{60}{MixARM1} & \rotatebox{60}{MixLinARM1} &       \rotatebox{60}{MixMonoXGB} &           \rotatebox{60}{MixXGB} &          \rotatebox{60}{MonoXGB} &              \rotatebox{60}{XGB} \\
\textbf{Dataset  } &                            &                            &                                  &                            &                                  &                            &                            &                            &                                  &                                  &                                  &                                  \\
\midrule
\endfirsthead

\toprule
\textbf{Classifier} &       \rotatebox{60}{NNLR} &    \rotatebox{60}{LinNNLR} &             \rotatebox{60}{ARM1} &       \rotatebox{60}{ARM2} &          \rotatebox{60}{LinARM1} &    \rotatebox{60}{LinARM2} &    \rotatebox{60}{MixARM1} & \rotatebox{60}{MixLinARM1} &       \rotatebox{60}{MixMonoXGB} &           \rotatebox{60}{MixXGB} &          \rotatebox{60}{MonoXGB} &              \rotatebox{60}{XGB} \\
\textbf{Dataset  } &                            &                            &                                  &                            &                                  &                            &                            &                            &                                  &                                  &                                  &                                  \\
\midrule
\endhead
\midrule
\multicolumn{13}{r}{{Continued on next page}} \\
\midrule
\endfoot

\bottomrule
\endlastfoot
\textbf{Australia} &  0.9286\,\hphantom{1}(5.0) &  0.9257\,\hphantom{1}(7.0) &                   0.8979\,(10.0) &  0.9112\,\hphantom{1}(8.0) &                   0.8758\,(11.0) &  0.9039\,\hphantom{1}(9.0) &  0.9282\,\hphantom{1}(6.0) &             0.7600\,(12.0) &        0.9376\,\hphantom{1}(2.0) &        0.9352\,\hphantom{1}(3.0) &        0.9325\,\hphantom{1}(4.0) &  {\bf 0.9385\,\hphantom{1}(1.0)} \\
\textbf{GMSC     } &             0.5609\,(11.5) &             0.5609\,(11.5) &  {\bf 0.8589\,\hphantom{1}(1.5)} &  0.8560\,\hphantom{1}(7.5) &        0.8586\,\hphantom{1}(3.0) &  0.8560\,\hphantom{1}(7.5) &  0.8559\,\hphantom{1}(9.0) &             0.8148\,(10.0) &        0.8580\,\hphantom{1}(6.0) &  {\bf 0.8589\,\hphantom{1}(1.5)} &        0.8584\,\hphantom{1}(5.0) &        0.8585\,\hphantom{1}(4.0) \\
\textbf{German   } &  0.7568\,\hphantom{1}(7.5) &  0.7568\,\hphantom{1}(7.5) &  {\bf 0.7790\,\hphantom{1}(1.0)} &             0.7486\,(12.0) &        0.7789\,\hphantom{1}(2.0) &             0.7499\,(11.0) &  0.7530\,\hphantom{1}(9.0) &             0.7522\,(10.0) &        0.7599\,\hphantom{1}(6.0) &        0.7648\,\hphantom{1}(5.0) &        0.7731\,\hphantom{1}(4.0) &        0.7735\,\hphantom{1}(3.0) \\
\textbf{HELOC    } &             0.7593\,(11.5) &             0.7593\,(11.5) &  {\bf 0.7963\,\hphantom{1}(1.5)} &  0.7952\,\hphantom{1}(4.0) &  {\bf 0.7963\,\hphantom{1}(1.5)} &  0.7958\,\hphantom{1}(3.0) &  0.7817\,\hphantom{1}(9.0) &  0.7818\,\hphantom{1}(8.0) &                   0.7757\,(10.0) &        0.7842\,\hphantom{1}(7.0) &        0.7908\,\hphantom{1}(6.0) &        0.7931\,\hphantom{1}(5.0) \\
\textbf{Japan    } &  0.9322\,\hphantom{1}(4.0) &  0.9298\,\hphantom{1}(6.0) &        0.9074\,\hphantom{1}(8.0) &  0.9063\,\hphantom{1}(9.0) &                   0.8970\,(11.0) &             0.9015\,(10.0) &  0.9229\,\hphantom{1}(7.0) &             0.7557\,(12.0) &  {\bf 0.9409\,\hphantom{1}(1.0)} &        0.9361\,\hphantom{1}(3.0) &        0.9303\,\hphantom{1}(5.0) &        0.9381\,\hphantom{1}(2.0) \\
\textbf{LC\_0    } &             0.5418\,(11.5) &             0.5418\,(11.5) &        0.9955\,\hphantom{1}(5.0) &  0.9836\,\hphantom{1}(7.0) &        0.9919\,\hphantom{1}(6.0) &             0.9798\,(10.0) &  0.9829\,\hphantom{1}(8.0) &  0.9808\,\hphantom{1}(9.0) &        0.9981\,\hphantom{1}(3.5) &        0.9981\,\hphantom{1}(3.5) &        0.9993\,\hphantom{1}(2.0) &  {\bf 0.9996\,\hphantom{1}(1.0)} \\
\textbf{LC\_1    } &             0.5247\,(11.5) &             0.5247\,(11.5) &  {\bf 0.9988\,\hphantom{1}(2.0)} &  0.9955\,\hphantom{1}(8.0) &        0.9964\,\hphantom{1}(6.5) &             0.9902\,(10.0) &  0.9964\,\hphantom{1}(6.5) &  0.9946\,\hphantom{1}(9.0) &        0.9965\,\hphantom{1}(4.5) &        0.9965\,\hphantom{1}(4.5) &  {\bf 0.9988\,\hphantom{1}(2.0)} &  {\bf 0.9988\,\hphantom{1}(2.0)} \\
\textbf{LC\_2    } &             0.5000\,(11.5) &             0.5000\,(11.5) &        0.9962\,\hphantom{1}(2.0) &  0.9886\,\hphantom{1}(8.0) &        0.9933\,\hphantom{1}(4.0) &             0.9826\,(10.0) &  0.9894\,\hphantom{1}(7.0) &  0.9862\,\hphantom{1}(9.0) &        0.9924\,\hphantom{1}(5.5) &        0.9924\,\hphantom{1}(5.5) &        0.9948\,\hphantom{1}(3.0) &  {\bf 0.9970\,\hphantom{1}(1.0)} \\
\textbf{LC\_3    } &             0.5000\,(11.5) &             0.5000\,(11.5) &        0.9969\,\hphantom{1}(2.0) &  0.9903\,\hphantom{1}(8.0) &        0.9939\,\hphantom{1}(4.0) &             0.9850\,(10.0) &  0.9917\,\hphantom{1}(7.0) &  0.9877\,\hphantom{1}(9.0) &        0.9934\,\hphantom{1}(5.5) &        0.9934\,\hphantom{1}(5.5) &        0.9952\,\hphantom{1}(3.0) &  {\bf 0.9970\,\hphantom{1}(1.0)} \\
\textbf{LC\_4    } &             0.5000\,(11.5) &             0.5000\,(11.5) &        0.9964\,\hphantom{1}(2.0) &  0.9901\,\hphantom{1}(8.0) &        0.9933\,\hphantom{1}(6.0) &             0.9841\,(10.0) &  0.9912\,\hphantom{1}(7.0) &  0.9876\,\hphantom{1}(9.0) &        0.9935\,\hphantom{1}(5.0) &        0.9936\,\hphantom{1}(4.0) &        0.9952\,\hphantom{1}(3.0) &  {\bf 0.9969\,\hphantom{1}(1.0)} \\
\textbf{LC\_5    } &             0.5536\,(11.5) &             0.5536\,(11.5) &        0.9966\,\hphantom{1}(3.0) &  0.9896\,\hphantom{1}(8.0) &        0.9934\,\hphantom{1}(4.0) &             0.9829\,(10.0) &  0.9905\,\hphantom{1}(7.0) &  0.9871\,\hphantom{1}(9.0) &        0.9931\,\hphantom{1}(6.0) &        0.9932\,\hphantom{1}(5.0) &        0.9969\,\hphantom{1}(2.0) &  {\bf 0.9971\,\hphantom{1}(1.0)} \\
\textbf{LC\_6    } &             0.5599\,(11.5) &             0.5599\,(11.5) &        0.9960\,\hphantom{1}(2.0) &  0.9884\,\hphantom{1}(8.0) &        0.9922\,\hphantom{1}(6.0) &             0.9822\,(10.0) &  0.9896\,\hphantom{1}(7.0) &  0.9843\,\hphantom{1}(9.0) &        0.9934\,\hphantom{1}(5.0) &        0.9935\,\hphantom{1}(4.0) &        0.9950\,\hphantom{1}(3.0) &  {\bf 0.9972\,\hphantom{1}(1.0)} \\
\textbf{LC\_7    } &             0.5000\,(11.5) &             0.5000\,(11.5) &        0.9941\,\hphantom{1}(3.0) &  0.9863\,\hphantom{1}(8.0) &        0.9905\,\hphantom{1}(6.0) &             0.9793\,(10.0) &  0.9869\,\hphantom{1}(7.0) &  0.9835\,\hphantom{1}(9.0) &        0.9930\,\hphantom{1}(4.5) &        0.9930\,\hphantom{1}(4.5) &        0.9964\,\hphantom{1}(2.0) &  {\bf 0.9967\,\hphantom{1}(1.0)} \\
\textbf{LC\_8    } &             0.5000\,(11.5) &             0.5000\,(11.5) &        0.9938\,\hphantom{1}(3.0) &  0.9865\,\hphantom{1}(7.0) &        0.9912\,\hphantom{1}(6.0) &             0.9793\,(10.0) &  0.9856\,\hphantom{1}(8.0) &  0.9824\,\hphantom{1}(9.0) &        0.9931\,\hphantom{1}(5.0) &        0.9932\,\hphantom{1}(4.0) &        0.9954\,\hphantom{1}(2.0) &  {\bf 0.9968\,\hphantom{1}(1.0)} \\
\textbf{LC\_9    } &             0.5000\,(11.5) &             0.5000\,(11.5) &        0.9974\,\hphantom{1}(3.0) &  0.9927\,\hphantom{1}(8.0) &        0.9949\,\hphantom{1}(6.0) &             0.9868\,(10.0) &  0.9937\,\hphantom{1}(7.0) &  0.9907\,\hphantom{1}(9.0) &        0.9963\,\hphantom{1}(5.0) &        0.9964\,\hphantom{1}(4.0) &  {\bf 0.9979\,\hphantom{1}(1.5)} &  {\bf 0.9979\,\hphantom{1}(1.5)} \\
\textbf{LC\_10   } &             0.5211\,(11.5) &             0.5211\,(11.5) &        0.9961\,\hphantom{1}(5.0) &  0.9882\,\hphantom{1}(8.0) &        0.9930\,\hphantom{1}(6.0) &             0.9831\,(10.0) &  0.9887\,\hphantom{1}(7.0) &  0.9871\,\hphantom{1}(9.0) &        0.9965\,\hphantom{1}(3.5) &        0.9965\,\hphantom{1}(3.5) &  {\bf 0.9973\,\hphantom{1}(1.5)} &  {\bf 0.9973\,\hphantom{1}(1.5)} \\
\textbf{LC\_11   } &             0.5000\,(11.5) &             0.5000\,(11.5) &        0.9962\,\hphantom{1}(3.0) &  0.9916\,\hphantom{1}(8.0) &        0.9925\,\hphantom{1}(7.0) &             0.9858\,(10.0) &  0.9934\,\hphantom{1}(6.0) &  0.9891\,\hphantom{1}(9.0) &        0.9953\,\hphantom{1}(5.0) &        0.9954\,\hphantom{1}(4.0) &  {\bf 0.9966\,\hphantom{1}(1.5)} &  {\bf 0.9966\,\hphantom{1}(1.5)} \\
\textbf{LC\_12   } &             0.6812\,(11.0) &             0.6811\,(12.0) &        0.9843\,\hphantom{1}(5.0) &  0.9800\,\hphantom{1}(8.0) &        0.9806\,\hphantom{1}(6.0) &             0.9758\,(10.0) &  0.9802\,\hphantom{1}(7.0) &  0.9790\,\hphantom{1}(9.0) &        0.9853\,\hphantom{1}(4.0) &        0.9864\,\hphantom{1}(3.0) &  {\bf 0.9914\,\hphantom{1}(1.5)} &  {\bf 0.9914\,\hphantom{1}(1.5)} \\
\textbf{Poland\_0} &             0.6240\,(11.5) &             0.6240\,(11.5) &        0.8833\,\hphantom{1}(5.0) &  0.8677\,\hphantom{1}(8.0) &        0.8752\,\hphantom{1}(6.0) &  0.8687\,\hphantom{1}(7.0) &  0.8644\,\hphantom{1}(9.0) &             0.7721\,(10.0) &        0.8912\,\hphantom{1}(4.0) &        0.8983\,\hphantom{1}(3.0) &        0.9283\,\hphantom{1}(2.0) &  {\bf 0.9315\,\hphantom{1}(1.0)} \\
\textbf{Poland\_1} &             0.5499\,(11.5) &             0.5499\,(11.5) &        0.8196\,\hphantom{1}(4.0) &  0.7893\,\hphantom{1}(7.0) &        0.8181\,\hphantom{1}(5.0) &  0.7910\,\hphantom{1}(6.0) &  0.7855\,\hphantom{1}(8.0) &             0.7205\,(10.0) &        0.8490\,\hphantom{1}(3.0) &        0.8515\,\hphantom{1}(2.0) &        0.7691\,\hphantom{1}(9.0) &  {\bf 0.8763\,\hphantom{1}(1.0)} \\
\textbf{Poland\_2} &             0.6102\,(11.5) &             0.6102\,(11.5) &        0.8502\,\hphantom{1}(2.0) &  0.8169\,\hphantom{1}(7.0) &        0.8477\,\hphantom{1}(3.0) &  0.8199\,\hphantom{1}(6.0) &  0.8074\,\hphantom{1}(8.0) &             0.7742\,(10.0) &        0.8336\,\hphantom{1}(5.0) &        0.8359\,\hphantom{1}(4.0) &        0.7908\,\hphantom{1}(9.0) &  {\bf 0.8755\,\hphantom{1}(1.0)} \\
\textbf{Poland\_3} &             0.6635\,(11.5) &             0.6635\,(11.5) &        0.8618\,\hphantom{1}(2.0) &  0.8242\,\hphantom{1}(8.0) &        0.8583\,\hphantom{1}(3.0) &  0.8273\,\hphantom{1}(7.0) &  0.8133\,\hphantom{1}(9.0) &             0.7866\,(10.0) &        0.8440\,\hphantom{1}(6.0) &        0.8456\,\hphantom{1}(5.0) &        0.8546\,\hphantom{1}(4.0) &  {\bf 0.8979\,\hphantom{1}(1.0)} \\
\textbf{Poland\_4} &             0.7423\,(12.0) &             0.7424\,(11.0) &        0.8997\,\hphantom{1}(3.0) &  0.8808\,\hphantom{1}(7.0) &        0.8869\,\hphantom{1}(6.0) &  0.8791\,\hphantom{1}(8.0) &  0.8754\,\hphantom{1}(9.0) &             0.8403\,(10.0) &        0.8927\,\hphantom{1}(5.0) &        0.8963\,\hphantom{1}(4.0) &        0.9151\,\hphantom{1}(2.0) &  {\bf 0.9266\,\hphantom{1}(1.0)} \\
\textbf{Taiwan   } &             0.6360\,(11.5) &             0.6360\,(11.5) &  {\bf 0.7771\,\hphantom{1}(1.0)} &  0.7688\,\hphantom{1}(8.0) &        0.7770\,\hphantom{1}(2.0) &  0.7696\,\hphantom{1}(7.0) &  0.7680\,\hphantom{1}(9.0) &             0.7361\,(10.0) &        0.7704\,\hphantom{1}(6.0) &        0.7705\,\hphantom{1}(5.0) &        0.7741\,\hphantom{1}(4.0) &        0.7748\,\hphantom{1}(3.0) \\
\end{longtable}

        }
\end{sidewaystable}

\begin{sidewaystable}
        \captionof{table}{Raw ECE scores (mean over 10 CV folds) along with their ranks. Scores in bold are best for a particular dataset.}
        \label{tab:ece_raw}
        {\tiny
        \begin{longtable}{lrrrrrrrrrrrr}
\toprule
\textbf{Classifier} &             \rotatebox{60}{NNLR} &          \rotatebox{60}{LinNNLR} &       \rotatebox{60}{ARM1} &       \rotatebox{60}{ARM2} & \rotatebox{60}{LinARM1} &    \rotatebox{60}{LinARM2} &          \rotatebox{60}{MixARM1} &       \rotatebox{60}{MixLinARM1} &       \rotatebox{60}{MixMonoXGB} &           \rotatebox{60}{MixXGB} &          \rotatebox{60}{MonoXGB} &        \rotatebox{60}{XGB} \\
\textbf{Dataset  } &                                  &                                  &                            &                            &                         &                            &                                  &                                  &                                  &                                  &                                  &                            \\
\midrule
\endfirsthead

\toprule
\textbf{Classifier} &             \rotatebox{60}{NNLR} &          \rotatebox{60}{LinNNLR} &       \rotatebox{60}{ARM1} &       \rotatebox{60}{ARM2} & \rotatebox{60}{LinARM1} &    \rotatebox{60}{LinARM2} &          \rotatebox{60}{MixARM1} &       \rotatebox{60}{MixLinARM1} &       \rotatebox{60}{MixMonoXGB} &           \rotatebox{60}{MixXGB} &          \rotatebox{60}{MonoXGB} &        \rotatebox{60}{XGB} \\
\textbf{Dataset  } &                                  &                                  &                            &                            &                         &                            &                                  &                                  &                                  &                                  &                                  &                            \\
\midrule
\endhead
\midrule
\multicolumn{13}{r}{{Continued on next page}} \\
\midrule
\endfoot

\bottomrule
\endlastfoot
\textbf{Australia} &        0.2100\,\hphantom{1}(7.0) &                   0.2242\,(10.0) &             0.2343\,(11.0) &  0.2142\,\hphantom{1}(8.0) &          0.2363\,(12.0) &  0.2167\,\hphantom{1}(9.0) &        0.1363\,\hphantom{1}(2.0) &  {\bf 0.1303\,\hphantom{1}(1.0)} &        0.1624\,\hphantom{1}(4.0) &        0.1610\,\hphantom{1}(3.0) &        0.2015\,\hphantom{1}(5.5) &  0.2015\,\hphantom{1}(5.5) \\
\textbf{GMSC     } &  {\bf 0.2166\,\hphantom{1}(3.5)} &  {\bf 0.2166\,\hphantom{1}(3.5)} &             0.2171\,(10.0) &  0.2169\,\hphantom{1}(9.0) &          0.2193\,(12.0) &             0.2185\,(11.0) &  {\bf 0.2166\,\hphantom{1}(3.5)} &  {\bf 0.2166\,\hphantom{1}(3.5)} &  {\bf 0.2166\,\hphantom{1}(3.5)} &  {\bf 0.2166\,\hphantom{1}(3.5)} &        0.2167\,\hphantom{1}(7.5) &  0.2167\,\hphantom{1}(7.5) \\
\textbf{German   } &        0.1297\,\hphantom{1}(8.0) &        0.1273\,\hphantom{1}(7.0) &             0.1689\,(12.0) &             0.1481\,(10.0) &          0.1660\,(11.0) &  0.1448\,\hphantom{1}(9.0) &        0.1191\,\hphantom{1}(4.0) &        0.1134\,\hphantom{1}(3.0) &  {\bf 0.1102\,\hphantom{1}(1.0)} &        0.1109\,\hphantom{1}(2.0) &        0.1260\,\hphantom{1}(5.0) &  0.1268\,\hphantom{1}(6.0) \\
\textbf{HELOC    } &        0.0991\,\hphantom{1}(6.0) &        0.0886\,\hphantom{1}(5.0) &             0.1172\,(11.0) &             0.1177\,(12.0) &          0.1121\,(10.0) &  0.1085\,\hphantom{1}(9.0) &        0.0530\,\hphantom{1}(3.0) &  {\bf 0.0472\,\hphantom{1}(1.0)} &        0.0523\,\hphantom{1}(2.0) &        0.0544\,\hphantom{1}(4.0) &        0.1033\,\hphantom{1}(7.0) &  0.1068\,\hphantom{1}(8.0) \\
\textbf{Japan    } &        0.1999\,\hphantom{1}(7.0) &        0.2039\,\hphantom{1}(8.0) &             0.2294\,(11.0) &  0.2081\,\hphantom{1}(9.0) &          0.2326\,(12.0) &             0.2159\,(10.0) &        0.1351\,\hphantom{1}(2.0) &  {\bf 0.1295\,\hphantom{1}(1.0)} &        0.1396\,\hphantom{1}(3.0) &        0.1518\,\hphantom{1}(4.0) &        0.1901\,\hphantom{1}(6.0) &  0.1874\,\hphantom{1}(5.0) \\
\textbf{LC\_0    } &  {\bf 0.1720\,\hphantom{1}(1.0)} &        0.1722\,\hphantom{1}(2.0) &  0.2370\,\hphantom{1}(7.0) &  0.2256\,\hphantom{1}(5.0) &          0.2414\,(12.0) &  0.2345\,\hphantom{1}(6.0) &        0.2222\,\hphantom{1}(3.0) &        0.2245\,\hphantom{1}(4.0) &        0.2396\,\hphantom{1}(8.5) &        0.2396\,\hphantom{1}(8.5) &                   0.2407\,(10.0) &             0.2411\,(11.0) \\
\textbf{LC\_1    } &  {\bf 0.1718\,\hphantom{1}(1.5)} &  {\bf 0.1718\,\hphantom{1}(1.5)} &             0.2458\,(10.0) &  0.2406\,\hphantom{1}(8.0) &          0.2476\,(12.0) &             0.2471\,(11.0) &        0.2396\,\hphantom{1}(7.0) &        0.2431\,\hphantom{1}(9.0) &        0.2267\,\hphantom{1}(3.5) &        0.2267\,\hphantom{1}(3.5) &        0.2329\,\hphantom{1}(5.5) &  0.2329\,\hphantom{1}(5.5) \\
\textbf{LC\_2    } &  {\bf 0.1642\,\hphantom{1}(1.5)} &  {\bf 0.1642\,\hphantom{1}(1.5)} &             0.2382\,(10.0) &  0.2299\,\hphantom{1}(8.0) &          0.2412\,(12.0) &             0.2387\,(11.0) &        0.2285\,\hphantom{1}(7.0) &        0.2315\,\hphantom{1}(9.0) &        0.2124\,\hphantom{1}(3.5) &        0.2124\,\hphantom{1}(3.5) &        0.2214\,\hphantom{1}(5.0) &  0.2258\,\hphantom{1}(6.0) \\
\textbf{LC\_3    } &  {\bf 0.1535\,\hphantom{1}(1.5)} &  {\bf 0.1535\,\hphantom{1}(1.5)} &             0.2395\,(10.0) &  0.2326\,\hphantom{1}(8.0) &          0.2426\,(12.0) &             0.2399\,(11.0) &        0.2307\,\hphantom{1}(7.0) &        0.2334\,\hphantom{1}(9.0) &        0.2098\,\hphantom{1}(4.0) &        0.2097\,\hphantom{1}(3.0) &        0.2234\,\hphantom{1}(5.0) &  0.2252\,\hphantom{1}(6.0) \\
\textbf{LC\_4    } &  {\bf 0.1500\,\hphantom{1}(1.5)} &  {\bf 0.1500\,\hphantom{1}(1.5)} &             0.2380\,(10.0) &  0.2321\,\hphantom{1}(8.0) &          0.2408\,(12.0) &             0.2393\,(11.0) &        0.2301\,\hphantom{1}(7.0) &        0.2332\,\hphantom{1}(9.0) &        0.2077\,\hphantom{1}(3.5) &        0.2077\,\hphantom{1}(3.5) &        0.2231\,\hphantom{1}(5.0) &  0.2248\,\hphantom{1}(6.0) \\
\textbf{LC\_5    } &        0.1463\,\hphantom{1}(2.0) &  {\bf 0.1462\,\hphantom{1}(1.0)} &             0.2381\,(10.0) &  0.2312\,\hphantom{1}(8.0) &          0.2415\,(12.0) &             0.2394\,(11.0) &        0.2293\,\hphantom{1}(7.0) &        0.2316\,\hphantom{1}(9.0) &        0.2045\,\hphantom{1}(4.0) &        0.2044\,\hphantom{1}(3.0) &        0.2227\,\hphantom{1}(5.0) &  0.2253\,\hphantom{1}(6.0) \\
\textbf{LC\_6    } &  {\bf 0.1490\,\hphantom{1}(1.5)} &  {\bf 0.1490\,\hphantom{1}(1.5)} &             0.2372\,(10.0) &  0.2293\,\hphantom{1}(9.0) &          0.2405\,(12.0) &             0.2375\,(11.0) &        0.2278\,\hphantom{1}(7.0) &        0.2279\,\hphantom{1}(8.0) &        0.2033\,\hphantom{1}(3.5) &        0.2033\,\hphantom{1}(3.5) &        0.2212\,\hphantom{1}(5.0) &  0.2257\,\hphantom{1}(6.0) \\
\textbf{LC\_7    } &  {\bf 0.1471\,\hphantom{1}(1.5)} &  {\bf 0.1471\,\hphantom{1}(1.5)} &             0.2349\,(10.0) &  0.2281\,\hphantom{1}(9.0) &          0.2384\,(12.0) &             0.2370\,(11.0) &        0.2259\,\hphantom{1}(6.0) &        0.2279\,\hphantom{1}(8.0) &        0.2007\,\hphantom{1}(4.0) &        0.2006\,\hphantom{1}(3.0) &        0.2234\,\hphantom{1}(5.0) &  0.2269\,\hphantom{1}(7.0) \\
\textbf{LC\_8    } &  {\bf 0.1399\,\hphantom{1}(1.5)} &  {\bf 0.1399\,\hphantom{1}(1.5)} &             0.2326\,(10.0) &  0.2282\,\hphantom{1}(9.0) &          0.2352\,(11.0) &             0.2367\,(12.0) &        0.2226\,\hphantom{1}(6.0) &        0.2242\,\hphantom{1}(7.0) &        0.1988\,\hphantom{1}(4.0) &        0.1987\,\hphantom{1}(3.0) &        0.2200\,\hphantom{1}(5.0) &  0.2272\,\hphantom{1}(8.0) \\
\textbf{LC\_9    } &  {\bf 0.1154\,\hphantom{1}(1.5)} &  {\bf 0.1154\,\hphantom{1}(1.5)} &             0.2374\,(10.0) &  0.2361\,\hphantom{1}(9.0) &          0.2390\,(11.0) &             0.2401\,(12.0) &        0.2314\,\hphantom{1}(5.0) &        0.2331\,\hphantom{1}(8.0) &        0.2118\,\hphantom{1}(4.0) &        0.2117\,\hphantom{1}(3.0) &        0.2316\,\hphantom{1}(6.5) &  0.2316\,\hphantom{1}(6.5) \\
\textbf{LC\_10   } &  {\bf 0.1215\,\hphantom{1}(1.0)} &        0.1216\,\hphantom{1}(2.0) &             0.2384\,(10.0) &  0.2351\,\hphantom{1}(9.0) &          0.2402\,(11.5) &             0.2402\,(11.5) &        0.2318\,\hphantom{1}(8.0) &        0.2313\,\hphantom{1}(5.0) &        0.2179\,\hphantom{1}(4.0) &        0.2178\,\hphantom{1}(3.0) &        0.2316\,\hphantom{1}(6.5) &  0.2316\,\hphantom{1}(6.5) \\
\textbf{LC\_11   } &  {\bf 0.1137\,\hphantom{1}(1.5)} &  {\bf 0.1137\,\hphantom{1}(1.5)} &             0.2396\,(10.0) &  0.2385\,\hphantom{1}(9.0) &          0.2419\,(11.0) &             0.2432\,(12.0) &        0.2350\,\hphantom{1}(8.0) &        0.2342\,\hphantom{1}(7.0) &        0.2189\,\hphantom{1}(4.0) &        0.2188\,\hphantom{1}(3.0) &        0.2311\,\hphantom{1}(5.5) &  0.2311\,\hphantom{1}(5.5) \\
\textbf{LC\_12   } &  {\bf 0.1325\,\hphantom{1}(1.0)} &        0.1327\,\hphantom{1}(2.0) &  0.2329\,\hphantom{1}(9.0) &             0.2353\,(10.0) &          0.2375\,(11.0) &             0.2428\,(12.0) &        0.2292\,\hphantom{1}(6.0) &        0.1348\,\hphantom{1}(3.0) &        0.2215\,\hphantom{1}(4.0) &        0.2219\,\hphantom{1}(5.0) &        0.2293\,\hphantom{1}(7.5) &  0.2293\,\hphantom{1}(7.5) \\
\textbf{Poland\_0} &        0.2311\,\hphantom{1}(5.5) &        0.2313\,\hphantom{1}(7.0) &             0.2367\,(10.0) &  0.2311\,\hphantom{1}(5.5) &          0.2414\,(12.0) &             0.2374\,(11.0) &        0.2309\,\hphantom{1}(4.0) &  {\bf 0.2307\,\hphantom{1}(1.0)} &        0.2308\,\hphantom{1}(2.5) &        0.2308\,\hphantom{1}(2.5) &        0.2318\,\hphantom{1}(9.0) &  0.2314\,\hphantom{1}(8.0) \\
\textbf{Poland\_1} &        0.2305\,\hphantom{1}(7.0) &        0.2305\,\hphantom{1}(7.0) &             0.2332\,(11.0) &  0.2306\,\hphantom{1}(9.0) &          0.2380\,(12.0) &             0.2330\,(10.0) &        0.2304\,\hphantom{1}(3.5) &  {\bf 0.2303\,\hphantom{1}(1.0)} &        0.2304\,\hphantom{1}(3.5) &        0.2304\,\hphantom{1}(3.5) &        0.2304\,\hphantom{1}(3.5) &  0.2305\,\hphantom{1}(7.0) \\
\textbf{Poland\_2} &        0.2268\,\hphantom{1}(8.5) &        0.2268\,\hphantom{1}(8.5) &             0.2304\,(10.0) &  0.2266\,\hphantom{1}(7.0) &          0.2361\,(12.0) &             0.2317\,(11.0) &  {\bf 0.2264\,\hphantom{1}(3.0)} &  {\bf 0.2264\,\hphantom{1}(3.0)} &  {\bf 0.2264\,\hphantom{1}(3.0)} &  {\bf 0.2264\,\hphantom{1}(3.0)} &  {\bf 0.2264\,\hphantom{1}(3.0)} &  0.2265\,\hphantom{1}(6.0) \\
\textbf{Poland\_3} &        0.2239\,\hphantom{1}(5.5) &        0.2239\,\hphantom{1}(5.5) &             0.2300\,(10.0) &  0.2244\,\hphantom{1}(9.0) &          0.2361\,(12.0) &             0.2307\,(11.0) &  {\bf 0.2237\,\hphantom{1}(2.5)} &  {\bf 0.2237\,\hphantom{1}(2.5)} &  {\bf 0.2237\,\hphantom{1}(2.5)} &  {\bf 0.2237\,\hphantom{1}(2.5)} &        0.2240\,\hphantom{1}(7.0) &  0.2241\,\hphantom{1}(8.0) \\
\textbf{Poland\_4} &        0.2164\,\hphantom{1}(5.0) &        0.2167\,\hphantom{1}(6.5) &             0.2292\,(11.0) &  0.2184\,\hphantom{1}(9.0) &          0.2356\,(12.0) &             0.2277\,(10.0) &  {\bf 0.2155\,\hphantom{1}(1.0)} &        0.2157\,\hphantom{1}(2.5) &        0.2157\,\hphantom{1}(2.5) &        0.2159\,\hphantom{1}(4.0) &        0.2169\,\hphantom{1}(8.0) &  0.2167\,\hphantom{1}(6.5) \\
\textbf{Taiwan   } &        0.1409\,\hphantom{1}(5.0) &        0.1411\,\hphantom{1}(6.0) &             0.1497\,(12.0) &  0.1480\,\hphantom{1}(9.0) &          0.1495\,(11.0) &             0.1483\,(10.0) &        0.1397\,\hphantom{1}(3.5) &  {\bf 0.1394\,\hphantom{1}(1.0)} &        0.1397\,\hphantom{1}(3.5) &        0.1395\,\hphantom{1}(2.0) &        0.1445\,\hphantom{1}(7.0) &  0.1450\,\hphantom{1}(8.0) \\
\end{longtable}

        }
\end{sidewaystable}

\begin{sidewaystable}
        \captionof{table}{Raw MCE scores (mean over 10 CV folds) along with their ranks. Scores in bold are best for a particular dataset.}
        \label{tab:mce_raw}
        {\tiny
        \begin{longtable}{lrrrrrrrrrrrr}
\toprule
\textbf{Classifier} &             \rotatebox{60}{NNLR} &          \rotatebox{60}{LinNNLR} &       \rotatebox{60}{ARM1} &       \rotatebox{60}{ARM2} &    \rotatebox{60}{LinARM1} &    \rotatebox{60}{LinARM2} &          \rotatebox{60}{MixARM1} &       \rotatebox{60}{MixLinARM1} &       \rotatebox{60}{MixMonoXGB} &           \rotatebox{60}{MixXGB} &    \rotatebox{60}{MonoXGB} &        \rotatebox{60}{XGB} \\
\textbf{Dataset  } &                                  &                                  &                            &                            &                            &                            &                                  &                                  &                                  &                                  &                            &                            \\
\midrule
\endfirsthead

\toprule
\textbf{Classifier} &             \rotatebox{60}{NNLR} &          \rotatebox{60}{LinNNLR} &       \rotatebox{60}{ARM1} &       \rotatebox{60}{ARM2} &    \rotatebox{60}{LinARM1} &    \rotatebox{60}{LinARM2} &          \rotatebox{60}{MixARM1} &       \rotatebox{60}{MixLinARM1} &       \rotatebox{60}{MixMonoXGB} &           \rotatebox{60}{MixXGB} &    \rotatebox{60}{MonoXGB} &        \rotatebox{60}{XGB} \\
\textbf{Dataset  } &                                  &                                  &                            &                            &                            &                            &                                  &                                  &                                  &                                  &                            &                            \\
\midrule
\endhead
\midrule
\multicolumn{13}{r}{{Continued on next page}} \\
\midrule
\endfoot

\bottomrule
\endlastfoot
\textbf{Australia} &        0.7280\,\hphantom{1}(3.0) &        0.7390\,\hphantom{1}(5.0) &  0.7604\,\hphantom{1}(6.0) &             0.7961\,(10.0) &  0.7262\,\hphantom{1}(2.0) &             0.8084\,(12.0) &  {\bf 0.7116\,\hphantom{1}(1.0)} &        0.7933\,\hphantom{1}(9.0) &                   0.8013\,(11.0) &        0.7318\,\hphantom{1}(4.0) &  0.7630\,\hphantom{1}(7.0) &  0.7821\,\hphantom{1}(8.0) \\
\textbf{GMSC     } &                   0.9116\,(11.0) &                   0.9208\,(12.0) &  0.8934\,\hphantom{1}(8.0) &  0.8871\,\hphantom{1}(7.0) &             0.9094\,(10.0) &  0.9045\,\hphantom{1}(9.0) &        0.7742\,\hphantom{1}(2.0) &  {\bf 0.6765\,\hphantom{1}(1.0)} &        0.7921\,\hphantom{1}(4.0) &        0.7904\,\hphantom{1}(3.0) &  0.8841\,\hphantom{1}(5.0) &  0.8853\,\hphantom{1}(6.0) \\
\textbf{German   } &        0.7185\,\hphantom{1}(3.0) &                   0.8270\,(11.0) &  0.7700\,\hphantom{1}(8.0) &             0.8261\,(10.0) &  0.8014\,\hphantom{1}(9.0) &             0.9089\,(12.0) &        0.7368\,\hphantom{1}(4.0) &        0.7452\,\hphantom{1}(5.0) &  {\bf 0.6809\,\hphantom{1}(1.0)} &        0.7183\,\hphantom{1}(2.0) &  0.7473\,\hphantom{1}(7.0) &  0.7468\,\hphantom{1}(6.0) \\
\textbf{HELOC    } &                   0.6564\,(12.0) &                   0.5876\,(11.0) &             0.5601\,(10.0) &  0.5074\,\hphantom{1}(7.0) &  0.5406\,\hphantom{1}(9.0) &  0.5399\,\hphantom{1}(8.0) &        0.4810\,\hphantom{1}(6.0) &        0.2916\,\hphantom{1}(3.0) &  {\bf 0.2401\,\hphantom{1}(1.0)} &        0.2908\,\hphantom{1}(2.0) &  0.4430\,\hphantom{1}(4.0) &  0.4691\,\hphantom{1}(5.0) \\
\textbf{Japan    } &        0.7409\,\hphantom{1}(6.0) &                   0.8187\,(12.0) &             0.8078\,(11.0) &  0.7169\,\hphantom{1}(2.0) &             0.7971\,(10.0) &  0.7427\,\hphantom{1}(7.0) &        0.7199\,\hphantom{1}(3.0) &        0.7242\,\hphantom{1}(4.0) &        0.7823\,\hphantom{1}(9.0) &  {\bf 0.6826\,\hphantom{1}(1.0)} &  0.7355\,\hphantom{1}(5.0) &  0.7563\,\hphantom{1}(8.0) \\
\textbf{LC\_0    } &        0.4441\,\hphantom{1}(2.0) &  {\bf 0.4430\,\hphantom{1}(1.0)} &  0.8355\,\hphantom{1}(6.0) &  0.8181\,\hphantom{1}(3.0) &             0.8423\,(10.0) &  0.8407\,\hphantom{1}(9.0) &        0.8318\,\hphantom{1}(5.0) &        0.8401\,\hphantom{1}(8.0) &                   0.8453\,(11.5) &                   0.8453\,(11.5) &  0.8315\,\hphantom{1}(4.0) &  0.8371\,\hphantom{1}(7.0) \\
\textbf{LC\_1    } &  {\bf 0.5235\,\hphantom{1}(1.0)} &        0.5636\,\hphantom{1}(2.0) &  0.8400\,\hphantom{1}(9.0) &  0.8349\,\hphantom{1}(7.0) &             0.8455\,(11.0) &             0.8542\,(12.0) &        0.8389\,\hphantom{1}(8.0) &                   0.8433\,(10.0) &        0.8243\,\hphantom{1}(4.0) &        0.8242\,\hphantom{1}(3.0) &  0.8248\,\hphantom{1}(5.0) &  0.8249\,\hphantom{1}(6.0) \\
\textbf{LC\_2    } &  {\bf 0.3284\,\hphantom{1}(1.5)} &  {\bf 0.3284\,\hphantom{1}(1.5)} &  0.8237\,\hphantom{1}(9.0) &  0.8067\,\hphantom{1}(5.0) &             0.8282\,(12.0) &             0.8275\,(11.0) &        0.8202\,\hphantom{1}(8.0) &                   0.8239\,(10.0) &        0.8077\,\hphantom{1}(7.0) &        0.8076\,\hphantom{1}(6.0) &  0.8043\,\hphantom{1}(4.0) &  0.8030\,\hphantom{1}(3.0) \\
\textbf{LC\_3    } &  {\bf 0.3069\,\hphantom{1}(1.5)} &  {\bf 0.3069\,\hphantom{1}(1.5)} &  0.8017\,\hphantom{1}(8.0) &  0.7893\,\hphantom{1}(7.0) &             0.8068\,(11.0) &             0.8069\,(12.0) &        0.8018\,\hphantom{1}(9.0) &                   0.8061\,(10.0) &        0.7854\,\hphantom{1}(5.5) &        0.7854\,\hphantom{1}(5.5) &  0.7824\,\hphantom{1}(3.0) &  0.7828\,\hphantom{1}(4.0) \\
\textbf{LC\_4    } &  {\bf 0.3001\,\hphantom{1}(1.5)} &  {\bf 0.3001\,\hphantom{1}(1.5)} &  0.7937\,\hphantom{1}(9.0) &  0.7796\,\hphantom{1}(7.0) &             0.7975\,(10.0) &             0.7984\,(11.0) &        0.7927\,\hphantom{1}(8.0) &                   0.8053\,(12.0) &        0.7746\,\hphantom{1}(6.0) &        0.7744\,\hphantom{1}(5.0) &  0.7740\,\hphantom{1}(4.0) &  0.7736\,\hphantom{1}(3.0) \\
\textbf{LC\_5    } &                   0.8344\,(12.0) &                   0.8173\,(11.0) &  0.7869\,\hphantom{1}(7.0) &  0.7722\,\hphantom{1}(5.0) &  0.7910\,\hphantom{1}(9.0) &             0.7921\,(10.0) &        0.7860\,\hphantom{1}(6.0) &        0.7904\,\hphantom{1}(8.0) &        0.7647\,\hphantom{1}(2.0) &  {\bf 0.7646\,\hphantom{1}(1.0)} &  0.7681\,\hphantom{1}(4.0) &  0.7668\,\hphantom{1}(3.0) \\
\textbf{LC\_6    } &                   0.8402\,(11.0) &                   0.9049\,(12.0) &  0.7898\,\hphantom{1}(8.0) &  0.7748\,\hphantom{1}(5.0) &             0.7954\,(10.0) &  0.7947\,\hphantom{1}(9.0) &        0.7869\,\hphantom{1}(6.0) &        0.7888\,\hphantom{1}(7.0) &  {\bf 0.7674\,\hphantom{1}(1.5)} &  {\bf 0.7674\,\hphantom{1}(1.5)} &  0.7730\,\hphantom{1}(4.0) &  0.7703\,\hphantom{1}(3.0) \\
\textbf{LC\_7    } &  {\bf 0.2943\,\hphantom{1}(1.5)} &  {\bf 0.2943\,\hphantom{1}(1.5)} &  0.7885\,\hphantom{1}(9.0) &  0.7727\,\hphantom{1}(7.0) &             0.7942\,(11.0) &             0.7945\,(12.0) &        0.7884\,\hphantom{1}(8.0) &                   0.7910\,(10.0) &        0.7680\,\hphantom{1}(5.0) &        0.7678\,\hphantom{1}(4.0) &  0.7660\,\hphantom{1}(3.0) &  0.7691\,\hphantom{1}(6.0) \\
\textbf{LC\_8    } &  {\bf 0.2799\,\hphantom{1}(1.5)} &  {\bf 0.2799\,\hphantom{1}(1.5)} &  0.7745\,\hphantom{1}(9.0) &  0.7577\,\hphantom{1}(7.0) &             0.7784\,(11.0) &             0.7790\,(12.0) &        0.7726\,\hphantom{1}(8.0) &                   0.7750\,(10.0) &        0.7536\,\hphantom{1}(5.0) &        0.7532\,\hphantom{1}(4.0) &  0.7499\,\hphantom{1}(3.0) &  0.7541\,\hphantom{1}(6.0) \\
\textbf{LC\_9    } &  {\bf 0.2308\,\hphantom{1}(1.5)} &  {\bf 0.2308\,\hphantom{1}(1.5)} &  0.7268\,\hphantom{1}(9.0) &  0.7190\,\hphantom{1}(7.0) &             0.7300\,(10.0) &             0.7306\,(11.0) &        0.7250\,\hphantom{1}(8.0) &                   0.7578\,(12.0) &        0.7078\,\hphantom{1}(4.0) &        0.7075\,\hphantom{1}(3.0) &  0.7158\,\hphantom{1}(5.5) &  0.7158\,\hphantom{1}(5.5) \\
\textbf{LC\_10   } &  {\bf 0.4295\,\hphantom{1}(1.0)} &        0.4489\,\hphantom{1}(2.0) &             0.7384\,(10.0) &  0.7289\,\hphantom{1}(7.0) &             0.7421\,(12.0) &             0.7417\,(11.0) &        0.7355\,\hphantom{1}(9.0) &        0.7338\,\hphantom{1}(8.0) &        0.7204\,\hphantom{1}(4.0) &        0.7200\,\hphantom{1}(3.0) &  0.7288\,\hphantom{1}(5.5) &  0.7288\,\hphantom{1}(5.5) \\
\textbf{LC\_11   } &  {\bf 0.2275\,\hphantom{1}(1.5)} &  {\bf 0.2275\,\hphantom{1}(1.5)} &             0.7206\,(10.0) &  0.7132\,\hphantom{1}(7.0) &             0.7259\,(11.0) &             0.7464\,(12.0) &        0.7192\,\hphantom{1}(9.0) &        0.7167\,\hphantom{1}(8.0) &        0.7010\,\hphantom{1}(4.0) &        0.7008\,\hphantom{1}(3.0) &  0.7094\,\hphantom{1}(5.5) &  0.7094\,\hphantom{1}(5.5) \\
\textbf{LC\_12   } &                   0.8079\,(12.0) &                   0.8050\,(11.0) &  0.7256\,\hphantom{1}(7.0) &  0.7186\,\hphantom{1}(5.5) &  0.7340\,\hphantom{1}(9.0) &             0.7349\,(10.0) &        0.7271\,\hphantom{1}(8.0) &  {\bf 0.5574\,\hphantom{1}(1.0)} &        0.7141\,\hphantom{1}(3.0) &        0.7139\,\hphantom{1}(2.0) &  0.7186\,\hphantom{1}(5.5) &  0.7185\,\hphantom{1}(4.0) \\
\textbf{Poland\_0} &                   0.9792\,(11.0) &                   0.9969\,(12.0) &  0.9555\,\hphantom{1}(8.0) &  0.9369\,\hphantom{1}(7.0) &             0.9645\,(10.0) &  0.9621\,\hphantom{1}(9.0) &        0.9342\,\hphantom{1}(6.0) &  {\bf 0.8171\,\hphantom{1}(1.0)} &        0.9266\,\hphantom{1}(5.0) &        0.8844\,\hphantom{1}(2.0) &  0.9162\,\hphantom{1}(3.0) &  0.9183\,\hphantom{1}(4.0) \\
\textbf{Poland\_1} &                   0.9690\,(11.0) &                   0.9707\,(12.0) &  0.9391\,\hphantom{1}(8.0) &  0.9189\,\hphantom{1}(6.0) &  0.9575\,\hphantom{1}(9.0) &             0.9603\,(10.0) &        0.8449\,\hphantom{1}(3.0) &  {\bf 0.7408\,\hphantom{1}(1.0)} &        0.8681\,\hphantom{1}(4.0) &        0.8282\,\hphantom{1}(2.0) &  0.8881\,\hphantom{1}(5.0) &  0.9272\,\hphantom{1}(7.0) \\
\textbf{Poland\_2} &        0.9495\,\hphantom{1}(9.0) &                   0.9651\,(12.0) &  0.9350\,\hphantom{1}(8.0) &  0.9213\,\hphantom{1}(6.0) &             0.9528\,(11.0) &             0.9525\,(10.0) &        0.8691\,\hphantom{1}(3.0) &  {\bf 0.8020\,\hphantom{1}(1.0)} &        0.8720\,\hphantom{1}(4.0) &        0.8532\,\hphantom{1}(2.0) &  0.9266\,\hphantom{1}(7.0) &  0.9156\,\hphantom{1}(5.0) \\
\textbf{Poland\_3} &                   0.9823\,(11.0) &                   0.9886\,(12.0) &  0.9265\,\hphantom{1}(8.0) &  0.9072\,\hphantom{1}(6.0) &             0.9436\,(10.0) &  0.9391\,\hphantom{1}(9.0) &        0.8688\,\hphantom{1}(4.0) &  {\bf 0.7704\,\hphantom{1}(1.0)} &        0.7981\,\hphantom{1}(2.0) &        0.8134\,\hphantom{1}(3.0) &  0.8989\,\hphantom{1}(5.0) &  0.9120\,\hphantom{1}(7.0) \\
\textbf{Poland\_4} &        0.9209\,\hphantom{1}(9.0) &                   0.9364\,(12.0) &  0.9167\,\hphantom{1}(8.0) &  0.9007\,\hphantom{1}(7.0) &             0.9309\,(11.0) &             0.9305\,(10.0) &        0.8248\,\hphantom{1}(4.0) &  {\bf 0.7812\,\hphantom{1}(1.0)} &        0.8077\,\hphantom{1}(3.0) &        0.8014\,\hphantom{1}(2.0) &  0.8900\,\hphantom{1}(6.0) &  0.8894\,\hphantom{1}(5.0) \\
\textbf{Taiwan   } &                   0.8167\,(11.0) &                   0.8455\,(12.0) &  0.7559\,\hphantom{1}(7.0) &             0.7970\,(10.0) &  0.7695\,\hphantom{1}(9.0) &  0.7688\,\hphantom{1}(8.0) &        0.5785\,\hphantom{1}(2.0) &  {\bf 0.4759\,\hphantom{1}(1.0)} &        0.6210\,\hphantom{1}(3.0) &        0.6467\,\hphantom{1}(4.0) &  0.7017\,\hphantom{1}(5.0) &  0.7105\,\hphantom{1}(6.0) \\
\end{longtable}

        }
\end{sidewaystable}

\begin{sidewaystable}
        \captionof{table}{Fraction of test examples with prediction in $\{0, 1\}$ (mean over 10 CV folds) along with their ranks. Scores in bold are best for a particular dataset.}
        \label{tab:full_certainty}
        {\tiny
        \begin{longtable}{lrrrrrrrrrrrr}
\toprule
\textbf{Classifier} &             \rotatebox{60}{NNLR} &          \rotatebox{60}{LinNNLR} &             \rotatebox{60}{ARM1} &             \rotatebox{60}{ARM2} & \rotatebox{60}{LinARM1} & \rotatebox{60}{LinARM2} &          \rotatebox{60}{MixARM1} &       \rotatebox{60}{MixLinARM1} &       \rotatebox{60}{MixMonoXGB} &           \rotatebox{60}{MixXGB} &          \rotatebox{60}{MonoXGB} &              \rotatebox{60}{XGB} \\
\textbf{Dataset  } &                                  &                                  &                                  &                                  &                         &                         &                                  &                                  &                                  &                                  &                                  &                                  \\
\midrule
\endfirsthead

\toprule
\textbf{Classifier} &             \rotatebox{60}{NNLR} &          \rotatebox{60}{LinNNLR} &             \rotatebox{60}{ARM1} &             \rotatebox{60}{ARM2} & \rotatebox{60}{LinARM1} & \rotatebox{60}{LinARM2} &          \rotatebox{60}{MixARM1} &       \rotatebox{60}{MixLinARM1} &       \rotatebox{60}{MixMonoXGB} &           \rotatebox{60}{MixXGB} &          \rotatebox{60}{MonoXGB} &              \rotatebox{60}{XGB} \\
\textbf{Dataset  } &                                  &                                  &                                  &                                  &                         &                         &                                  &                                  &                                  &                                  &                                  &                                  \\
\midrule
\endhead
\midrule
\multicolumn{13}{r}{{Continued on next page}} \\
\midrule
\endfoot

\bottomrule
\endlastfoot
\textbf{Australia} &        0.0043\,\hphantom{1}(8.0) &                   0.4623\,(10.0) &        0.0681\,\hphantom{1}(9.0) &  {\bf 0.0000\,\hphantom{1}(4.0)} &          0.7884\,(12.0) &          0.5681\,(11.0) &  {\bf 0.0000\,\hphantom{1}(4.0)} &  {\bf 0.0000\,\hphantom{1}(4.0)} &  {\bf 0.0000\,\hphantom{1}(4.0)} &  {\bf 0.0000\,\hphantom{1}(4.0)} &  {\bf 0.0000\,\hphantom{1}(4.0)} &  {\bf 0.0000\,\hphantom{1}(4.0)} \\
\textbf{GMSC     } &  {\bf 0.0000\,\hphantom{1}(5.0)} &                   0.0002\,(10.0) &  {\bf 0.0000\,\hphantom{1}(5.0)} &  {\bf 0.0000\,\hphantom{1}(5.0)} &          0.0481\,(12.0) &          0.0323\,(11.0) &  {\bf 0.0000\,\hphantom{1}(5.0)} &  {\bf 0.0000\,\hphantom{1}(5.0)} &  {\bf 0.0000\,\hphantom{1}(5.0)} &  {\bf 0.0000\,\hphantom{1}(5.0)} &  {\bf 0.0000\,\hphantom{1}(5.0)} &  {\bf 0.0000\,\hphantom{1}(5.0)} \\
\textbf{German   } &  {\bf 0.0000\,\hphantom{1}(5.0)} &                   0.0010\,(10.0) &  {\bf 0.0000\,\hphantom{1}(5.0)} &  {\bf 0.0000\,\hphantom{1}(5.0)} &          0.1930\,(12.0) &          0.0540\,(11.0) &  {\bf 0.0000\,\hphantom{1}(5.0)} &  {\bf 0.0000\,\hphantom{1}(5.0)} &  {\bf 0.0000\,\hphantom{1}(5.0)} &  {\bf 0.0000\,\hphantom{1}(5.0)} &  {\bf 0.0000\,\hphantom{1}(5.0)} &  {\bf 0.0000\,\hphantom{1}(5.0)} \\
\textbf{HELOC    } &  {\bf 0.0000\,\hphantom{1}(5.0)} &                   0.0138\,(10.0) &  {\bf 0.0000\,\hphantom{1}(5.0)} &  {\bf 0.0000\,\hphantom{1}(5.0)} &          0.0483\,(12.0) &          0.0379\,(11.0) &  {\bf 0.0000\,\hphantom{1}(5.0)} &  {\bf 0.0000\,\hphantom{1}(5.0)} &  {\bf 0.0000\,\hphantom{1}(5.0)} &  {\bf 0.0000\,\hphantom{1}(5.0)} &  {\bf 0.0000\,\hphantom{1}(5.0)} &  {\bf 0.0000\,\hphantom{1}(5.0)} \\
\textbf{Japan    } &        0.0043\,\hphantom{1}(8.0) &                   0.4783\,(10.0) &        0.1058\,\hphantom{1}(9.0) &  {\bf 0.0000\,\hphantom{1}(4.0)} &          0.7725\,(12.0) &          0.5942\,(11.0) &  {\bf 0.0000\,\hphantom{1}(4.0)} &  {\bf 0.0000\,\hphantom{1}(4.0)} &  {\bf 0.0000\,\hphantom{1}(4.0)} &  {\bf 0.0000\,\hphantom{1}(4.0)} &  {\bf 0.0000\,\hphantom{1}(4.0)} &  {\bf 0.0000\,\hphantom{1}(4.0)} \\
\textbf{LC\_0    } &  {\bf 0.0000\,\hphantom{1}(4.5)} &        0.0042\,\hphantom{1}(9.0) &                   0.1452\,(10.0) &  {\bf 0.0000\,\hphantom{1}(4.5)} &          0.8707\,(12.0) &          0.7672\,(11.0) &  {\bf 0.0000\,\hphantom{1}(4.5)} &  {\bf 0.0000\,\hphantom{1}(4.5)} &  {\bf 0.0000\,\hphantom{1}(4.5)} &  {\bf 0.0000\,\hphantom{1}(4.5)} &  {\bf 0.0000\,\hphantom{1}(4.5)} &  {\bf 0.0000\,\hphantom{1}(4.5)} \\
\textbf{LC\_1    } &  {\bf 0.0000\,\hphantom{1}(5.0)} &  {\bf 0.0000\,\hphantom{1}(5.0)} &                   0.2359\,(10.0) &  {\bf 0.0000\,\hphantom{1}(5.0)} &          0.9441\,(11.0) &          0.9499\,(12.0) &  {\bf 0.0000\,\hphantom{1}(5.0)} &  {\bf 0.0000\,\hphantom{1}(5.0)} &  {\bf 0.0000\,\hphantom{1}(5.0)} &  {\bf 0.0000\,\hphantom{1}(5.0)} &  {\bf 0.0000\,\hphantom{1}(5.0)} &  {\bf 0.0000\,\hphantom{1}(5.0)} \\
\textbf{LC\_2    } &  {\bf 0.0000\,\hphantom{1}(5.0)} &  {\bf 0.0000\,\hphantom{1}(5.0)} &                   0.1602\,(10.0) &  {\bf 0.0000\,\hphantom{1}(5.0)} &          0.8668\,(12.0) &          0.8584\,(11.0) &  {\bf 0.0000\,\hphantom{1}(5.0)} &  {\bf 0.0000\,\hphantom{1}(5.0)} &  {\bf 0.0000\,\hphantom{1}(5.0)} &  {\bf 0.0000\,\hphantom{1}(5.0)} &  {\bf 0.0000\,\hphantom{1}(5.0)} &  {\bf 0.0000\,\hphantom{1}(5.0)} \\
\textbf{LC\_3    } &  {\bf 0.0000\,\hphantom{1}(5.0)} &  {\bf 0.0000\,\hphantom{1}(5.0)} &                   0.1760\,(10.0) &  {\bf 0.0000\,\hphantom{1}(5.0)} &          0.8894\,(12.0) &          0.8700\,(11.0) &  {\bf 0.0000\,\hphantom{1}(5.0)} &  {\bf 0.0000\,\hphantom{1}(5.0)} &  {\bf 0.0000\,\hphantom{1}(5.0)} &  {\bf 0.0000\,\hphantom{1}(5.0)} &  {\bf 0.0000\,\hphantom{1}(5.0)} &  {\bf 0.0000\,\hphantom{1}(5.0)} \\
\textbf{LC\_4    } &  {\bf 0.0000\,\hphantom{1}(5.0)} &  {\bf 0.0000\,\hphantom{1}(5.0)} &                   0.1738\,(10.0) &  {\bf 0.0000\,\hphantom{1}(5.0)} &          0.8737\,(12.0) &          0.8708\,(11.0) &  {\bf 0.0000\,\hphantom{1}(5.0)} &  {\bf 0.0000\,\hphantom{1}(5.0)} &  {\bf 0.0000\,\hphantom{1}(5.0)} &  {\bf 0.0000\,\hphantom{1}(5.0)} &  {\bf 0.0000\,\hphantom{1}(5.0)} &  {\bf 0.0000\,\hphantom{1}(5.0)} \\
\textbf{LC\_5    } &  {\bf 0.0000\,\hphantom{1}(4.5)} &        0.0002\,\hphantom{1}(9.0) &                   0.1718\,(10.0) &  {\bf 0.0000\,\hphantom{1}(4.5)} &          0.8773\,(12.0) &          0.8720\,(11.0) &  {\bf 0.0000\,\hphantom{1}(4.5)} &  {\bf 0.0000\,\hphantom{1}(4.5)} &  {\bf 0.0000\,\hphantom{1}(4.5)} &  {\bf 0.0000\,\hphantom{1}(4.5)} &  {\bf 0.0000\,\hphantom{1}(4.5)} &  {\bf 0.0000\,\hphantom{1}(4.5)} \\
\textbf{LC\_6    } &  {\bf 0.0000\,\hphantom{1}(4.5)} &        0.0003\,\hphantom{1}(9.0) &                   0.1658\,(10.0) &  {\bf 0.0000\,\hphantom{1}(4.5)} &          0.8675\,(12.0) &          0.8608\,(11.0) &  {\bf 0.0000\,\hphantom{1}(4.5)} &  {\bf 0.0000\,\hphantom{1}(4.5)} &  {\bf 0.0000\,\hphantom{1}(4.5)} &  {\bf 0.0000\,\hphantom{1}(4.5)} &  {\bf 0.0000\,\hphantom{1}(4.5)} &  {\bf 0.0000\,\hphantom{1}(4.5)} \\
\textbf{LC\_7    } &  {\bf 0.0000\,\hphantom{1}(5.0)} &  {\bf 0.0000\,\hphantom{1}(5.0)} &                   0.1479\,(10.0) &  {\bf 0.0000\,\hphantom{1}(5.0)} &          0.8318\,(11.0) &          0.8395\,(12.0) &  {\bf 0.0000\,\hphantom{1}(5.0)} &  {\bf 0.0000\,\hphantom{1}(5.0)} &  {\bf 0.0000\,\hphantom{1}(5.0)} &  {\bf 0.0000\,\hphantom{1}(5.0)} &  {\bf 0.0000\,\hphantom{1}(5.0)} &  {\bf 0.0000\,\hphantom{1}(5.0)} \\
\textbf{LC\_8    } &  {\bf 0.0000\,\hphantom{1}(5.0)} &  {\bf 0.0000\,\hphantom{1}(5.0)} &                   0.1512\,(10.0) &  {\bf 0.0000\,\hphantom{1}(5.0)} &          0.8118\,(11.0) &          0.8452\,(12.0) &  {\bf 0.0000\,\hphantom{1}(5.0)} &  {\bf 0.0000\,\hphantom{1}(5.0)} &  {\bf 0.0000\,\hphantom{1}(5.0)} &  {\bf 0.0000\,\hphantom{1}(5.0)} &  {\bf 0.0000\,\hphantom{1}(5.0)} &  {\bf 0.0000\,\hphantom{1}(5.0)} \\
\textbf{LC\_9    } &  {\bf 0.0000\,\hphantom{1}(5.0)} &  {\bf 0.0000\,\hphantom{1}(5.0)} &                   0.1797\,(10.0) &  {\bf 0.0000\,\hphantom{1}(5.0)} &          0.8998\,(11.0) &          0.9143\,(12.0) &  {\bf 0.0000\,\hphantom{1}(5.0)} &  {\bf 0.0000\,\hphantom{1}(5.0)} &  {\bf 0.0000\,\hphantom{1}(5.0)} &  {\bf 0.0000\,\hphantom{1}(5.0)} &  {\bf 0.0000\,\hphantom{1}(5.0)} &  {\bf 0.0000\,\hphantom{1}(5.0)} \\
\textbf{LC\_10   } &  {\bf 0.0000\,\hphantom{1}(4.5)} &        0.0001\,\hphantom{1}(9.0) &                   0.0752\,(10.0) &  {\bf 0.0000\,\hphantom{1}(4.5)} &          0.9016\,(11.0) &          0.9070\,(12.0) &  {\bf 0.0000\,\hphantom{1}(4.5)} &  {\bf 0.0000\,\hphantom{1}(4.5)} &  {\bf 0.0000\,\hphantom{1}(4.5)} &  {\bf 0.0000\,\hphantom{1}(4.5)} &  {\bf 0.0000\,\hphantom{1}(4.5)} &  {\bf 0.0000\,\hphantom{1}(4.5)} \\
\textbf{LC\_11   } &  {\bf 0.0000\,\hphantom{1}(5.0)} &  {\bf 0.0000\,\hphantom{1}(5.0)} &                   0.1416\,(10.0) &  {\bf 0.0000\,\hphantom{1}(5.0)} &          0.9116\,(11.0) &          0.9487\,(12.0) &  {\bf 0.0000\,\hphantom{1}(5.0)} &  {\bf 0.0000\,\hphantom{1}(5.0)} &  {\bf 0.0000\,\hphantom{1}(5.0)} &  {\bf 0.0000\,\hphantom{1}(5.0)} &  {\bf 0.0000\,\hphantom{1}(5.0)} &  {\bf 0.0000\,\hphantom{1}(5.0)} \\
\textbf{LC\_12   } &        0.0001\,\hphantom{1}(8.0) &                   0.0757\,(10.0) &        0.0064\,\hphantom{1}(9.0) &  {\bf 0.0000\,\hphantom{1}(4.0)} &          0.8648\,(11.0) &          0.9223\,(12.0) &  {\bf 0.0000\,\hphantom{1}(4.0)} &  {\bf 0.0000\,\hphantom{1}(4.0)} &  {\bf 0.0000\,\hphantom{1}(4.0)} &  {\bf 0.0000\,\hphantom{1}(4.0)} &  {\bf 0.0000\,\hphantom{1}(4.0)} &  {\bf 0.0000\,\hphantom{1}(4.0)} \\
\textbf{Poland\_0} &        0.0027\,\hphantom{1}(9.0) &                   0.0129\,(10.0) &        0.0021\,\hphantom{1}(8.0) &  {\bf 0.0000\,\hphantom{1}(4.0)} &          0.6449\,(12.0) &          0.3763\,(11.0) &  {\bf 0.0000\,\hphantom{1}(4.0)} &  {\bf 0.0000\,\hphantom{1}(4.0)} &  {\bf 0.0000\,\hphantom{1}(4.0)} &  {\bf 0.0000\,\hphantom{1}(4.0)} &  {\bf 0.0000\,\hphantom{1}(4.0)} &  {\bf 0.0000\,\hphantom{1}(4.0)} \\
\textbf{Poland\_1} &        0.0008\,\hphantom{1}(9.0) &                   0.0049\,(10.0) &  {\bf 0.0000\,\hphantom{1}(4.5)} &  {\bf 0.0000\,\hphantom{1}(4.5)} &          0.3841\,(12.0) &          0.1512\,(11.0) &  {\bf 0.0000\,\hphantom{1}(4.5)} &  {\bf 0.0000\,\hphantom{1}(4.5)} &  {\bf 0.0000\,\hphantom{1}(4.5)} &  {\bf 0.0000\,\hphantom{1}(4.5)} &  {\bf 0.0000\,\hphantom{1}(4.5)} &  {\bf 0.0000\,\hphantom{1}(4.5)} \\
\textbf{Poland\_2} &        0.0006\,\hphantom{1}(9.0) &                   0.0043\,(10.0) &  {\bf 0.0000\,\hphantom{1}(4.5)} &  {\bf 0.0000\,\hphantom{1}(4.5)} &          0.4032\,(12.0) &          0.2062\,(11.0) &  {\bf 0.0000\,\hphantom{1}(4.5)} &  {\bf 0.0000\,\hphantom{1}(4.5)} &  {\bf 0.0000\,\hphantom{1}(4.5)} &  {\bf 0.0000\,\hphantom{1}(4.5)} &  {\bf 0.0000\,\hphantom{1}(4.5)} &  {\bf 0.0000\,\hphantom{1}(4.5)} \\
\textbf{Poland\_3} &        0.0011\,\hphantom{1}(9.0) &                   0.0083\,(10.0) &  {\bf 0.0000\,\hphantom{1}(4.5)} &  {\bf 0.0000\,\hphantom{1}(4.5)} &          0.4335\,(12.0) &          0.2175\,(11.0) &  {\bf 0.0000\,\hphantom{1}(4.5)} &  {\bf 0.0000\,\hphantom{1}(4.5)} &  {\bf 0.0000\,\hphantom{1}(4.5)} &  {\bf 0.0000\,\hphantom{1}(4.5)} &  {\bf 0.0000\,\hphantom{1}(4.5)} &  {\bf 0.0000\,\hphantom{1}(4.5)} \\
\textbf{Poland\_4} &        0.0034\,\hphantom{1}(9.0) &                   0.0298\,(10.0) &  {\bf 0.0000\,\hphantom{1}(4.5)} &  {\bf 0.0000\,\hphantom{1}(4.5)} &          0.5885\,(12.0) &          0.3249\,(11.0) &  {\bf 0.0000\,\hphantom{1}(4.5)} &  {\bf 0.0000\,\hphantom{1}(4.5)} &  {\bf 0.0000\,\hphantom{1}(4.5)} &  {\bf 0.0000\,\hphantom{1}(4.5)} &  {\bf 0.0000\,\hphantom{1}(4.5)} &  {\bf 0.0000\,\hphantom{1}(4.5)} \\
\textbf{Taiwan   } &        0.0001\,\hphantom{1}(9.0) &                   0.0059\,(10.0) &  {\bf 0.0000\,\hphantom{1}(4.5)} &  {\bf 0.0000\,\hphantom{1}(4.5)} &          0.0349\,(12.0) &          0.0213\,(11.0) &  {\bf 0.0000\,\hphantom{1}(4.5)} &  {\bf 0.0000\,\hphantom{1}(4.5)} &  {\bf 0.0000\,\hphantom{1}(4.5)} &  {\bf 0.0000\,\hphantom{1}(4.5)} &  {\bf 0.0000\,\hphantom{1}(4.5)} &  {\bf 0.0000\,\hphantom{1}(4.5)} \\
\end{longtable}

        }
\end{sidewaystable}
\restoregeometry
\clearpage
\section{Statistical Calculations}
\label{sec:stats}

\paragraph{Friedman omnibus test.}
The Friedman test~\cite{Friedman1940} is a non-parametric version of the ANOVA test.
Let $r_{i,j}$ be the rank of the $i$\textsuperscript{th} of $k$ algorithms on the $j$\textsuperscript{th} of $N$ datasets, where a rank of 1 corresponds to the best algorithm, 2 the second-best, and so on.
Ties are assigned the arithmetic mean of the constituent ranks, for instance, if two algorithms are joint first, then they are assigned the rank $\frac{1 + 2}{2} = 1.5$.

The Friedman test is a comparison of the average ranks over all of the datasets, $R_i = \frac{1}{N}\sum^N_{j=1} r_{i, j}$.
Under the null hypothesis that the average ranks are all equal, the Friedman statistic
\begin{equation}
    \label{eq:friedman}
    \chi^2_F = \frac{12 N}{k (k + 1)} \qty[\sum_{i=1}^k R_i^2 - \frac{k(k+1)^2}{4}]
\end{equation}
is distributed according to $\chi^2_F$ with $k - 1$ degrees of freedom.
It is well known that the Friedman statistic is often unnecessarily conservative, so we use a more accurate $F_F$ statistic~\cite{Iman1980} defined as
\begin{equation}
    \label{eq:imandavenport}
    F_F = \frac{(N-1) \chi^2_F}{ N(k - 1)- \chi^2_F}
\end{equation}
Under the null hypothesis the $F_F$ statistic is distributed according to the $F$ distribution with $k - 1$ and $(k - 1)(N - 1)$ degrees of freedom.

\paragraph{Post-hoc analysis.}
If the null hypothesis is rejected, we proceed with pairwise comparisons between the $k$ algorithms.
Fixing such a pair, let $d_j$ be the difference in performance on the $j$\textsuperscript{th} of $N$ datasets.
The Wilcoxon signed-rank test~\cite{Wilcoxon1945} is a non-parametric version of the paired $t$-test.
The differences $d_j$ are ranked according to their absolute values with average ranks being assigned in case of ties. 
Let $R^+$ be the sum of ranks for the data sets on which the second algorithm outperformed the first, and $R^-$ the sum of ranks for the converse.
Ranks of $d_j = 0$ are split evenly among the sums. 
If there is an odd number of ties, one is ignored. More precisely
\begin{equation}
    \label{eq:Rplusminus}
    R^{+} = \sum_{d_j > 0} \operatorname{rank}(d_j) + \frac{1}{2}\sum_{d_j = 0} \operatorname{rank}(d_j), \quad 
    R^{-} = \sum_{d_j < 0} \operatorname{rank}(d_j) + \frac{1}{2}\sum_{d_j = 0} \operatorname{rank}(d_j)
\end{equation}
Let $T$ be the smaller of the sums, $T = \min(R^+, R^-)$.
At $\alpha=0.05$ the exact critical value for $N=24$ is 81, that is if the smaller of $R^+$, $R^-$ is less that 81, we reject the null hypothesis.
For exact $p$-values under the null hypothesis that $d_j=0$ for all $j \in \{1, \ldots, N\}$, we use the precomputed values provided in the \texttt{scipy.stats.wilcoxon} python module~\cite{2020SciPy-NMeth}.
The $k(k - 1) / 2$ $p$-values are computed in this manner between each pair of the $k$ algorithms.

In Holm’s method of multiple hypothesis testing, the individual $p$-values are compared with adjusted $\alpha$ values as follows.
First, the $p$-values are sorted so that $p_1 \leq p_2 \leq \ldots \leq p_{k(k-1)/2}$.
Then, each $p_i$ is compared to $\frac{\alpha}{ k(k-1)/2 - i + 1}$ sequentially. 
So the most significant $p$-value, $p_1$, is compared with $\frac{\alpha}{k(k-1)/2}$. 
If $p_1$ is below $\frac{\alpha}{k(k-1)/2}$, the corresponding hypothesis is rejected and we continue to compare $p_2$ with $\frac{\alpha}{k(k-1)/2 - 1}$, and so on. 
As soon as a certain null
hypothesis cannot be rejected, all the remaining hypotheses must be retained as well.

\section{Calibration Differences Across Classifiers}
Tables~\ref{tab:comparison_ece}~and~\ref{tab:comparison_mce} respectively show point estimates for the difference in performance of the models under consideration for ECE and MCE calibration metrics.

\setlength{\tabcolsep}{2pt}
\begin{table}
\centering
        \caption{Difference in ECE scores between classifiers. Value in cell $(i, j)$ corresponds to pseudomedian over all datasets of classifier $i$ cross-validated score minus classifier $j$ cross-validated score. Bold values indicate a difference that is statistically significant.}
        \label{tab:comparison_ece}
        {\tiny
        \begin{tabular}{lrrrrrrrrrrrr}
\toprule
{} & \rotatebox{0}{NNLR} & \rotatebox{0}{LinNNLR} & \rotatebox{0}{ARM1} & \rotatebox{0}{ARM2} & \rotatebox{0}{LinARM1} & \rotatebox{0}{LinARM2} & \rotatebox{0}{MixARM1} & \rotatebox{0}{MixLinARM1} & \rotatebox{0}{MixMonoXGB} & \rotatebox{0}{MixXGB} & \rotatebox{0}{MonoXGB} & \rotatebox{0}{XGB} \\
\midrule
NNLR       &                  --- &                   0.000 &       \textbf{0.055} &       \textbf{0.046} &          \textbf{0.057} &          \textbf{0.052} &                   0.039 &                      0.039 &                      0.028 &                  0.028 &          \textbf{0.038} &               0.040 \\
LinNNLR    &                  --- &                     --- &       \textbf{0.053} &       \textbf{0.046} &          \textbf{0.057} &          \textbf{0.052} &                   0.039 &                      0.039 &                      0.028 &                  0.028 &          \textbf{0.039} &               0.040 \\
ARM1       &                  --- &                     --- &                  --- &      \textbf{-0.006} &          \textbf{0.003} &                   0.001 &         \textbf{-0.009} &            \textbf{-0.008} &            \textbf{-0.023} &        \textbf{-0.023} &         \textbf{-0.010} &     \textbf{-0.009} \\
ARM2       &                  --- &                     --- &                  --- &                  --- &          \textbf{0.009} &          \textbf{0.006} &         \textbf{-0.003} &                     -0.003 &            \textbf{-0.019} &        \textbf{-0.019} &         \textbf{-0.006} &     \textbf{-0.004} \\
LinARM1    &                  --- &                     --- &                  --- &                  --- &                     --- &         \textbf{-0.003} &         \textbf{-0.012} &            \textbf{-0.012} &            \textbf{-0.027} &        \textbf{-0.026} &         \textbf{-0.014} &     \textbf{-0.012} \\
LinARM2    &                  --- &                     --- &                  --- &                  --- &                     --- &                     --- &         \textbf{-0.010} &            \textbf{-0.009} &            \textbf{-0.024} &        \textbf{-0.024} &         \textbf{-0.011} &     \textbf{-0.010} \\
MixARM1    &                  --- &                     --- &                  --- &                  --- &                     --- &                     --- &                     --- &                     -0.000 &                     -0.009 &                 -0.008 &                   0.000 &               0.001 \\
MixLinARM1 &                  --- &                     --- &                  --- &                  --- &                     --- &                     --- &                     --- &                        --- &                     -0.008 &                 -0.008 &                   0.001 &               0.002 \\
MixMonoXGB &                  --- &                     --- &                  --- &                  --- &                     --- &                     --- &                     --- &                        --- &                        --- &                 -0.000 &          \textbf{0.011} &               0.014 \\
MixXGB     &                  --- &                     --- &                  --- &                  --- &                     --- &                     --- &                     --- &                        --- &                        --- &                    --- &          \textbf{0.011} &               0.014 \\
MonoXGB    &                  --- &                     --- &                  --- &                  --- &                     --- &                     --- &                     --- &                        --- &                        --- &                    --- &                     --- &               0.001 \\
XGB        &                  --- &                     --- &                  --- &                  --- &                     --- &                     --- &                     --- &                        --- &                        --- &                    --- &                     --- &                 --- \\
\bottomrule
\end{tabular}

        }
\end{table}

\begin{table}
\centering
        \caption{Difference in MCE scores between classifiers. Value in cell $(i, j)$ corresponds to pseudomedian over all datasets of classifier $i$ cross-validated score minus classifier $j$ cross-validated score. Bold values indicate a difference that is statistically significant.}
        \label{tab:comparison_mce}
        {\tiny
        \begin{tabular}{lrrrrrrrrrrrr}
\toprule
{} & \rotatebox{0}{NNLR} & \rotatebox{0}{LinNNLR} & \rotatebox{0}{ARM1} & \rotatebox{0}{ARM2} & \rotatebox{0}{LinARM1} & \rotatebox{0}{LinARM2} & \rotatebox{0}{MixARM1} & \rotatebox{0}{MixLinARM1} & \rotatebox{0}{MixMonoXGB} & \rotatebox{0}{MixXGB} & \rotatebox{0}{MonoXGB} & \rotatebox{0}{XGB} \\
\midrule
NNLR       &                  --- &                   0.009 &                0.206 &                0.194 &                   0.208 &                   0.213 &                   0.160 &                      0.127 &                      0.146 &                  0.152 &                   0.180 &               0.184 \\
LinNNLR    &                  --- &                     --- &                0.191 &                0.183 &                   0.195 &                   0.208 &                   0.168 &                      0.121 &                      0.139 &                  0.138 &                   0.167 &               0.172 \\
ARM1       &                  --- &                     --- &                  --- &               -0.013 &                   0.006 &                   0.009 &         \textbf{-0.034} &                     -0.068 &            \textbf{-0.042} &        \textbf{-0.056} &         \textbf{-0.021} &     \textbf{-0.018} \\
ARM2       &                  --- &                     --- &                  --- &                  --- &                   0.020 &          \textbf{0.022} &                  -0.019 &                     -0.056 &                     -0.028 &        \textbf{-0.047} &                  -0.007 &              -0.005 \\
LinARM1    &                  --- &                     --- &                  --- &                  --- &                     --- &                   0.000 &         \textbf{-0.040} &            \textbf{-0.077} &            \textbf{-0.052} &        \textbf{-0.062} &         \textbf{-0.030} &     \textbf{-0.027} \\
LinARM2    &                  --- &                     --- &                  --- &                  --- &                     --- &                     --- &         \textbf{-0.045} &            \textbf{-0.084} &            \textbf{-0.054} &        \textbf{-0.068} &         \textbf{-0.031} &     \textbf{-0.027} \\
MixARM1    &                  --- &                     --- &                  --- &                  --- &                     --- &                     --- &                     --- &                     -0.033 &                     -0.014 &                 -0.018 &                   0.005 &               0.013 \\
MixLinARM1 &                  --- &                     --- &                  --- &                  --- &                     --- &                     --- &                     --- &                        --- &                      0.015 &                  0.003 &                   0.051 &               0.050 \\
MixMonoXGB &                  --- &                     --- &                  --- &                  --- &                     --- &                     --- &                     --- &                        --- &                        --- &                 -0.000 &                   0.014 &               0.025 \\
MixXGB     &                  --- &                     --- &                  --- &                  --- &                     --- &                     --- &                     --- &                        --- &                        --- &                    --- &          \textbf{0.029} &               0.034 \\
MonoXGB    &                  --- &                     --- &                  --- &                  --- &                     --- &                     --- &                     --- &                        --- &                        --- &                    --- &                     --- &               0.002 \\
XGB        &                  --- &                     --- &                  --- &                  --- &                     --- &                     --- &                     --- &                        --- &                        --- &                    --- &                     --- &                 --- \\
\bottomrule
\end{tabular}

        }
\end{table}



\section{Model training and hyperparameters}

\paragraph{Model Hyperparameters.}
For the ARM models on the HELOC and German Credit datasets, we replicate the hyperparameters (namely feature bin edges $\theta_j$) of~\cite{CHEN2022113647}. 
For the remaining datasets we restrict each continuous feature using feature binning to 5 bins and the \texttt{DecisionTreeClassifier} from scikit-learn~\cite{scikit-learn} is trained on each individual feature, with the splits being used to decide the bins.
NNLR-based models have regularisation parameter $C=0$ in all cases.
All XGB models use the following hyperparameters, with the remaining being the default parameters used in version 1.4.2~\cite{xgboost}.
\begin{lstlisting}[language=Python, caption=XGBoost hyperparameters]
xgb_base_params = {
    "max_depth": 2,
    "n_estimators": 50,
    "learning_rate": 0.1,
    "eval_metric": 'logloss', 
    "use_label_encoder": False,
    "missing": self.binariser_kwargs["special_value_threshold"]
}
\end{lstlisting}
The value \texttt{self.binariser\_kwargs[\textcolor{codepurple}{"special\_value\_threshold"}]} corresponds to $\phi_u$ in the main text and is decided for each dataset separately.

\paragraph{Dataset preprocessing}
In terms of the subscale groupings $\mathcal{S}$, monotone constraints $(\mathcal{I}, \mathcal{D}, \mathcal{U})$ and feature lower bounds $\{\phi_u\}_{u \in [d]}$ for each dataset, for the HELOC and German datasets we replicate the assignments from~\cite{CHEN2022113647}.
In the code listings below we show all of the subscale groupings  $\mathcal{S}$ as a Python \texttt{OrderedDict} with name \texttt{<dataset>\_RC\_FEATURE\_MAPPING}, with keys corresponding to the names of the subscale, and values comprising a list of the constituent features.
Monotone constraints are stored in a Python \texttt{dict} called \texttt{<dataset>\_MONOTONE\_CONSTRAINTS}, with keys corresponding to feature names, and values in $\{-1, 1, 0\}$ corresponding to monotone decreasing, increasing and no constraint respectively.
\texttt{<dataset>\_SPECIAL\_VALUES\_DICT} contains special values for each feature that are subsequently one-hot encoded, and \texttt{<dataset>\_SPECIAL\_VALUE\_THRESHOLD} corresponds to a global feature lower bound $\phi_u$ for each dataset.
The \texttt{dict} called   \texttt{<dataset>\_MAX\_LEAF\_NODES\_DICT} corresponds to the number of allowed bins for each (continuous) feature.
The Poland dataset has uninformative attribute names, which we replace with informative names from~\cite{zikeba2016ensemble} using the Python \texttt{dict} called \texttt{POLAND\_BANKRUPTCY\_FEATURE\_MAPPING}.

\begin{lstlisting}[language=Python, caption=HELOC preprocessing]
HELOC_RC_FEATURE_MAPPING = OrderedDict({
    "ExternalRiskEstimate": ["ExternalRiskEstimate"],
    "TradeOpenTime": ["MSinceOldestTradeOpen", "MSinceMostRecentTradeOpen", "AverageMInFile"],
    "NumSatisfactoryTrades": ["NumSatisfactoryTrades"],
    "TradeFrequency": ["NumTrades60Ever2DerogPubRec", "NumTrades90Ever2DerogPubRec", "NumTotalTrades", "NumTradesOpeninLast12M"],
    "Delinquency": ["PercentTradesNeverDelq", "MSinceMostRecentDelq", "MaxDelq2PublicRecLast12M", "MaxDelqEver"],
    "Installment": ["PercentInstallTrades", "NetFractionInstallBurden", "NumInstallTradesWBalance"],
    "Inquiry": ["MSinceMostRecentInqexcl7days", "NumInqLast6M", "NumInqLast6Mexcl7days"],
    "RevolvingBalance": ["NetFractionRevolvingBurden", "NumRevolvingTradesWBalance"],
    "Utilization": ["NumBank2NatlTradesWHighUtilization"],
    "TradeWBalance": ["PercentTradesWBalance"]
})


HELOC_MONOTONE_CONSTRAINTS = {
    'ExternalRiskEstimate': -1,
    'MSinceOldestTradeOpen': -1,
    'MSinceMostRecentTradeOpen': -1,
    'AverageMInFile': -1,
    'NumSatisfactoryTrades': -1,
    'NumTrades60Ever2DerogPubRec': 1,
    'NumTrades90Ever2DerogPubRec': 1,
    'NumTotalTrades': 0,
    'NumTradesOpeninLast12M': +1,
    'PercentTradesNeverDelq': -1,
    'MSinceMostRecentDelq': -1,
    'MaxDelq2PublicRecLast12M': 0,
    'MaxDelqEver': 0,
    'PercentInstallTrades': 0,
    'NetFractionInstallBurden': +1,
    'NumInstallTradesWBalance': 0,
    'MSinceMostRecentInqexcl7days': -1,
    'NumInqLast6M': +1,
    'NumInqLast6Mexcl7days': +1,
    'NetFractionRevolvingBurden': +1,
    'NumRevolvingTradesWBalance': 0,
    'NumBank2NatlTradesWHighUtilization': +1,
    'PercentTradesWBalance': 0 
}

HELOC_CATEGORICAL_COLS = ["MaxDelq2PublicRecLast12M", "MaxDelqEver"]

HELOC_MAX_LEAF_NODES_DICT = {
    'ExternalRiskEstimate': 5,
    'MSinceOldestTradeOpen': 4,
    'MSinceMostRecentTradeOpen': 2,
    'AverageMInFile': 4,
    'NumSatisfactoryTrades': 5,
    'NumTrades60Ever2DerogPubRec': 5,
    'NumTrades90Ever2DerogPubRec': 4,
    'NumTotalTrades': 5,
    'NumTradesOpeninLast12M': 5,
    'PercentTradesNeverDelq': 5,
    'MSinceMostRecentDelq': 4,
    'MaxDelq2PublicRecLast12M': 3,
    'MaxDelqEver': 2,
    'PercentInstallTrades': 5,
    'NetFractionInstallBurden': 3,
    'NumInstallTradesWBalance': 5,
    'MSinceMostRecentInqexcl7days': 5,
    'NumInqLast6M': 4,
    'NumInqLast6Mexcl7days': 2,
    'NetFractionRevolvingBurden': 4,
    'NumRevolvingTradesWBalance': 5,
    'NumBank2NatlTradesWHighUtilization': 5,
    'PercentTradesWBalance': 5
}

HELOC_SPECIAL_VALUES_DICT = { feature: [-7, -8, -9] 
                              for feature in HELOC_MAX_LEAF_NODES_DICT.keys()
                            } 

HELOC_SPECIAL_VALUE_THRESHOLD = -0.5
\end{lstlisting}

\begin{lstlisting}[language=Python, caption=German preprocessing]
GERMAN_CREDIT_HEADERS =["Status of existing checking account","Duration in month","Credit history",\
         "Purpose","Credit amount","Savings account/bonds","Present employment since",\
         "Installment rate in percentage of disposable income","Personal status and sex",\
         "Other debtors / guarantors","Present residence since","Property","Age in years",\
        "Other installment plans","Housing","Number of existing credits at this bank",\
        "Job","Number of people being liable to provide maintenance for","Telephone","foreign worker", "Target"]

GERMAN_CREDIT_FEATURES = copy(GERMAN_CREDIT_HEADERS); GERMAN_CREDIT_FEATURES.remove("Target")

GERMAN_CREDIT_CATEGORICAL_COLS = [
                                  "Credit history",
                                  "Purpose",
                                  "Present employment since",
                                  "Personal status and sex",
                                  "Other debtors / guarantors",
                                  "Property",
                                  "Other installment plans",
                                  "Housing",
                                  "Job",
                                  "Telephone",
                                  "foreign worker"
                                 ]

GERMAN_CREDIT_NON_CATEGORICAL_COLS = list(set(GERMAN_CREDIT_FEATURES) - set(GERMAN_CREDIT_CATEGORICAL_COLS))

GERMAN_CREDIT_RC_FEATURE_MAPPING = OrderedDict({
    "CreditLoanInfo": [
        "Status of existing checking account", 
        "Credit history",
        "Purpose",
        "Savings account/bonds"
    ],
    "PersonalInfo": [
        "Present employment since",
        "Personal status and sex",
        "Other debtors / guarantors",
        "Property",
        "Other installment plans",
        "Housing",
        "Job",
        "Telephone",
        "foreign worker"
    ]
})

GERMAN_CREDIT_MONOTONE_CONSTRAINTS = {feature: 0 for feature in GERMAN_CREDIT_FEATURES}
GERMAN_CREDIT_MONOTONE_CONSTRAINTS["Status of existing checking account"] = 1
GERMAN_CREDIT_MONOTONE_CONSTRAINTS["Savings account/bonds"] = 1

GERMAN_CREDIT_MAX_LEAF_NODES_DICT = {feature: 5 for feature in GERMAN_CREDIT_NON_CATEGORICAL_COLS}

GERMAN_CREDIT_SPECIAL_VALUE_THRESHOLD = -1000

GERMAN_CREDIT_SPECIAL_VALUES_DICT = { feature: [GERMAN_CREDIT_SPECIAL_VALUE_THRESHOLD] 
                              for feature in GERMAN_CREDIT_NON_CATEGORICAL_COLS
                            } 
\end{lstlisting}

\begin{lstlisting}[language=Python, caption=Taiwan preprocessing]
TAIWAN_CREDIT_FEATURES = ['LIMIT_BAL', 'SEX', 'EDUCATION', 'MARRIAGE', 'AGE', 'PAY_0', 'PAY_2',
       'PAY_3', 'PAY_4', 'PAY_5', 'PAY_6', 'BILL_AMT1', 'BILL_AMT2',
       'BILL_AMT3', 'BILL_AMT4', 'BILL_AMT5', 'BILL_AMT6', 'PAY_AMT1',
       'PAY_AMT2', 'PAY_AMT3', 'PAY_AMT4', 'PAY_AMT5', 'PAY_AMT6']

TAIWAN_CREDIT_CATEGORICAL_COLS = [ 
    'SEX', 'EDUCATION', 'MARRIAGE',
    'PAY_0', 'PAY_2', 'PAY_3', 'PAY_4', 'PAY_5', 'PAY_6'
]

TAIWAN_CREDIT_NON_CATEGORICAL_COLS = list(set(TAIWAN_CREDIT_FEATURES) - set(TAIWAN_CREDIT_CATEGORICAL_COLS))

TAIWAN_CREDIT_RC_FEATURE_MAPPING = OrderedDict({
    "CreditLoanInfo": [
        "LIMIT_BAL"
    ],
    "PersonalInfo": [
        'SEX', 'EDUCATION', 'MARRIAGE', 'AGE'
    ],
    "RepaymentStatus": [
        'PAY_0', 'PAY_2', 'PAY_3', 'PAY_4', 'PAY_5', 'PAY_6'
    ],
    "BillAmount": [
        'BILL_AMT1', 'BILL_AMT2', 'BILL_AMT3', 'BILL_AMT4', 'BILL_AMT5', 'BILL_AMT6'
    ],
    "PaymentAmounts": [
        'PAY_AMT1', 'PAY_AMT2', 'PAY_AMT3', 'PAY_AMT4', 'PAY_AMT5', 'PAY_AMT6'
    ]
})

TAIWAN_CREDIT_MONOTONE_CONSTRAINTS = {feature: 0 for feature in TAIWAN_CREDIT_FEATURES}
TAIWAN_CREDIT_MONOTONE_CONSTRAINTS['PAY_0'] = 1
for n in range(1, 7):
    TAIWAN_CREDIT_MONOTONE_CONSTRAINTS[f'PAY_AMT{n}'] = -1

TAIWAN_CREDIT_MAX_LEAF_NODES_DICT = {feature: 5 for feature in TAIWAN_CREDIT_NON_CATEGORICAL_COLS}

TAIWAN_CREDIT_SPECIAL_VALUE_THRESHOLD = -400000

TAIWAN_CREDIT_SPECIAL_VALUES_DICT = { feature: [TAIWAN_CREDIT_SPECIAL_VALUE_THRESHOLD] 
                              for feature in TAIWAN_CREDIT_NON_CATEGORICAL_COLS
                            } 
\end{lstlisting}

\begin{lstlisting}[language=Python, caption=Give Me Some Credit preprocessing]
GIVE_ME_SOME_CREDIT_FEATURES = [
    'RevolvingUtilizationOfUnsecuredLines', 
    'age',
    'NumberOfTime30-59DaysPastDueNotWorse',
    'DebtRatio',
    'MonthlyIncome',
    'NumberOfOpenCreditLinesAndLoans',
    'NumberOfTimes90DaysLate',
    'NumberRealEstateLoansOrLines',
    'NumberOfTime60-89DaysPastDueNotWorse',
    'NumberOfDependents'
]

GIVE_ME_SOME_CREDIT_CATEGORICAL_COLS = []

GIVE_ME_SOME_CREDIT_SPECIAL_VALUES_DICT = {feature: [-1] for feature in GIVE_ME_SOME_CREDIT_FEATURES}

GIVE_ME_SOME_CREDIT_MAX_LEAF_NODES_DICT = {feature: 5 for feature in GIVE_ME_SOME_CREDIT_FEATURES}

GIVE_ME_SOME_CREDIT_SPECIAL_VALUE_THRESHOLD = -0.5

GIVE_ME_SOME_CREDIT_MONOTONE_CONSTRAINTS = {feature: 0 for feature in GIVE_ME_SOME_CREDIT_FEATURES}
for feature in ['RevolvingUtilizationOfUnsecuredLines', 'NumberOfTime30-59DaysPastDueNotWorse', 'NumberOfTime60-89DaysPastDueNotWorse', 'NumberOfTimes90DaysLate']:
    GIVE_ME_SOME_CREDIT_MONOTONE_CONSTRAINTS[feature] = 1
    
for feature in ['MonthlyIncome']:
    GIVE_ME_SOME_CREDIT_MONOTONE_CONSTRAINTS[feature] = -1

GIVE_ME_SOME_CREDIT_REASON_CODE_MAPPING = OrderedDict({
    "HistoricalLatePayments": [
        'NumberOfTime30-59DaysPastDueNotWorse', 'NumberOfTime60-89DaysPastDueNotWorse', 'NumberOfTimes90DaysLate'
    ],
    "FinancialObligations": [
        'RevolvingUtilizationOfUnsecuredLines', 'NumberOfOpenCreditLinesAndLoans', 'NumberRealEstateLoansOrLines', 'NumberOfDependents'
    ],
    "FinancialCapabilities": [
        'MonthlyIncome'
    ],
    "Demographics": [
        "age"
    ]
})
\end{lstlisting}

\begin{lstlisting}[language=Python, caption=Japan preprocessing]
# From https://www.rpubs.com/kuhnrl30/CreditScreen
JAPAN_CREDIT_PUTATIVE_FEATURE_NAMES = [
    "Male",
    "Age",
    "Debt",
    "Married",
    "BankCustomer",
    "EducationLevel",
    "Ethnicity",
    "YearsEmployed",
    "PriorDefault",
    "Employed",
    "CreditScore",
    "DriversLicense",
    "Citizen",
    "ZipCode",
    "Income"
]

JAPAN_CREDIT_NUMERIC_COLS = ["Age", "Debt", "YearsEmployed", "CreditScore", "Income"]
JAPAN_CREDIT_CATEGORICAL_COLS = list(set(JAPAN_CREDIT_PUTATIVE_FEATURE_NAMES) - set(JAPAN_CREDIT_NUMERIC_COLS))

JAPAN_CREDIT_RC_FEATURE_MAPPING = OrderedDict({
    "Demographic": ["Male", "Age", "Married", "Ethnicity", "ZipCode", "Citizen"],
    "Career": ["EducationLevel", "YearsEmployed", "Employed", "DriversLicense"],
    "FinancialObligations": ["Debt", "PriorDefault"],
    "FinancialCapabilities": ["CreditScore", "BankCustomer", "Income"]
    
})

JAPAN_CREDIT_MONOTONE_CONSTRAINTS = {feature: 0 for feature in JAPAN_CREDIT_PUTATIVE_FEATURE_NAMES}
for feature in ['Debt']:
    JAPAN_CREDIT_MONOTONE_CONSTRAINTS[feature] = 1
for feature in ['Income', 'CreditScore']:
    JAPAN_CREDIT_MONOTONE_CONSTRAINTS[feature] = -1

JAPAN_CREDIT_MAX_LEAF_NODES_DICT = {feature: 5 for feature in JAPAN_CREDIT_NUMERIC_COLS}

JAPAN_CREDIT_SPECIAL_VALUES_DICT = {feature: [-1] for feature in JAPAN_CREDIT_NUMERIC_COLS}

JAPAN_CREDIT_SPECIAL_VALUE_THRESHOLD = -0.5
\end{lstlisting}

\begin{lstlisting}[language=Python, caption=Australia preprocessing]
# From https://www.rpubs.com/kuhnrl30/CreditScreen
AUSTRALIA_CREDIT_PUTATIVE_FEATURE_NAMES = [
    "Male",
    "Age",
    "Debt",
    "BankCustomer",
    "EducationLevel",
    "Ethnicity",
    "YearsEmployed",
    "PriorDefault",
    "Employed",
    "CreditScore",
    "DriversLicense",
    "Citizen",
    "ZipCode",
    "Income"
]

AUSTRALIA_CREDIT_NUMERIC_COLS = ["Age", "Debt", "YearsEmployed", "CreditScore", "Income"]
AUSTRALIA_CREDIT_CATEGORICAL_COLS = list(set(AUSTRALIA_CREDIT_PUTATIVE_FEATURE_NAMES) - set(AUSTRALIA_CREDIT_NUMERIC_COLS))

AUSTRALIA_CREDIT_RC_FEATURE_MAPPING = OrderedDict({
    "Demographic": ["Male", "Age", "Ethnicity", "ZipCode", "Citizen"],
    "Career": ["EducationLevel", "YearsEmployed", "Employed", "DriversLicense"],
    "FinancialObligations": ["Debt", "PriorDefault"],
    "FinancialCapabilities": ["CreditScore", "BankCustomer", "Income"]
    
})

AUSTRALIA_CREDIT_MONOTONE_CONSTRAINTS = {feature: 0 for feature in AUSTRALIA_CREDIT_PUTATIVE_FEATURE_NAMES}
for feature in ['Debt']:
    AUSTRALIA_CREDIT_MONOTONE_CONSTRAINTS[feature] = 1
for feature in ['Income', 'CreditScore']:
    AUSTRALIA_CREDIT_MONOTONE_CONSTRAINTS[feature] = -1

AUSTRALIA_CREDIT_MAX_LEAF_NODES_DICT = {feature: 5 for feature in AUSTRALIA_CREDIT_NUMERIC_COLS}

AUSTRALIA_CREDIT_SPECIAL_VALUES_DICT = {feature: [-1] for feature in AUSTRALIA_CREDIT_NUMERIC_COLS}

AUSTRALIA_CREDIT_SPECIAL_VALUE_THRESHOLD = -0.5

\end{lstlisting}

\begin{lstlisting}[language=Python, caption=Poland preprocessing]
POLAND_BANKRUPTCY_FEATURE_MAPPING = {
    "Attr1" : "net profit / total assets",
    "Attr2" : "total liabilities / total assets",
    "Attr3" : "working capital / total assets",
    "Attr4" : "current assets / short-term liabilities",
    "Attr5" : "((cash + short-term securities + receivables - short-term liabilities) / (operating expenses - depreciation)) * 365",
    "Attr6" : "retained earnings / total assets",
    "Attr7" : "EBIT / total assets",
    "Attr8" : "book value of equity / total liabilities",
    "Attr9" : "sales / total assets",
    "Attr10" : "equity / total assets",
    "Attr11" : "(gross profit + extraordinary items + financial expenses) / total assets",
    "Attr12" : "gross profit / short-term liabilities",
    "Attr13" : "(gross profit + depreciation) / sales",
    "Attr14" : "(gross profit + interest) / total assets",
    "Attr15" : "(total liabilities * 365) / (gross profit + depreciation)",
    "Attr16" : "(gross profit + depreciation) / total liabilities",
    "Attr17" : "total assets / total liabilities",
    "Attr18" : "gross profit / total assets",
    "Attr19" : "gross profit / sales",
    "Attr20" : "(inventory * 365) / sales",
    "Attr21" : "sales (n) / sales (n-1)",
    "Attr22" : "profit on operating activities / total assets",
    "Attr23" : "net profit / sales",
    "Attr24" : "gross profit (in 3 years) / total assets",
    "Attr25" : "(equity - share capital) / total assets",
    "Attr26" : "(net profit + depreciation) / total liabilities",
    "Attr27" : "profit on operating activities / financial expenses",
    "Attr28" : "working capital / fixed assets",
    "Attr29" : "logarithm of total assets",
    "Attr30" : "(total liabilities - cash) / sales",
    "Attr31" : "(gross profit + interest) / sales",
    "Attr32" : "(current liabilities * 365) / cost of products sold",
    "Attr33" : "operating expenses / short-term liabilities",
    "Attr34" : "operating expenses / total liabilities",
    "Attr35" : "profit on sales / total assets",
    "Attr36" : "total sales / total assets",
    "Attr37" : "(current assets - inventories) / long-term liabilities",
    "Attr38" : "constant capital / total assets",
    "Attr39" : "profit on sales / sales",
    "Attr40" : "(current assets - inventory - receivables) / short-term liabilities",
    "Attr41" : "total liabilities / ((profit on operating activities + depreciation) * (12/365))",
    "Attr42" : "profit on operating activities / sales",
    "Attr43" : "rotation receivables + inventory turnover in days",
    "Attr44" : "(receivables * 365) / sales",
    "Attr45" : "net profit / inventory",
    "Attr46" : "(current assets - inventory) / short-term liabilities",
    "Attr47" : "(inventory * 365) / cost of products sold",
    "Attr48" : "EBITDA (profit on operating activities - depreciation) / total assets",
    "Attr49" : "EBITDA (profit on operating activities - depreciation) / sales",
    "Attr50" : "current assets / total liabilities",
    "Attr51" : "short-term liabilities / total assets",
    "Attr52" : "(short-term liabilities * 365) / cost of products sold)",
    "Attr53" : "equity / fixed assets",
    "Attr54" : "constant capital / fixed assets",
    "Attr55" : "working capital",
    "Attr56" : "(sales - cost of products sold) / sales",
    "Attr57" : "(current assets - inventory - short-term liabilities) / (sales - gross profit - depreciation)",
    "Attr58" : "total costs /total sales",
    "Attr59" : "long-term liabilities / equity",
    "Attr60" : "sales / inventory",
    "Attr61" : "sales / receivables",
    "Attr62" : "(short-term liabilities *365) / sales",
    "Attr63" : "sales / short-term liabilities",
    "Attr64" : "sales / fixed assets"
}

POLAND_BANKRUPTCY_CATEGORICAL_COLS = []
POLAND_BANKRUPTCY_NUMERIC_COLS = [
    'net profit / total assets', 'total liabilities / total assets', 'working capital / total assets', 
    'current assets / short-term liabilities', '((cash + short-term securities + receivables - short-term liabilities) / (operating expenses - depreciation)) * 365',
    'retained earnings / total assets', 'EBIT / total assets', 'book value of equity / total liabilities',
    'sales / total assets', 'equity / total assets', '(gross profit + extraordinary items + financial expenses) / total assets',
    'gross profit / short-term liabilities', '(gross profit + depreciation) / sales', '(gross profit + interest) / total assets',
    '(total liabilities * 365) / (gross profit + depreciation)', '(gross profit + depreciation) / total liabilities',
    'total assets / total liabilities', 'gross profit / total assets', 'gross profit / sales',
    '(inventory * 365) / sales', 'sales (n) / sales (n-1)', 'profit on operating activities / total assets',
    'net profit / sales', 'gross profit (in 3 years) / total assets', '(equity - share capital) / total assets', '(net profit + depreciation) / total liabilities',
    'profit on operating activities / financial expenses', 'working capital / fixed assets', 'logarithm of total assets',
    '(total liabilities - cash) / sales', '(gross profit + interest) / sales', '(current liabilities * 365) / cost of products sold',
    'operating expenses / short-term liabilities', 'operating expenses / total liabilities', 'profit on sales / total assets',
    'total sales / total assets', 'constant capital / total assets', 'profit on sales / sales',
    '(current assets - inventory - receivables) / short-term liabilities', 'total liabilities / ((profit on operating activities + depreciation) * (12/365))',
    'profit on operating activities / sales', 'rotation receivables + inventory turnover in days', '(receivables * 365) / sales',
    'net profit / inventory', '(current assets - inventory) / short-term liabilities', '(inventory * 365) / cost of products sold',
    'EBITDA (profit on operating activities - depreciation) / total assets', 'EBITDA (profit on operating activities - depreciation) / sales',
    'current assets / total liabilities', 'short-term liabilities / total assets', '(short-term liabilities * 365) / cost of products sold)',
    'equity / fixed assets', 'constant capital / fixed assets', 'working capital', '(sales - cost of products sold) / sales',
    '(current assets - inventory - short-term liabilities) / (sales - gross profit - depreciation)', 'total costs /total sales',
    'long-term liabilities / equity', 'sales / inventory', 'sales / receivables', '(short-term liabilities *365) / sales',
    'sales / short-term liabilities', 'sales / fixed assets'
]

POLAND_BANKRUPTCY_MONOTONE_CONSTRAINTS = {
    'net profit / total assets': -1,
    'total liabilities / total assets': 1,
    'working capital / total assets': -1,
    'current assets / short-term liabilities': -1,
    '((cash + short-term securities + receivables - short-term liabilities) / (operating expenses - depreciation)) * 365': -1,
    'retained earnings / total assets': -1,
    'EBIT / total assets': -1,
    'book value of equity / total liabilities': -1,
    'sales / total assets': 0,
    'equity / total assets': -1,
    '(gross profit + extraordinary items + financial expenses) / total assets': -1,
    'gross profit / short-term liabilities': -1,
    '(gross profit + depreciation) / sales': -1,
    '(gross profit + interest) / total assets': -1,
    '(total liabilities * 365) / (gross profit + depreciation)': 0,
    '(gross profit + depreciation) / total liabilities': -1,
    'total assets / total liabilities': -1,
    'gross profit / total assets': -1,
    'gross profit / sales': -1,
    '(inventory * 365) / sales': 0,
    'sales (n) / sales (n-1)': 0,
    'profit on operating activities / total assets': 0,
    'net profit / sales': -1,
    'gross profit (in 3 years) / total assets': -1,
    '(equity - share capital) / total assets': -1,
    '(net profit + depreciation) / total liabilities': -1,
    'profit on operating activities / financial expenses': 0,
    'working capital / fixed assets': 0,
    'logarithm of total assets': 0,
    '(total liabilities - cash) / sales': 1,
    '(gross profit + interest) / sales': -1,
    '(current liabilities * 365) / cost of products sold': 0,
    'operating expenses / short-term liabilities': 0,
    'operating expenses / total liabilities': 0,
    'profit on sales / total assets': 0,
    'total sales / total assets': 0,
    'constant capital / total assets': -1,
    'profit on sales / sales': 0,
    '(current assets - inventory - receivables) / short-term liabilities': -1,
    'total liabilities / ((profit on operating activities + depreciation) * (12/365))': 0,
    'profit on operating activities / sales': -1,
    'rotation receivables + inventory turnover in days': 0,
    '(receivables * 365) / sales': 0,
    'net profit / inventory': -1,
    '(current assets - inventory) / short-term liabilities': -1,
    '(inventory * 365) / cost of products sold': 0,
    'EBITDA (profit on operating activities - depreciation) / total assets': 0,
    'EBITDA (profit on operating activities - depreciation) / sales': 0,
    'current assets / total liabilities': -1,
    'short-term liabilities / total assets': 1,
    '(short-term liabilities * 365) / cost of products sold)': 0,
    'equity / fixed assets': 0,
    'constant capital / fixed assets': 0,
    'working capital': -1,
    '(sales - cost of products sold) / sales': 0,
    '(current assets - inventory - short-term liabilities) / (sales - gross profit - depreciation)': 0,
    'total costs /total sales': 0,
    'long-term liabilities / equity': 0,
    'sales / inventory': 0,
    'sales / receivables': 0,
    '(short-term liabilities *365) / sales': 1,
    'sales / short-term liabilities': -1,
    'sales / fixed assets': 0
}

POLAND_BANKRUPTCY_MAX_LEAF_NODES_DICT = {feature: 5 for feature in POLAND_BANKRUPTCY_NUMERIC_COLS}

POLAND_BANKRUPTCY_SPECIAL_VALUES_DICT = {feature: [-6] for feature in POLAND_BANKRUPTCY_NUMERIC_COLS}

POLAND_BANKRUPTCY_SPECIAL_VALUE_THRESHOLD = -5.5

POLAND_BANKRUPTCY_RC_FEATURE_MAPPING = OrderedDict({
    "Financing": [
        "(equity - share capital) / total assets"
    ],
    "CurrentRatio": [
        "total liabilities / ((profit on operating activities + depreciation) * (12/365))"
    ],
    "WorkingCapital":[
        "working capital / total assets", "working capital / fixed assets", "constant capital / total assets", "working capital",
        "constant capital / fixed assets", "logarithm of total assets"
    ],
    "LiabilitiesTurnoverRatios": [
        "equity / fixed assets", "book value of equity / total liabilities", "equity / total assets", "(net profit + depreciation) / total liabilities",
        "sales / short-term liabilities", "(current liabilities * 365) / cost of products sold", "operating expenses / short-term liabilities",
        "(short-term liabilities * 365) / cost of products sold)", "short-term liabilities / total assets", "(current assets - inventory) / short-term liabilities",
        "(current assets - inventory - receivables) / short-term liabilities", "operating expenses / total liabilities"
    ],
    "ProfitabilityRatios": [
        "(gross profit + depreciation) / sales", "profit on operating activities / total assets", "(gross profit + interest) / sales",
        "rotation receivables + inventory turnover in days", "net profit / total assets",
        "(gross profit + extraordinary items + financial expenses) / total assets", "(gross profit + interest) / total assets",
        "gross profit / total assets", "gross profit (in 3 years) / total assets"
        
    ],
    "LeverageRatios": [
        "(total liabilities * 365) / (gross profit + depreciation)", "total liabilities / total assets", "current assets / short-term liabilities",
        "total assets / total liabilities", "gross profit / short-term liabilities", "(gross profit + depreciation) / total liabilities",
        "long-term liabilities / equity",
    ],
    "TurnoverRatios": [
        "(inventory * 365) / sales", "sales (n) / sales (n-1)", "(receivables * 365) / sales", "net profit / inventory",
        "(current assets - inventory) / short-term liabilities", "(inventory * 365) / cost of products sold",
        "current assets / total liabilities",
    ],
    "OperatingPerformanceRatios": [
        "sales / total assets", "total sales / total assets", "EBITDA (profit on operating activities - depreciation) / sales",
        "((cash + short-term securities + receivables - short-term liabilities) / (operating expenses - depreciation)) * 365",
        "retained earnings / total assets", "EBIT / total assets", "EBITDA (profit on operating activities - depreciation) / total assets",
        "(current assets - inventory - short-term liabilities) / (sales - gross profit - depreciation)", "profit on operating activities / financial expenses"
    ],
    "SalesInventoryRatios": [
        "(sales - cost of products sold) / sales",
        "total costs /total sales", "sales / inventory", "sales / receivables", "net profit / inventory"
    ],
    "SalesLiabilityRatios": [
        "(short-term liabilities *365) / sales", "(total liabilities - cash) / sales"
    ],
    "ProfitabilitySalesRatios": [
        "gross profit / sales", "net profit / sales", "profit on sales / sales", "profit on operating activities / sales"
    ],
    "SalesCapitalRatios": [
        "profit on sales / total assets", "sales / fixed assets"
    ]
})

\end{lstlisting}

\begin{lstlisting}[language=Python, caption=Lending Club preprocessing]
LENDING_CLUB_SPECIAL_VALUE_THRESHOLD = -0.5

LENDING_CLUB_CATEGORICAL_COLS = [
    'term', 'emp_length', 'home_ownership', 'verification_status',
    'pymnt_plan', 'purpose', 'initial_list_status', 'application_type',
    'hardship_flag', 'disbursement_method', 'debt_settlement_flag'
]

LENDING_CLUB_NON_CATEGORICAL_COLS = [
    'loan_amnt', 'funded_amnt', 'funded_amnt_inv', 'int_rate', 'installment', 'sub_grade',
    'annual_inc', 'issue_d', 'dti', 'delinq_2yrs', 'fico_range_low', 'fico_range_high',
    'inq_last_6mths', 'mths_since_last_delinq', 'mths_since_last_record', 'open_acc', 'pub_rec', 
    'revol_bal', 'revol_util', 'total_acc', 'out_prncp', 'out_prncp_inv',
    'total_pymnt', 'total_pymnt_inv', 'total_rec_prncp', 'total_rec_int', 'total_rec_late_fee', 
    'recoveries', 'collection_recovery_fee', 'last_pymnt_amnt', 'last_fico_range_high', 
    'last_fico_range_low', 'collections_12_mths_ex_med', 'mths_since_last_major_derog', 'acc_now_delinq',
    'tot_coll_amt', 'tot_cur_bal', 'open_acc_6m', 'open_act_il',
    'open_il_12m', 'open_il_24m', 'mths_since_rcnt_il', 'total_bal_il', 'il_util',
    'open_rv_12m', 'open_rv_24m', 'max_bal_bc', 'all_util',
    'total_rev_hi_lim', 'inq_fi', 'total_cu_tl', 'inq_last_12m',
    'acc_open_past_24mths', 'avg_cur_bal', 'bc_open_to_buy', 'bc_util', 'chargeoff_within_12_mths',
    'delinq_amnt', 'mo_sin_old_il_acct', 'mo_sin_old_rev_tl_op', 'mo_sin_rcnt_rev_tl_op',
    'mo_sin_rcnt_tl', 'mort_acc', 'mths_since_recent_bc', 'mths_since_recent_bc_dlq',
    'mths_since_recent_inq', 'mths_since_recent_revol_delinq', 'num_accts_ever_120_pd', 'num_actv_bc_tl',
    'num_actv_rev_tl', 'num_bc_sats', 'num_bc_tl', 'num_il_tl',
    'num_op_rev_tl', 'num_rev_accts', 'num_rev_tl_bal_gt_0', 'num_sats',
    'num_tl_120dpd_2m', 'num_tl_30dpd', 'num_tl_90g_dpd_24m', 'num_tl_op_past_12m',
    'pct_tl_nvr_dlq', 'percent_bc_gt_75', 'pub_rec_bankruptcies', 'tax_liens',
    'tot_hi_cred_lim', 'total_bal_ex_mort', 'total_bc_limit', 'total_il_high_credit_limit'
]

LENDING_CLUB_MONOTONE_CONSTRAINTS = OrderedDict({
    'issue_d': 1,
    'tot_coll_amt': 0,
    'num_bc_sats': 1,
    'total_rev_hi_lim': -1,
    'last_fico_range_low': -1,
    'num_rev_tl_bal_gt_0': 1,
    'mths_since_recent_bc': -1,
    'revol_bal': 0,
    'mo_sin_rcnt_rev_tl_op': -1,
    'out_prncp_inv': 1,
    'total_bal_ex_mort': 0,
    'mths_since_last_delinq': 0,
    'recoveries': 0,
    'chargeoff_within_12_mths': 1,
    'fico_range_high': -1,
    'total_pymnt': -1,
    'mths_since_rcnt_il': 0,
    'open_act_il': 0,
    'bc_util': 1,
    'revol_util': 1,
    'open_il_24m': 1,
    'total_pymnt_inv': -1,
    'il_util': 1,
    'mths_since_recent_bc_dlq': 0,
    'num_rev_accts': 0,
    'pub_rec': 0,
    'num_sats': 1,
    'num_il_tl': 0,
    'out_prncp': 1,
    'all_util': 1,
    'num_tl_30dpd': 1,
    'collections_12_mths_ex_med': 1,
    'total_acc': 0,
    'num_actv_bc_tl': 1,
    'delinq_amnt': 1,
    'num_tl_op_past_12m': 1,
    'mths_since_last_major_derog': 0,
    'tot_cur_bal': 0,
    'total_rec_prncp': 0,
    'inq_last_12m': 1,
    'inq_fi': 1,
    'dti': 1,
    'num_bc_tl': 0,
    'total_rec_int': 0,
    'tot_hi_cred_lim': 0,
    'avg_cur_bal': 0,
    'funded_amnt_inv': 0,
    'num_tl_120dpd_2m': 1,
    'open_acc_6m': 1,
    'funded_amnt': 0,
    'tax_liens': 0,
    'open_rv_24m': 1,
    'percent_bc_gt_75': 0,
    'mo_sin_old_rev_tl_op': 0,
    'num_op_rev_tl': 1,
    'int_rate': 1,
    'total_cu_tl': 0,
    'pct_tl_nvr_dlq': 0,
    'num_accts_ever_120_pd': 0,
    'sub_grade': 1,
    'total_il_high_credit_limit': 0,
    'mo_sin_old_il_acct': 0,
    'mths_since_recent_revol_delinq': 0,
    'open_rv_12m': 1,
    'acc_now_delinq': 1,
    'last_fico_range_high': -1,
    'mths_since_recent_inq': -1,
    'mort_acc': -1,
    'total_bal_il': 0,
    'inq_last_6mths': 1,
    'last_pymnt_amnt': 0,
    'max_bal_bc': 0,
    'collection_recovery_fee': 0,
    'num_actv_rev_tl': 1,
    'open_il_12m': 0,
    'delinq_2yrs': 1,
    'mo_sin_rcnt_tl': -1,
    'bc_open_to_buy': -1,
    'loan_amnt': 0,
    'mths_since_last_record': 0,
    'installment': 0,
    'fico_range_low': -1,
    'total_rec_late_fee': 0,
    'open_acc': 1,
    'acc_open_past_24mths': 1,
    'annual_inc': -1,
    'num_tl_90g_dpd_24m': 0,
    'total_bc_limit': -1,
    'pub_rec_bankruptcies': 1
})

LENDING_CLUB_MONOTONE_CONSTRAINTS.update({col: 0 for col in LENDING_CLUB_CATEGORICAL_COLS})


LENDING_CLUB_SPECIAL_VALUES_DICT = { feature: [LENDING_CLUB_SPECIAL_VALUE_THRESHOLD] 
                              for feature in LENDING_CLUB_NON_CATEGORICAL_COLS
                            } 

LENDING_CLUB_RC_FEATURE_MAPPING = OrderedDict({
    "LoanInfo": [
        "loan_amnt", "funded_amnt", "funded_amnt_inv", "term", "sub_grade", "issue_d", 
        "pymnt_plan", "purpose", "initial_list_status", "application_type", "disbursement_method"
    ],
    "LoanStatus": [
        "out_prncp", "out_prncp_inv", "total_pymnt", "total_pymnt_inv", 
        "total_rec_prncp", "total_rec_int", "total_rec_late_fee", "recoveries", 
        "collection_recovery_fee", "last_pymnt_amnt"
    ],
    "PersonalInfo": [
        "emp_length", "home_ownership"
    ],
    "FinancialCapabilities": [
        "annual_inc", "verification_status", "open_acc", "total_acc", "tot_cur_bal",
        "bc_util", "bc_open_to_buy", "mort_acc", "num_actv_bc_tl",
        "num_bc_sats", "num_bc_tl", "num_sats", "tot_hi_cred_lim", 
        "total_bc_limit", "total_il_high_credit_limit", "avg_cur_bal"
    ],
    "FinancialLiabilities": [
        "dti", "percent_bc_gt_75", "pub_rec_bankruptcies", "tax_liens", "hardship_flag",
        "debt_settlement_flag", "total_bal_ex_mort"
    ],
    "ExternalRiskEstimate": [
        "fico_range_low", "fico_range_high", "last_fico_range_high", "last_fico_range_low"
    ],
    "TradeOpenTime": [
        "total_cu_tl", "acc_open_past_24mths", "mths_since_recent_bc"
    ],
    "TradeQuality": [
        "pub_rec", "pct_tl_nvr_dlq"
    ],
    "TradeFrequency": [
        "mths_since_last_record", "open_acc_6m", "mo_sin_old_il_acct", "mo_sin_old_rev_tl_op", 
        "mo_sin_rcnt_rev_tl_op", "mo_sin_rcnt_tl", "num_actv_rev_tl", "num_tl_op_past_12m"
    ],
    "Delinquency": [
        "delinq_2yrs", "mths_since_last_delinq", "collections_12_mths_ex_med", 
        "mths_since_last_major_derog", "acc_now_delinq", "tot_coll_amt", "chargeoff_within_12_mths", 
        "delinq_amnt", "mths_since_recent_bc_dlq", "mths_since_recent_revol_delinq", 
        "num_accts_ever_120_pd", "num_tl_120dpd_2m", "num_tl_30dpd", "num_tl_90g_dpd_24m", 
    ],
    "Installment": [
        "int_rate", "installment", "open_act_il", "open_il_12m", "open_il_24m",
        "mths_since_rcnt_il", "total_bal_il", "il_util", "num_il_tl"
    ],
    "Inquiry": [
        "inq_last_6mths", "inq_fi", "inq_last_12m", "mths_since_recent_inq"
    ],
    "RevolvingBalance": [
        "revol_bal", "open_rv_12m", "open_rv_24m", "max_bal_bc", "total_rev_hi_lim", 
        "num_op_rev_tl", "num_rev_accts", "num_rev_tl_bal_gt_0", 
    ],
    "Utilization": [
        "revol_util", "all_util"
    ],
})


LENDING_CLUB_MAX_LEAF_NODES_DICT = {feature: 5 for feature in LENDING_CLUB_NON_CATEGORICAL_COLS}
\end{lstlisting}

\end{document}